%% file: main.tex
\pdfoutput=1
\documentclass[11pt]{article}

\usepackage[]{acl2023}

\usepackage{times}
\usepackage{latexsym}
\usepackage[T1]{fontenc}
\usepackage[utf8]{inputenc}
\usepackage{microtype}
\usepackage{inconsolata}
\usepackage{graphicx}
\usepackage{booktabs}
\usepackage{placeins}
\usepackage{float}
\usepackage{multicol}
\usepackage{amsmath,amssymb,amsthm}
\theoremstyle{definition}
\newcommand{\twoline}[2]{\begin{tabular}[c]{@{}l@{}}#1\\#2\end{tabular}}
\newtheorem{definition}{Definition}
\usepackage{xcolor}
\usepackage[normalem]{ulem}

\input{numbers}

\input{numbers_arm}

\title{Executable Prefix Values for Locating Agent Failures}

\author{
Bingyu Shen\textsuperscript{2} \quad
Jiaqi Tang\textsuperscript{1} \quad
Lan Wei\textsuperscript{1} \quad
Yi Sheng\textsuperscript{3} \quad
J. Jenny Li\textsuperscript{1} \quad
Boyang Li\textsuperscript{1}\thanks{Corresponding author.} \\[0.5ex]
\textsuperscript{1}Department of Computer Science and Technology,
Kean University, Union, NJ, USA \\
\textsuperscript{2}Independent Researcher \\
\textsuperscript{3}Bellini College of Artificial Intelligence, \\Cybersecurity and Computing, University of South Florida, Tampa, FL, USA \\[0.5ex]
\texttt{bingyu.shen@hotmail.com} \quad
\texttt{sheng1@usf.edu} \\
\texttt{\{tangjiaq, welan, juli, boli\}@kean.edu}
}

\begin{document}
\maketitle

\begin{abstract}
Work on debugging LLM agents treats the ``first mistake'' marked by a human or LLM judge as ground truth for where a trajectory failed. We ask whether such labels point at \emph{causes} or \emph{symptoms}, and make the distinction executable: we restart the agent from reconstructed intermediate states of a failed trajectory and estimate its prefix value $V(h_t)$, the probability of success from each step, with a certified point of no return (PNR) for a failure. On \nTasks{} SWE-bench Verified tasks with two frozen open-weight agents, we build such curves for \nStudy{} and \armPrelimResolved{} failed trajectories and probe every step labeled by four frontier judges under the Traverse rubric. We found that the labels did not accurately indicate what a controller needs. Judges of different lineages agree on the exact step in only \pairFiveFiveVsOthreeExact{} cases, and the newest judges flag mistakes in \SolFlaggedDecoys{} and \armSolFlaggedDecoys{} successful runs. For \rqOneZero{} of the labels, the agent’s success probability does not drop at the labeled step; only \rqOneOn{} answers name the decisive step. Restarting at the measured restart point raises expected success from \restartNeverSuccess{} to \restartOracleSuccess{} offline, and real restarts confirm the advantage, whereas restarting at a judge's label does not (\restartJudgeGainMin{} to \restartJudgeGainMax). We release our code, curves, and labels.
\footnote{codebase:
https://anonymous.4open.science/r/laf27/}

\end{abstract}

\section{Introduction}

\begin{figure}[t]
\centering
\includegraphics[width=\columnwidth]{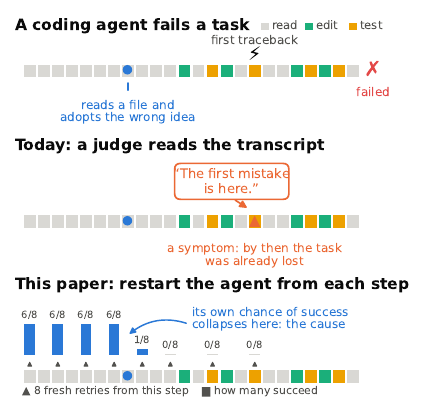}
\caption{Symptom or cause? A coding agent fails a task (top; one square per step). A judge that reads the transcript points at the first visible mistake, typically the first traceback (middle), but by then the task was already lost. We restart the agent from every step of its own run and watch its chance of success (bottom): it collapses at the step where the agent adopted the wrong idea, which is the cause. 
Figure~\ref{fig:vcurves} shows measured curves.}
\vspace{-10pt}
\label{fig:teaser}
\end{figure}

When a long-horizon agent fails, where did it go wrong? The question has two kinds of answer. A \emph{symptom} is where the failure becomes visible: the traceback, the failing test, the fix that does not apply. A \emph{cause} is the earlier step that made the symptom inevitable (but not visible yet): a wrong diagnosis adopted, a file misread, a hypothesis committed to. Debugging, restarting and learning from failures all depend on finding the cause: a controller that restarts at the symptom restarts too late, and a verifier trained on symptoms recognizes failure after the fact.

Current practice locates the first mistake by judgment: a human annotator or an LLM judge reads the trajectory and marks the earliest erroneous step \citep{rahman2026traverse,zhu2025agentdebug,zhang2025whowhen,zhang2026longrca}; benchmarks are built from such labels, verifiers are trained to reproduce them, and localizers are ranked by agreement with them. A label is a judgment about a text, made in hindsight, with no access to what the agent would have done had it continued. Nothing in the rubric distinguishes the step where the evidence appears from the step where the task was lost, and which one labels land on has not been measured: there was nothing to measure them against.

We make the distinction executable. For a stochastic agent $\pi$ the \emph{prefix value} $V(h_t)$ is the probability that the agent succeeds if it continues from history $h_t$ with the remaining budget. A failed run starts at $V(h_0)=p(x)$, the task's pass rate, and ends at $V(h_T)=0$ (Figure~\ref{fig:teaser}); the cause is where $V$ drops, by the standard of the agent's own average behaviour, and the symptom is whatever appears after. $V$ can be estimated by restarting the agent from reconstructed intermediate states of its real trajectory, repeatedly, and evaluating the outcomes with the task's own tests; we call the result a V-curve. The same quantity is what inference-time controllers need \citep{dahal2026metareasoning,wang2026failfast}: a run should be restarted when a fresh start with the remaining budget, $R(x,B-t)$, is worth more than continuing, $V(h_t)$.


We build V-curves for failed trajectories of two frozen open-weight agents on SWE-bench Verified \citep{jimenez2024swebench,openai2024verified}, with an adaptive schedule, a certified point of no return and controls for sampling noise and the shrinking budget, and place the first-mistake labels of four frontier judges on them. We ask: \textbf{RQ1} where do labels fall relative to the drop in $V$, and does $V$ move at all where a judge points? \textbf{RQ2} how are failures distributed between \emph{cliffs} and \emph{erosion}, and what is the decisive step? \textbf{RQ3} what does a restart policy gain from the measured signal, from a judge's label, or from a fixed rule, and how cheaply can the signal be obtained?

Our contributions: (i) a formulation of failure localization and of the restart decision in terms of the agent's own success probability, with graded responsibility shares and a certified point of no return (\S\ref{sec:defs}); (ii) a method that measures it on real SWE-bench trajectories, with its controls, replay checks and cost (\S\ref{sec:instrument}, \S\ref{sec:cheap}); (iii) an offline and an online evaluation of restart policies (\S\ref{sec:restart}); (iv) measurements of frontier judges against the executable signal on two agents (\S\ref{sec:judges}--\ref{sec:cases}, \S\ref{sec:arm}). The labels turn out to be neither causes nor consistently symptoms: they scatter around the decisive drop, and for \rqOneZero{} of the labels the agent's success probability does not move at the labeled step at all. What a controller needs, and what the labels do not carry, is the point at which the agent's own prospects collapse; the curves give it.

\section{Related work}

\paragraph{First-mistake labels.} Traverse \citep{rahman2026traverse} annotates several thousand agent failures with a rubric for the earliest erroneous step, finds frontier judges poor at the task, and trains Scout to reproduce the labels; Who\&When \citep{zhang2025whowhen} and LongRCA \citep{zhang2026longrca} pose failure attribution in multi-agent and long-horizon settings with low exact-step accuracy for strong models; AgentDebug \citep{zhu2025agentdebug} builds a taxonomy and a labeled benchmark and uses root-cause feedback for recovery. In all of these the reference is a judgment; we ask what the judgment measures.

\paragraph{Inference-time control.} Meta-reasoning controllers \citep{dahal2026metareasoning} decide which runs to continue, restart or halt within a budget; Fail-Fast/Restart-Smart \citep{wang2026failfast} predicts failing SWE-agent runs early and restarts them. Both need the value of a prefix; we define it, measure it, and evaluate restart policies against it. DoVer \citep{ma2025dover} replaces attribution with targeted interventions, and Prefix Sampling \citep{zhu2026rollout} resamples from prefixes for RL; the V-curve is a non-targeted intervention of the same kind, used for measurement.

\section{Definitions}
\label{sec:defs}

\paragraph{Setting.} An agent is a stochastic policy $\pi$ (an LLM with its scaffold, sampled at temperature $>0$) that attempts a task $x$ under a step budget $B$. A run is a sequence of agent actions $a_t$ and environment observations $o_t$; the \emph{history} after $t$ steps is $h_t=(x,a_1,o_1,\dots,a_t,o_t)$, with $h_0=(x)$, and the run ends after $T\le B$ steps with a binary outcome $Y\in\{0,1\}$ from the task's evaluator. We assume \emph{replayability}: the environment state reached after $h_t$ can be reconstructed exactly, so that $\pi$ can be resumed from $h_t$.

\begin{definition}[Prefix value]
For a history $h_t$ of a run under budget $B$,
\begin{equation}
V(h_t)=\Pr\!\big[Y=1 \,\big|\, h_t,\ \pi,\ \text{budget } B-t\big],
\end{equation}
the probability that the agent succeeds, resumed from $h_t$ with the remaining budget. $V(h_0)=p(x)$ is the task's pass rate under $\pi$, and $V(h_T)=Y$.
\end{definition}

\begin{definition}[Value drop and responsibility share]
The drop at step $t$ and its share of the loss are
\begin{equation}
\Delta_t=V(h_{t-1})-V(h_t),\qquad \rho_t=\Delta_t/p(x).
\end{equation}
In a deterministic environment $\Delta_t=-A^{\pi}(h_{t-1},a_t)$, the negative advantage of the action actually taken over the agent's own average behaviour at $h_{t-1}$. The drops telescope, $\sum_{t=1}^{T}\Delta_t=p(x)-Y$, so on a failed run the shares sum to one: responsibility is graded, can be spread over several steps, and a negative $\rho_t$ marks a partial recovery.
\end{definition}

\begin{definition}[Failure points]
Let $\overline{U}_{\alpha}(h_t)$ be the one-sided $(1-\alpha)$ upper confidence bound on $V(h_t)$ from its estimate, and $p_t$ the one-sided Fisher exact $p$-value for $V(h_{t-1})>V(h_t)$. The \emph{decisive step}, \emph{onset} and \emph{point of no return} (PNR) are
\begin{align}
t^*&=\arg\max_t \Delta_t,\\
t_{\text{on}}&=\min\{t : p_t<\alpha\},\\
t_{\text{pnr}}&=\min\{t : \overline{U}_{\alpha}(h_s)\le\epsilon\ \text{for all } s\ge t\}.
\end{align}
The \emph{first symptom} $t_{\text{sym}}$ is the first step after the agent's first repository edit whose observation carries an explicit failure signal (traceback, failing test, non-zero exit of a script or test run). The \emph{labeled first mistake} $m$ is the step a human or LLM judge marks under the Traverse rubric \citep{rahman2026traverse}; judges return a 0-based index $i$ and we store $m=i+1$, so the drop at the label is $\Delta_m=V(h_{m-1})-V(h_m)$.
\end{definition}

\begin{definition}[Failure shape]
With $\Delta_t^{+}=\max(\Delta_t,0)$, the concentration of the loss is
\begin{equation}
C=\max_t \Delta_t^{+}\Big/\sum_t \Delta_t^{+}\in(0,1],
\end{equation}
computed on a monotone fit of the curve (\S\ref{sec:instrument}). A failure is a \emph{cliff} if $C\ge0.5$ and \emph{erosion} otherwise; a task with $p(x)\le\epsilon_0$ ($\epsilon_0=\epsZero$ here) is \emph{doomed from the start} and has nothing to localize.
\end{definition}

\begin{definition}[Restart value and restart point]
For a budget $b$, the restart value is
\begin{equation}
R(x,b)=\Pr\!\big[Y=1,\ T\le b \,\big|\, h_0=(x),\ \pi\big],
\end{equation}
the probability that a fresh run of $x$ succeeds within $b$ steps, estimated from the $n$ base runs as $\hat R(x,b)=\tfrac{1}{n}\sum_i \mathbb{1}[Y_i=1,\ T_i\le b]$; $R(x,B)=p(x)$. A controller that restarts after $h_t$ obtains $R(x,B-t)$ in expectation and forgoes $V(h_t)$; for a margin $\mu$ the \emph{restart point} is
\begin{equation}
t_r=\min\{t : R(x,B-t)-V(h_t)\ge\mu\},
\end{equation}
and $t_{\text{pnr}}$ is its certified, task-independent counterpart, since $V(h_t)\le\epsilon$ there whatever $R$ is.
\end{definition}

\paragraph{Why $V$ is the reference.} $V$ is relative to the policy and the budget by design: it measures where \emph{this} agent lost the task, which is what a controller for this agent needs, and it involves no judgment about the text. A label $m$ can then be scored by a measurement rather than by another label: the drop at $m$ says whether anything was lost there, the position of $m$ relative to $t^*$ and $t_{\text{pnr}}$ whether it is a cause or a symptom, and the value a controller obtains by acting on $m$ what the label is worth. Every claim in this paper is one of these three comparisons.

\section{Method}
\label{sec:instrument}

Estimating $V(h_t)$ needs the exact environment state the agent saw after $h_t$ and many fresh continuations from it; we describe how each is obtained, estimated and checked.

\paragraph{Replay and anchors.} For each selected trajectory we start a fresh container from the task image and re-execute the recorded commands in order, each with its recorded timeout; after every step we compare the new observation with the recorded one and a state fingerprint (the set of changed files from \texttt{docker diff} outside caches and logs, the return code, \texttt{git diff} and \texttt{git status}) with the recorded fingerprint. Observations are compared after a normalization that masks values with a recognizable context (timestamps, durations, temporary paths, memory addresses, labelled hashes, process ids under a label), fixed on scripted dry runs before the first agent run (\nMaskRules{} rules). On the first mismatch the trajectory is non-replayable: it is dropped and the selection is back-filled from the same stratum, so that no trajectory is accepted with a reconstructed state that differs from the recorded one. At every anchor $t$ (\mbox{$t=0$}, every 5 steps to step 50, every 10 after, plus $T-1$) the container is committed as an image. A continuation from a non-anchor $h_t$ starts from the nearest earlier image, replays the steps up to $t$ under the same check, and is itself committed for later continuations from $h_t$; a mismatch there drops the trajectory as well.

\paragraph{Continuations.} A continuation from $h_t$ hands the agent the recorded messages up to $o_t$, nothing added, and lets it run with the remaining budget $B-t$ and a fresh sampling seed: from the agent's side it is the same run, resumed. Its submission is evaluated by the official harness through a cache keyed by the patch hash, so that identical patches are evaluated once and a random 10\% of unique patches is re-evaluated as a flake check. Every continuation is keyed by (trajectory, $t$, seed) and is never run twice; raising the sample size at a point adds the next seeds.

\paragraph{Schedule} (Table~\ref{tab:alg} in the appendix). (1) \emph{Screening}: $K=\Kscreen$ continuations at every anchor. (2) \emph{Bisection}: for the up to \maxBisect{} adjacent anchor pairs whose estimates differ by at least $\deltaScreen$, largest first, continuations at the midpoint, recursively, until the drop is bracketed by consecutive steps. (3) \emph{Confirmation}: both ends of the largest bracketed drop are raised to $K=\Kconfirm$ cumulative samples; a second resolved drop of at least \secondDropMin{} is confirmed too. (4) \emph{Point of no return}: from the first anchor with no success, points are raised to $K=\Kpnr$ until the one-sided 95\% Wilson upper bound \citep{wilson1927} on $V$ is below $\epsilon=\epsPnr$; the earliest certified point is $t_{\text{pnr}}$, so a certification is a statement about the agent with a known error rate, not a point estimate. (5) \emph{Label probes}: for each distinct step a judge labeled (at most four per trajectory), $V$ at $m-1$ and $m$ is estimated at $K=\Klabel$, so the drop at the label is measured directly rather than read off the fit. A curve costs \prelimMeanContPerTraj{} continuations on average.

\paragraph{Estimation.} $\hat V(h_t)$ is the success fraction of the continuations from $h_t$, with Wilson score intervals. Raw differences of a noisy curve produce spurious positive drops, which would inflate the denominator of $C$ and make every failure look like erosion. $C$ and the decisive interval are therefore read off the isotonic (monotone non-increasing, weighted least-squares) fit of the points, with the raw value reported next to it. The fit also flattens genuine recoveries, which occur on \nRecoveries{} curves (a rise beyond the 95\% intervals; Appendix~\ref{app:gallery}). $t^*$ is the largest drop of the fit between two consecutive resolved points $a<b$; we call $(a,b]$ the \emph{decisive interval} and set $t^*=a+1$, so that $b=a+1=t^*$ when bisection has resolved the drop to a single step. The onset $t_{\text{on}}$ uses a one-sided Fisher exact test at $\alpha=\alphaLevel$ between consecutive resolved points. Shares over labels carry Wilson intervals and a task-level cluster bootstrap. All parameters are in Table~\ref{tab:params}; the estimator was validated on \nSynthetic{} synthetic curves with known $t^*$ before any real curve was built (Appendix~\ref{app:synthetic}).

\paragraph{Controls and exclusions.} Three controls run on the first trajectories of the fixed processing order (Appendix~\ref{app:controls}): the continuations from $h_0$ are tested against the task's base runs (Fisher's exact test, two-sided); screening is repeated at three anchors with fresh seeds (test--retest); and the screening anchors are run again with the \emph{full} budget $B$ instead of $B-t$, to separate the loss of value from the loss of budget. Agent containers run without network and under a memory watchdog; a continuation in which a process was killed, or whose evaluation could not complete, has no outcome and is run again with the same key; the counts are reported (\S\ref{sec:checks}).

\section{Experimental setup}
\label{sec:setup}

\paragraph{Policy and scaffold.} The agent is mini-swe-agent \scaffoldVersion{} \citep{miniswe}, the minimal descendant of SWE-agent \citep{yang2024sweagent}, with its default SWE-bench configuration and prompts untouched, driving \texttt{\policyId} served by vLLM \vllmVersion{} \citep{kwon2023vllm} with thinking disabled, FP8 weights and KV cache, a \maxModelLen-token context and temperature \samplingTemperature{} (top-$p$ \samplingTopP). Every parameter is frozen before the first base run and identical for base runs and continuations. The step budget is $B=\stepBudget$.

\paragraph{Tasks and base runs.} From SWE-bench Verified (revision \texttt{\datasetRevision}) we sample \nPool{} instances uniformly (seed \taskSeed) and keep the first \nTasks{} whose images pull cleanly. Each task is run \nRunsPerTask{} times (\nBaseRuns{} runs). Patches are evaluated by the official harness \citep{jimenez2024swebench}, version \swebenchVersion; an empty patch fails without a run.

\paragraph{Selection.} Tasks with 2 to 6 successes of \nRunsPerTask{} form the \emph{band}; failed runs of band tasks are stratified by length tercile (L0--L2) and end type (submitted vs.\ step cap) and sampled with a fixed seed, at most \maxPerTask{} per task; every capped run lies in the top tercile, so the strata are L0-submit, L1-submit and L2-cap. Ten successful runs of band tasks are included blind as \emph{decoys}. Selection is fixed before any curve is seen.

\paragraph{Judges.} Each selected trajectory and each decoy is converted to Traverse's step format and given, with the Traverse SWE-bench prompt \citep{rahman2026traverse}, to four judges: GPT-6.1 Sol (primary; \texttt{gpt-6.1-sol}), GPT-5.5 (\texttt{gpt-5.5-2026-04-23}), OpenAI o3 (\texttt{o3-2025-04-16}) and GPT-6 Astra (\texttt{gpt-6-astra}). Each judge makes one pass, and three passes on a 20-trajectory subset for self-consistency. GPT-6 Luna (\texttt{gpt-6-luna}), a small model, is run as a contrast and kept out of every consensus. Each judge returns a single step index in the prompt's answer format, with a sentinel value for ``no mistake''; answers that do not follow the format are counted as unparsed (Appendix~\ref{app:judges}). 

\paragraph{An annotator-protocol label source.} To check that the results are not an artefact of one prompt, a fifth model, Claude Opus 5.5 (\texttt{claude-opus-5-5}), labelled a fixed subset of \opusN{} trajectories (the first \opusNFailed{} failed trajectories in processing order and the \opusNDecoys{} decoys) the way a human annotator would: from blind static pages showing only the task and the steps, in random order, under the same rubric. It is the only non-OpenAI model in the study, is reported separately, and enters no consensus or probe. A human annotator with a software-engineering background, blind to outcomes, curves and judge labels, labelled the same \humanN{} trajectories from the same pages under the same rubric (Appendix~\ref{app:annotator}).

\section{Results}
\label{sec:results}

\subsection{The policy and the pool}

The agent resolves \passRate{} of base runs; outcomes are strongly bimodal: \nTasksAllPass{} of \nTasks{} tasks succeed in all \nRunsPerTask{} runs and \nTasksNonePass{} in none; only \nBand{} tasks (\bandShare{} of the pool) lie in the band where a failed run is informative. Runs are \lengthMean{} steps long on average (median \lengthMedian); \capHitShare{} hit the step cap; the flake check found \flakeDisagree{} disagreements in \flakeRechecked{} re-evaluations (Appendix~\ref{app:pool}).

From the \nCandidates{} failed band runs the rule selects \nSelected{} trajectories from \nTasksSelected{} tasks (target \nTarget; at most \maxPerTask{} per task, raised from 2 before selection because the band is small). Replay reconstructs \replayableAll{} of \replayedAll{} selected, decoy and spare trajectories (\replayRateAll); with \nBackfill{} back-fills, \nStudy{} failed trajectories enter the study (decoys are judged, not continued). Non-replayable trajectories print genuinely non-deterministic values (Appendix~\ref{app:replay}).

\subsection{What the judges label}
\label{sec:judges}

Table~\ref{tab:judges} summarizes the judges before any curve is used (Appendix~\ref{app:judges}). First, the judges disagree: the most similar pair of distinct models, GPT-5.5 and OpenAI o3, marks the same step in \pairFiveFiveVsOthreeExact{} trajectories and lies within three steps in \pairFiveFiveVsOthreeWithinThree; all three main judges of different lineages agree exactly on \consensusExact{}. Second, the two GPT-6 judges are near-duplicates (\pairSolVsAstraExact{} exact), which a consensus over judges must take into account. The GPT-6 judges almost never answer ``no mistake'' on a failed trajectory (\SolNoMistakeFailed) but flag \SolFlaggedDecoys{} of the successful decoys; GPT-5.5 and OpenAI o3 are the opposite, answering ``no mistake'' on \FiveFiveNoMistakeFailed{} and \OthreeNoMistakeFailed{} of the failed trajectories, and are far less self-consistent (\FiveFiveSelfExact{} and \OthreeSelfExact{} identical answers in three passes, against \SolSelfExact{} and \AstraSelfExact).

The two annotators behave alike: they flag \opusFlaggedDecoys{} and \humanFlaggedDecoys{} decoys, answer ``no mistake'' on \opusNoMistakeFailed{} and \humanNoMistakeFailed{} failed trajectories, and agree with each other (\humanVsOpusExact{} exact) more than with any judge. 
Yet $V$ does not drop at \tabOpusZero{} of the probed labels of the model and \rqOneHumanZero{} of the human's
(Appendix~\ref{app:annotator}): not over-flagging and finding the drop are different abilities.

\begin{table}[t]
\centering
\footnotesize
\setlength{\tabcolsep}{4pt}
\begin{tabular}{lcccc}
\toprule
Judge & no mistake & flagged & position & 3 passes \\
 & (failed) & decoys & median & identical \\
\midrule
GPT-6.1 Sol & \SolNoMistakeFailed & \SolFlaggedDecoys & \SolPositionMedian & \SolSelfExact \\
GPT-6 Astra & \AstraNoMistakeFailed & \AstraFlaggedDecoys & \AstraPositionMedian & \AstraSelfExact \\
GPT-5.5 & \FiveFiveNoMistakeFailed & \FiveFiveFlaggedDecoys & \FiveFivePositionMedian & \FiveFiveSelfExact \\
OpenAI o3 & \OthreeNoMistakeFailed & \OthreeFlaggedDecoys & \OthreePositionMedian & \OthreeSelfExact \\
\midrule
\twoline{GPT-6 Luna}{(contrast)} & \LunaNoMistakeFailed & \LunaFlaggedDecoys & \LunaPositionMedian & \LunaSelfExact \\
\bottomrule
\end{tabular}
\caption{Judges on the \nStudy{} failed trajectories and \nDecoys{} decoys (pass 1). Position is the labeled step as a fraction of the trajectory. A label on a decoy is over-flagging. Self-consistency on the 20-trajectory subset.}
\vspace{-5pt}
\label{tab:judges}
\end{table}

\begin{figure*}[t]
\centering
\includegraphics[width=\textwidth]{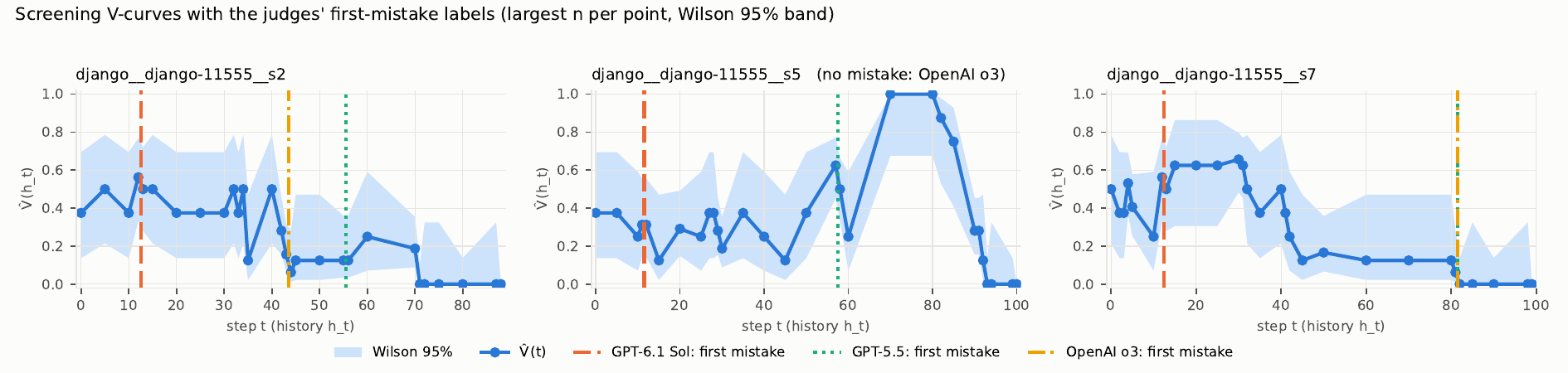}
\caption{V-curves of three resolved trajectories (largest-$n$ estimate at every point, Wilson 95\% band) with the first-mistake labels of the three main judges of different lineages drawn between $h_{m-1}$ and $h_m$. Every curve of the study is in Appendix~\ref{app:gallery}, every trajectory's numbers in Table~\ref{tab:alltraj}.}
\vspace{-5pt}

\label{fig:vcurves}
\end{figure*}

\subsection{Labels against the curves}
\label{sec:rq1}

On the \prelimResolved{} trajectories of the study (\prelimContinuations{} continuations, \prelimMeanContPerTraj{} per trajectory; \prelimPnrCertified{} with a certified point of no return), the four main judges give \prelimLabels{} labels and \prelimNone{} ``no mistake'' answers. Measured against the decisive interval $(a,b]\ni t^*$ (both endpoints at $K\ge\Kconfirm$ in \argmaxConfirmed{} of \rqOneTraj{} trajectories; elsewhere a smaller unconfirmed interval became the argmax after the confirmed one shrank) and against $t_{\text{pnr}}$, the labels scatter on both sides of the drop: \prelimBefore{} fall before the interval, \prelimAt{} inside it, \prelimAfter{} after it but at or before the point of no return, and \prelimPostPnr{} after the point of no return, where the task was already certified lost (Table~\ref{tab:rq1}). The direct measurement is the more telling one: of the \prelimProbes{} distinct labeled steps probed at $K\ge\Klabel$ on both sides, \prelimProbesZero{} (\prelimZeroShare) show no drop at all, \prelimProbesSmall{} a drop below $0.2$ and \prelimProbesBig{} one of $0.2$ or more. 
Aggregated across all four judges, \rqOneHOne of the labels (\rqOneHOneK{}; 95\% Wilson interval: \rqOneHOneCI; task-cluster bootstrap: \rqOneHOneClusterCI)fall after the decisive drop or at a step where $V$ does not drop at all — the symptom reading — and only 7/162 name the decisive step.
Only \rqOneOn{} judge answers on failed runs point at the decisive step itself. The share is \rqOneCliffHOne{} on cliffs (\rqOneCliffN{} labels) and \rqOneErosionHOne{} on erosion (\rqOneErosionN). Nor are the labels the first symptom: on runs with an explicit failure signal (\rqOneSilent{} are silent), \rqOneNearTstar{} labels lie within one step of $t^*$ and \rqOneNearTsym{} within one step of $t_{\text{sym}}$; judges mark the step at which a wrong diagnosis is stated in full or acted upon. The GPT-6 models miss early, OpenAI o3 and GPT-5.5 late (Appendix~\ref{app:rq1}).

\begin{table}[t]
\centering
\footnotesize
\setlength{\tabcolsep}{2pt}
\begin{tabular}{lccccccc}
\toprule
Source & pre & on & post & post- & none & zero- & symptom \\
 & & & & PNR & & drop & \\
\midrule
GPT-6.1 Sol & \tabSolPre & \tabSolOn & \tabSolPost & \tabSolPostPnr & \tabSolNone & \tabSolZero & \tabSolHOne \\
GPT-6 Astra & \tabAstraPre & \tabAstraOn & \tabAstraPost & \tabAstraPostPnr & \tabAstraNone & \tabAstraZero & \tabAstraHOne \\
GPT-5.5 & \tabFiveFivePre & \tabFiveFiveOn & \tabFiveFivePost & \tabFiveFivePostPnr & \tabFiveFiveNone & \tabFiveFiveZero & \tabFiveFiveHOne \\
OpenAI o3 & \tabOthreePre & \tabOthreeOn & \tabOthreePost & \tabOthreePostPnr & \tabOthreeNone & \tabOthreeZero & \tabOthreeHOne \\
\midrule
four judges & \tabPoolPre & \tabPoolOn & \tabPoolPost & \tabPoolPostPnr & \tabPoolNone & \tabPoolZero & \tabPoolHOne \\
\midrule
\twoline{GPT-6 Luna}{(contrast)} & \tabLunaPre & \tabLunaOn & \tabLunaPost & \tabLunaPostPnr & \tabLunaNone & \tabLunaZero & \tabLunaHOne \\
\twoline{Claude Opus 5.5}{(annotator)} & \tabOpusPre & \tabOpusOn & \tabOpusPost & \tabOpusPostPnr & \tabOpusNone & \tabOpusZero & \tabOpusHOne \\
\twoline{Human}{(annotator)} & \tabHumanPre & \tabHumanOn & \tabHumanPost & \tabHumanPostPnr & \tabHumanNone & \tabHumanZero & \tabHumanHOne \\
\bottomrule
\end{tabular}
\caption{Where the labels fall on the \rqOneTraj{} resolved trajectories: before the decisive interval $(a,b]$, inside it, after it, after the certified point of no return, or ``no mistake''; zero-drop counts labeled steps whose probed drop $V(h_{m-1})-V(h_m)$ is $\le0$; the last column is the share of labels that are post-cause or zero-drop.}
\label{tab:rq1}
\end{table}\vspace{-5pt}

\vspace{-5pt}
\subsection{A close reading of four trajectories}
\label{sec:cases}
\vspace{-5pt}

Appendix~\ref{app:cases} reads four trajectories step by step; two regularities recur. The decisive step is usually the step at which the agent adopts a wrong diagnosis, often an innocuous read or a probe script (on \texttt{matplotlib-25960} a \texttt{find} at step \caseMplTstar{} that turns to the wrong class, $\hat V$ from \caseMplVFive{} to \caseMplVSix); judges of the previous generation mark the action that follows, the fix or the failing test (the symptom); the GPT-6 judges mark the step where the diagnosis is stated in full, and also mark one in \SolFlaggedDecoys{} successful decoys, where no drop exists.

\subsection{What a restart policy gains}
\label{sec:restart}
\vspace{-5pt}

We evaluate restart policies offline on the measured curves: a policy that restarts a failed trajectory after $h_t$ obtains $R(x,B-t)$; A run that would have succeeded on its own forfeits that success if restarted. Figure~\ref{fig:policies} (right; every policy and trajectory in Appendix~\ref{app:restart}) reports the gain in expected success of a random run of the affected tasks under each policy, assuming (curves exist for failed runs only) that the value-based policies do not restart a run that is going to succeed, while a judge policy restarts a successful run as often as it labels a decoy.

Restarting at the restart point $t_r$ ($\mu=\restartMargin$) raises expected success from \restartNeverSuccess{} to \restartOracleSuccess{} (\restartOracleGain), at the certified point of no return to \restartPnrSuccess, and at the decisive step to \restartTstarSuccess. Restarting at a judge's label lowers it or leaves it where it was (GPT-6.1 Sol \restartSolGain, GPT-6 Astra \restartAstraGain, GPT-5.5 \restartFiveFiveGain, OpenAI o3 \restartOthreeGain; without the L2 stratum, \S\ref{sec:checks}, \sensRestartSolGain, \sensRestartAstraGain, \sensRestartFiveFiveGain, \sensRestartOthreeGain): on failed runs the labels restart late (OpenAI o3 after \restartOthreeMeanT{} steps on average) or not at all, and on successful runs the newest judges restart \decoyFalseRestartsSol{} of the decoys at an expected loss of \decoyLossSol{} each. A fixed early restart (step 10: \restartFixedTenGain) loses for the same reason. The labels do not carry the signal that distinguishes a failing prefix from a succeeding one.

\emph{Online check.} To test the value of a restart, $R(x,B-t)$, taken above from the base runs, we restarted the agent for real: \Kcheck{} fresh runs with budget $B-t_r$ and new seeds for each of the \onlineTraj{} trajectories with a restart point (\onlineRuns{} runs, mean $t_r=\onlineMeanT$). They succeed in \onlineObserved{} of cases (Wilson \onlineObservedCI) against a predicted \onlinePredicted{} and a value of continuing of \onlineContinue{}: the prediction is slightly optimistic (\onlineDiff, task-cluster 95\% interval \onlineDiffCI), as selecting restart points where $\hat R$ exceeds $\hat V$ implies, and restarting beats continuing by a wide margin (Appendix~\ref{app:online}).

\subsection{Cheaper estimates, and whether judges can guide them}
\label{sec:cheap}

Resolving a curve costs approximately \prelimMeanContPerTraj{} continuations, including around \labelProbeCost{} label probes designed strictly to test the judges. To ask how much of the rest is needed, we re-run the schedule without probes by parametric bootstrap on the measured curves, drawing outcomes from the isotonic fit of each trajectory and comparing the estimated decisive step with the fit's own largest drop (\cheapNTraj{} trajectories, \cheapReps{} replications each; the full schedule then costs \cheapFullN{} continuations; Table~\ref{tab:cheap-full} in the appendix).

Two things stand out. First, even the full schedule pins the decisive step exactly in only \cheapFullExact{} of replications (\cheapFullWithinThree{} within three steps; \cheapFullCaptured{} of the drop captured): at $K=\Kconfirm$ a drop of $0.3$ has a standard error of about $0.08$ on each side, and on erosion curves several small drops compete. Unlike a single ``decisive step'', the point of no return is robust because it is certified rather than located. This undermines benchmarks that score localizers based on exact steps. Dropping confirmation halves the cost (\cheapScreenBisectN{} continuations) at \cheapScreenBisectExact{} exact; sparse anchors (\cheapSparseN) do as well as the full schedule. Second, judge labels do not help: a schedule that probes GPT-6.1 Sol's label first and accepts it if the drop there is at least $\deltaScreen$ verifies it in \cheapJudgeSolVerified{} replications and otherwise falls back to a search, ending near the full schedule's cost (\cheapJudgeSolN{} against \cheapFullN) at the same accuracy. A proposal that is almost never where the value moves cannot save the search.

\subsection{Failure shapes and the decisive step}
\label{sec:shapes}

Of the \rqTwoN{} resolved trajectories, \rqTwoCliff{} are cliffs and \rqTwoErosion{} erosion (median concentration $C=\rqTwoCMedian$). The decisive step comes early (median $t^*/T=\rqTwoTstarFracMedian$), in \rqTwoBeforeEdit{} trajectories at or before the agent's first edit: the task is lost while the agent is still exploring. The decisive action is usually unremarkable: a read (\catTstarRead{} of decisive steps, against \catAllRead{} of all steps), rarely an edit (\catTstarEdit); the judges' labels fall disproportionately on edits (\catLabelEdit), where the wrong idea is written into the code, not where it is formed. The point of no return follows $t^*$ by a median of \rqTwoPnrGapMedian{} steps.

\emph{Interpretation and information use.} These findings suggest that consequential deterioration in agent performance frequently precedes the visible modification of the environment. In such cases, the important event may not be the acquisition of information itself, but the interpretation subsequently adopted by the agent. This distinction motivates further investigation of how agents represent evidence, distinguish observations from assumptions, and recognize when available information does not adequately support a proposed course of action.

\emph{Budget control.} On the first \bcN{} trajectories, screened again with the full budget $B$, the decisive step agrees within three steps in \bcWithinThree{} and the shape in \bcSameShape{} (Table~\ref{tab:controls}); where they differ the full budget moves $t^*$ later, so part of a late drop under $B-t$ is the shrinking budget.

\subsection{Ranking localizers}
\label{sec:rq3}

Table~\ref{tab:rq3-full} (appendix) scores every localizer two ways. The first is the value it captures: the share of the trajectory's total loss that lies within $k$ steps of the proposed step $m$, read off the isotonic fit, with a proposal of ``no mistake'' scoring zero,
\begin{equation}
\mathrm{VC}_k(m)=\frac{\max\big(0,\,V(h_{m-1-k})-V(h_{m+k})\big)}{V(h_0)-V(h_T)}.
\end{equation}
The second is agreement within one step with the labels of the judges of other lineages, the way a benchmark built from one judge's labels would score it (a lineage consensus exists on only \rqThreeConsensus{} trajectories, too few to rank on). All localizers capture little of the drop: the best, the GPT-6 judges, \vcSolOne{} of it within one step, against \vcRandomOne{} for a random step. The two rankings are only moderately related (Kendall $\tau=\rqThreeTau$, permutation $p=\rqThreeP$) and disagree where it matters: GPT-6 Luna, the small contrast model, ranks \rankLaLuna{} by agreement and \rankVcLuna{} by value captured, the GPT-6 judges \rankVcAstra{} and \rankVcSol{} by value but \rankLaSol{} and \rankLaAstra{} by agreement, and the first edit, a free heuristic, ranks \rankVcFirstEdit{} by value (\vcFirstEditOne{} against \vcFiveFiveOne{} for GPT-5.5). Agreement with labels rewards labelling like the other judges, not finding where the task was lost (Appendix~\ref{app:rq3}).

\subsection{A second policy}
\label{sec:arm}

To test agent dependence, we reran the pipeline with Qwen3.5-9B (FP8 at load time; same scaffold, tasks, seeds, selection rule, anchors and $K$).
It resolves \armPassRate{} of base runs with runs of \armLengthMean{} steps on average (median \armLengthMedian; \armCapHitShare{} hit the cap); its outcomes are less bimodal (\armNBand{} band tasks; \armNTasksNonePass{} tasks fail in all runs, \armNTasksAllPass{} succeed in all), so selection yields the full target of \armNSelected{} trajectories from \armNTasksSelected{} tasks. Replay reconstructs \armReplayRateAll{} of the selected, decoy and spare trajectories; with \armNBackfill{} back-fills, \armNStudy{} failed trajectories enter the study.

The judges behave as in \S\ref{sec:judges} (Table~\ref{tab:judges-arm}: the GPT-6 judges flag \armSolFlaggedDecoys{} decoys and agree with each other on \armPairSolVsAstraExact{} labels). The curves repeat the pattern (Figure~\ref{fig:policies}; full tables in Appendix~\ref{arm:app:all}). On the \armRqOneTraj{} resolved trajectories (\armPrelimContinuations{} continuations) the four judges give \armRqOneLabels{} labels, of which \armRqOneHOneK{} (\armRqOneHOne) are post-cause or zero-drop and only \armRqOneOn{} answers fall on the decisive step; at \armRqOneZero{} of the probed labels the success probability does not drop, and a drop of $0.2$ or more is found at \armRqOneBig. Failures are \armRqTwoErosion{} erosions and \armRqTwoCliff{} cliffs; the decisive step again comes early (median $t^*/T=\armRqTwoTstarFracMedian$). The ordering of restart policies is the same with smaller gains (\armRestartOracleGain{} at $t_r$, \armRestartTstarGain{} at $t^*$; every judge-label policy loses, \armRestartJudgeGainMin{} to \armRestartJudgeGainMax), because a fresh start is worth little to an agent that resolves \armPassRate{} of tasks and rarely within few steps. The same property makes this policy fail one check of \S\ref{sec:checks}: \armCapHitOverall{} of its continuations exhaust the remaining budget (\armCapHitLtwo{} on capped base runs, \armNBandCapShare{} of its set) and budget control moves $t^*$ later in all \armBcN{} trajectories. Consequently, ``lost under the remaining budget'' and ``lost'' nearly coincide. We present its curves as a replication of the label results, leaving its restart evaluation for the appendix.

\begin{figure}[t]
\centering
\includegraphics[width=\columnwidth]{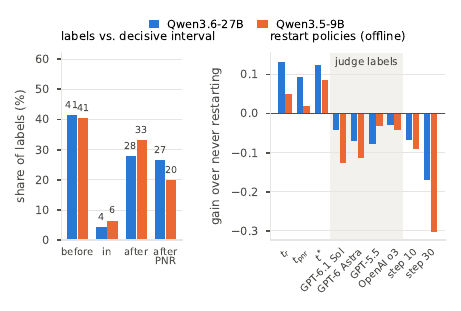}
\caption{Both policies. Left: where the four main judges' labels fall relative to the decisive interval (share of labels; before it, inside it, after it, after the certified point of no return). Right: gain over never restarting of each restart policy on a random run of the affected tasks, evaluated offline (\S\ref{sec:restart}); shaded: restarting at a judge's label (GPT-6.1 Sol, GPT-6 Astra, GPT-5.5, OpenAI o3).}
\vspace{-5pt}
\label{fig:policies}
\end{figure}

\vspace{-5pt}
\subsection{Checks on the method}
\label{sec:checks}
\vspace{-5pt}

Table~\ref{tab:instrument} (Appendix~\ref{app:criteria}) collects the checks required before any claim is made; all pass on the main policy. \nInterventions{} watchdog or kernel interventions led to \nBaseExcluded{} excluded base runs and \nContExcludedSoFar{} excluded continuations, most on one task whose natural wrong fix is an unbounded loop (excluded runs are re-sampled, which can only raise $\hat V$ there; \auditTrajErrors{} curves of that task still count \auditStaleRuns{} such runs). Continuations have a shrinking budget $B-t$, and \capHitOverall{} of the \capHitN{} continuations end by exhausting it, \capHitLzeroOne{} on the strata whose base run submitted and \capHitLtwo{} on the stratum whose base run itself hit the cap (L2); every such continuation was evaluated and failed, which is what $V(h_t)$ measures. Without the L2 stratum (\sensTraj{} trajectories; Appendix~\ref{app:rq1}, \ref{app:restart}) the symptom share is \sensHOne{} (full set \rqOneHOne) and the restart gains are \sensRestartOracleGain{} at $t_r$ and \sensRestartSolGain{} to \sensRestartFiveFiveGain{} at the judges' labels (full set \restartOracleGain; \restartJudgeGainMin{} to \restartJudgeGainMax). The second policy passes every check but this one (\armCapHitOverall; \S\ref{sec:arm}).

\vspace{-5pt}
\section{Discussion}
\vspace{-5pt}

\textbf{What do the labels measure?} The judges' labels are not simply late. On cliffs most land after the decisive drop (\rqOneCliffPostAll{} of \rqOneCliffLabels{} labels), where the wrong decision becomes visible; on erosions, which lose value late, many land before it (\rqOneErosionPre{} of \rqOneErosionLabels), on an early step that reads as a mistake but cost nothing. Judges read for evidence, which appears when a wrong decision is acted upon; the V-curve registers the decision when it is made. The GPT-6 judges, which mark the stated diagnosis, come closest; they also mark one in most successful decoys (\SolFlaggedDecoys), where no drop exists, which is why a controller that trusts them loses on both agents.

\textbf{The first mistake as a point.} \rqTwoErosion{} of the \rqTwoN{} failures are erosion: value is lost over many steps and no single step carries much of it, so even the method cannot name one step reliably (\S\ref{sec:cheap}); a rubric that asks for one forces a point onto a slope. The stable quantities are the ones a controller uses: the certified point of no return, the restart point and the value captured by a proposed step; localizers should be scored by value captured, not by agreement with a label. The curve measures the agent, the judge measures the text; the curve is the reference against which localizers and the value estimators controllers need can be built and ranked.

\vspace{-5pt}
\section*{Limitations}
\vspace{-5pt}
Two policies of one family, one scaffold, one benchmark subset; $V$ is relative to the agent and its budget (\S\ref{sec:arm}). The restart evaluation is offline and assumes a value-based policy never restarts a run that would have succeeded. Replay failures, the \capHitOverall{} of continuations that exhaust the shrinking budget and resampled killed runs are all counted and bounded by the controls (\S\ref{sec:checks}). The judges come from one vendor; one human annotator leaves inter-annotator agreement unmeasured.



\clearpage

\bibliography{custom}
\bibliographystyle{acl_natbib}

\clearpage

\appendix

\section*{Contents of the appendix}

\noindent{\small\setlength{\tabcolsep}{3pt}
\begin{tabular}{@{}lp{0.80\columnwidth}@{}}
\toprule
\ref{app:protocol} & Protocol decisions taken during the study \\
\ref{app:alg} & The adaptive schedule, step by step (Table~\ref{tab:alg}) \\
\ref{app:criteria} & Checks on the method: criteria, thresholds, values (Table~\ref{tab:instrument}) \\
\ref{app:armjudges} & Judges on the second policy (Table~\ref{tab:judges-arm}) \\
\ref{app:pairs} & Pairwise judge agreement incl.\ decoys (Table~\ref{tab:judge-pairs}) \\
\ref{app:cases} & Four trajectories read step by step \\
\ref{app:prompt} & The judge prompt, verbatim \\
\ref{app:params} & Every frozen estimator and schedule parameter (Table~\ref{tab:params}) \\
\ref{app:pool} & The base-run pool: pass@$k$, band, successes histogram (Table~\ref{tab:pool}) \\
\ref{app:trajectories} & Every trajectory: curve quantities and every source's label (Tables~\ref{tab:alltraj}--\ref{tab:alllabels-2}) \\
\ref{app:schedule} & Continuations by schedule stage, cost of a curve (Table~\ref{tab:stages}) \\
\ref{app:replay} & Replay by role and stratum, mismatch causes (Tables~\ref{tab:replay}--\ref{tab:replay-levels}) \\
\ref{app:judges} & Judge agreement matrix, per-judge detail, model identifiers (Tables~\ref{tab:judge-matrix}--\ref{tab:judge-detail}) \\
\ref{app:controls} & $t=0$ sanity, test--retest, budget control (Tables~\ref{tab:sanity}--\ref{tab:controls}) \\
\ref{app:rq1} & Labels against the curves, every source and quantity; without L2 (Tables~\ref{tab:rq1-full}--\ref{tab:rq1-noL2}) \\
\ref{app:restart} & Restart policies: summary, decoys, per trajectory; without L2 (Tables~\ref{tab:restart-full}--\ref{tab:restart-noL2}) \\
\ref{app:online} & Online restart check, per trajectory (Table~\ref{tab:online}) \\
\ref{app:cheap} & Cheaper schedules: cost, accuracy, judge-guided paths, by shape (Tables~\ref{tab:cheap-full}--\ref{tab:cheap-shape}) \\
\ref{app:rq3} & Every localizer by value captured and by label agreement (Table~\ref{tab:rq3-full}) \\
\ref{app:synthetic} & Synthetic validation of the estimator (Table~\ref{tab:synthetic}) \\
\ref{app:annotator} & The two annotators, Claude Opus 5.5 and a human (Tables~\ref{tab:annotator}--\ref{tab:human}) \\
\ref{app:gallery} & All 47 V-curves with one diagnostic note per curve (Figures~\ref{fig:gallery1}--\ref{fig:gallery4}) \\
\ref{arm:app:all} & The second policy (Qwen3.5-9B): the same tables and curves for its \armRqOneTraj{} trajectories \\
\bottomrule
\end{tabular}}

\section{Protocol decisions}
\label{app:protocol}

Estimator parameters, serving configuration and sampling were fixed before the first base run and are identical for base runs and continuations. Decisions taken during the study that bear on how the results read: the cap of trajectories per task was raised from 2 to 3 before selection (the band has \nBand{} tasks); the judge set consists of OpenAI models only, GPT-6 Astra being added after a first look at six trajectories (added, nothing removed); the $t=0$ sanity check is a two-sample Fisher test, decided before any Stage~6 analysis; continuations whose process was killed by the memory watchdog are excluded and resampled, a key excluded three times is retired; the study set of each policy is the set of trajectories resolved when its schedule, fixed at selection time, was stopped.

\section{The schedule}
\label{app:alg}

Table~\ref{tab:alg} states the adaptive schedule of \S\ref{sec:instrument} step by step. Its logic is to spend continuations where they change a conclusion: screening at every anchor is cheap and locates the region of the largest drop; bisection narrows that region to consecutive steps; confirmation raises only the two bracketing points to $K=\Kconfirm$, so that the decisive step rests on a drop with a standard error of about $0.08$; the point-of-no-return stage raises points from the first zero-success anchor until the one-sided Wilson bound certifies $V\le\epsPnr$, and bisects backward so that $t_{\text{pnr}}$ is the earliest certified step; the label probes, run last, raise $m-1$ and $m$ for every distinct judge label so that the drop at a label is a direct measurement. Continuations are cumulative per history and keyed by (trajectory, $t$, seed), so a stage never discards what an earlier stage sampled. A curve costs \prelimMeanContPerTraj{} continuations on average, of which the label probes are \labelProbeCost{} (Appendix~\ref{app:schedule}).

\begin{table}[htbp]
\centering
\footnotesize
\begin{tabular}{@{}p{0.96\columnwidth}@{}}
\toprule
\textbf{Input:} a failed trajectory $h_T$, anchors $A$, the judges' labeled steps $M$. \\
1. \emph{Screen:} for $t\in A$, $K=\Kscreen$ continuations from $h_t$; $V(h_T)=0$ exactly. \\
2. \emph{Bisect:} for the up to three adjacent anchor pairs with $\hat V(a)-\hat V(b)\ge\deltaScreen$, in order of drop size: sample $K=\Kscreen$ at $\lfloor(a+b)/2\rfloor$ and recurse into the half with the larger drop until $b-a=1$. \\
3. \emph{Confirm:} raise both ends of the largest bracketed drop to $K=\Kconfirm$; a second resolved drop of at least \secondDropMin{} is confirmed too. \\
4. \emph{Point of no return:} at the first anchor with no success, raise to $K=\Kpnr$; if its one-sided Wilson upper bound is below $\epsPnr$ (certified), bisect backward to the last earlier point with a success and raise the final candidate to $K=\Kpnr$; $t_{\text{pnr}}$ is the earliest certified point. If the anchor does not certify, try the next anchor with no success. \\
5. \emph{Label probes:} for each $m\in M$ (at most four), raise $m-1$ and $m$ to $K=\Klabel$ (to \KlabelRaised{} if the drop there is at least \labelRaiseDrop). \\
\textbf{Output:} points $(t,n,\text{successes})$, isotonic fit, $t^*$ with its interval, $C$, shape, $t_{\text{on}}$, $t_{\text{pnr}}$. \\
\bottomrule
\end{tabular}
\caption{The adaptive schedule of Stage~4. Continuations are keyed by (trajectory, $t$, seed) and reused across stages; raising $K$ at a point adds continuations with the next seeds.}
\label{tab:alg}
\end{table}\vspace{-5pt}

\section{Checks on the method: criteria}
\label{app:criteria}

Table~\ref{tab:instrument} lists the criteria the method must meet before any claim is made and their values on the frozen set. Replay must reconstruct at least 70\% of the selected trajectories (\gateCReplay); the harness must not flake (\flakeDisagree{} disagreements in \flakeRechecked{} re-evaluations, below the 3\% that would trigger double evaluation); continuations from the empty history must reproduce the base pass rate (\sanityPass{} trajectories pass Fisher's exact test); two screenings of the same history must agree (median gap \retestMedian{} over \retestN{} anchors); continuations must not exhaust their budget more often than the threshold (\capHitOverall{} of \capHitN{} against $<\capHitMax$; L0+L1 \capHitLzeroOne, L2 \capHitLtwo); and at least 60\% of trajectories must have a confirmed drop or a certified point of no return (\gateCLocalized). Every evaluation went through the patch cache (\evalCacheRuns{} runs, \evalCacheUnique{} unique patches). All criteria are met; the cap-hit rate is discussed in \S\ref{sec:checks}, where the analyses are repeated without the L2 stratum.

\begin{table}[htbp]
\centering
\footnotesize
\setlength{\tabcolsep}{3pt}
\begin{tabular}{p{0.40\columnwidth}p{0.25\columnwidth}p{0.27\columnwidth}}
\toprule
Check & Criterion & Value \\
\midrule
Replay match & $\ge70\%$ of selected & \gateCReplay \\
Harness flake rate & double-eval if $>3\%$ & \flakeDisagree/\flakeRechecked \\
$t=0$ sanity, 5 traj. & Fisher exact, $p\ge.05$ & \sanityPass \\
Test--retest, $K=\Kscreen$ & median gap $\le0.2$ & \retestMedian{} (\retestN{} anchors; max \retestMax) \\
Cap hits, continuations & $<\capHitMax$ & \capHitOverall{} of \capHitN{} (L0+L1 \capHitLzeroOne; L2 \capHitLtwo) \\
Confirmed drop or certified $t_{\text{pnr}}$ & $\ge60\%$ of traj. & \gateCLocalized \\
Evaluations via cache & all & \evalCacheRuns{} (\evalCacheUnique{} unique patches) \\
\bottomrule
\end{tabular}
\caption{Criteria the method must meet before any claim is made, and their values on the frozen set (\S\ref{sec:checks}).}
\label{tab:instrument}
\end{table}\vspace{-5pt}

\section{Judges on the second policy}
\label{app:armjudges}

Table~\ref{tab:judges-arm} repeats Table~\ref{tab:judges} for the second policy, Qwen3.5-9B, on its \armNStudy{} failed trajectories and \armNDecoys{} decoys. The lineage pattern of the main policy repeats: the GPT-6 judges answer ``no mistake'' on \armSolNoMistakeFailed{} and \armAstraNoMistakeFailed{} failed trajectories but flag \armSolFlaggedDecoys{} and \armAstraFlaggedDecoys{} of the successful decoys, while GPT-5.5 and OpenAI o3 answer ``no mistake'' on \armFiveFiveNoMistakeFailed{} and \armOthreeNoMistakeFailed{} and flag \armFiveFiveFlaggedDecoys{} and \armOthreeFlaggedDecoys. The GPT-6 judges again label early (median position \armSolPositionMedian{} and \armAstraPositionMedian{} of the trajectory against \armFiveFivePositionMedian{} and \armOthreePositionMedian) and are again the more self-consistent (\armSolSelfExact{} and \armAstraSelfExact{} identical answers in three passes against \armFiveFiveSelfExact{} and \armOthreeSelfExact). The second policy's runs are longer and more often capped (\S\ref{sec:arm}), and the judges flag all of its decoys more readily: on a weaker agent a successful run looks more like a failed one.

\begin{table}[h]
\centering
\footnotesize
\setlength{\tabcolsep}{4pt}
\begin{tabular}{lcccc}
\toprule
Judge & no mistake & flagged & position & 3 passes \\
 & (failed) & decoys & median & identical \\
\midrule
GPT-6.1 Sol & \armSolNoMistakeFailed & \armSolFlaggedDecoys & \armSolPositionMedian & \armSolSelfExact \\
GPT-6 Astra & \armAstraNoMistakeFailed & \armAstraFlaggedDecoys & \armAstraPositionMedian & \armAstraSelfExact \\
GPT-5.5 & \armFiveFiveNoMistakeFailed & \armFiveFiveFlaggedDecoys & \armFiveFivePositionMedian & \armFiveFiveSelfExact \\
OpenAI o3 & \armOthreeNoMistakeFailed & \armOthreeFlaggedDecoys & \armOthreePositionMedian & \armOthreeSelfExact \\
\midrule
\twoline{GPT-6 Luna}{(contrast)} & \armLunaNoMistakeFailed & \armLunaFlaggedDecoys & \armLunaPositionMedian & \armLunaSelfExact \\
\bottomrule
\end{tabular}
\caption{Judges on the second policy's \armNStudy{} failed trajectories and \armNDecoys{} decoys (pass 1), as in Table~\ref{tab:judges}.}
\label{tab:judges-arm}
\end{table}\vspace{-5pt}

\section{Judge pairs}
\label{app:pairs}

Table~\ref{tab:judge-pairs} gives the agreement of every pair of judges on the \nJudged{} judged trajectories of the main policy (failed trajectories and decoys together; two ``no mistake'' answers count as agreement, a ``no mistake'' never agrees with a step). The two GPT-6 judges agree exactly on \pairSolVsAstraExact{} and within three steps on \pairSolVsAstraWithinThree; the best pair of distinct lineages, GPT-5.5 and OpenAI o3, on \pairFiveFiveVsOthreeExact{} and \pairFiveFiveVsOthreeWithinThree; a GPT-6 judge and a GPT-5.5/OpenAI o3 judge on \pairSolVsFiveFiveExact{} to \pairOthreeVsAstraExact. The contrast model GPT-6 Luna agrees with GPT-5.5 (\pairFiveFiveVsLunaExact) more than with GPT-6.1 Sol (\pairSolVsLunaExact), although it shares the GPT-6 name: agreement follows the labelling style, early or late, rather than the model family's name. Appendix~\ref{app:judges} gives the same matrix restricted to failed trajectories with the within-one-step counts.

\begin{table}[h]
\centering
\small\setlength{\tabcolsep}{3pt}
\begin{tabular}{lcc}
\toprule
Pair & exact & within 3 \\
\midrule
GPT-6.1 Sol vs.\ GPT-6 Astra & \pairSolVsAstraExact & \pairSolVsAstraWithinThree \\
GPT-6.1 Sol vs.\ GPT-5.5 & \pairSolVsFiveFiveExact & \pairSolVsFiveFiveWithinThree \\
GPT-6.1 Sol vs.\ OpenAI o3 & \pairSolVsOthreeExact & \pairSolVsOthreeWithinThree \\
GPT-5.5 vs.\ OpenAI o3 & \pairFiveFiveVsOthreeExact & \pairFiveFiveVsOthreeWithinThree \\
GPT-5.5 vs.\ GPT-6 Astra & \pairFiveFiveVsAstraExact & \pairFiveFiveVsAstraWithinThree \\
OpenAI o3 vs.\ GPT-6 Astra & \pairOthreeVsAstraExact & \pairOthreeVsAstraWithinThree \\
\midrule
GPT-6.1 Sol vs.\ GPT-6 Luna & \pairSolVsLunaExact & \pairSolVsLunaWithinThree \\
GPT-5.5 vs.\ GPT-6 Luna & \pairFiveFiveVsLunaExact & \pairFiveFiveVsLunaWithinThree \\
OpenAI o3 vs.\ GPT-6 Luna & \pairOthreeVsLunaExact & \pairOthreeVsLunaWithinThree \\
\bottomrule
\end{tabular}
\caption{Agreement of pass-1 labels over \nJudged{} trajectories (failed and decoys). Two ``no mistake'' answers agree; ``no mistake'' never agrees with a step.}
\label{tab:judge-pairs}
\end{table}\vspace{-5pt}

\section{Four trajectories, read closely}
\label{app:cases}

The resolved trajectories show what the numbers mean. On \texttt{matplotlib-25960} (a \caseMplShape, $C=\caseMplC$) the agent, after confirming the \texttt{subfigures} spacing bug, reads the \texttt{SubFigure} class at step 5 ($\hat V=\caseMplVFive$) and at step \caseMplTstar{} turns to \texttt{GridSpec} with the thought that the spacing must be handled there ($\hat V=\caseMplVSix$, $K=32$ on both sides). The action itself is a harmless \texttt{find}; what the drop records is the commitment to a wrong hypothesis. Three judges mark step 17 ($\hat V=\caseMplVSeventeen$), where the agent first states the wrong diagnosis in full, and OpenAI o3 marks step 49 ($\caseMplVFortyNine$), a passing test run; by step 7 the task was already lost under this policy. On \texttt{requests-6028} (a \caseReqShape{} at $t^*=\caseReqTstar$, $t_{\text{pnr}}=\caseReqPnr$) the agent decides at step 24 that a Python security fix to \texttt{urlparse} explains the proxy failure and at step 25 writes a script to test that idea; $\hat V$ falls from \caseReqVTwentyFour{} to \caseReqVTwentyFive. The two GPT-6 judges mark step 24, GPT-5.5 and OpenAI o3 mark step 35 ($\caseReqVThirtyFive$), where the fix is applied, after the point of no return. On \texttt{sympy-13877} (\caseSymShape, $C=\caseSymC$, $t_{\text{pnr}}=\caseSymPnr$) value leaks from \caseSymVFive{} at step 5 over twenty steps while the agent pursues the idea, formed at step 7 ($\caseSymVSeven$), that \texttt{cancel()} introduces \texttt{NaN}; the GPT-6 judges mark step 7 and the other two mark step 34 ($\caseSymVThirtyFour$), the application of the fix, six steps after the point of no return. On \texttt{django-16938} (\caseDjShape, $t^*=t_{\text{pnr}}=\caseDjTstar$) the drop from \caseDjVTen{} to \caseDjVEleven{} coincides with a traceback from a probe script the agent wrote: the GPT-6 judges mark it exactly, GPT-5.5 marks the planning step five steps later ($\caseDjVSixteen$), and OpenAI o3 finds no mistake in a failed run.

\section{Judge prompt}
\label{app:prompt}

The prompt is the SWE-bench variant of Traverse's first-mistake prompt \citep{rahman2026traverse}, reproduced from the paper's source without change; each step is rendered as a numbered, tagged block containing its Thought, Action and Observation. Its fixed parts, in order (the task statement and the trajectory are inserted between them):

\begin{quote}\small
The following is a coding task and an agent's step-by-step trajectory attempting to solve it (enclosed with tags and indexed from 0). Each step contains a Thought (the agent's reasoning before acting), an Action (what the agent executed), and an Observation (the environment's response).

[Context] The agent was given a GitHub issue from a real software repository and tasked with modifying the codebase to resolve it. The agent can read files, edit code, and run tests within the repository.

[Task] \ldots\quad [Agent Trajectory] \ldots

[Definition of Error] An error is a state-changing Thought or Action that is wrong given the evidence available to the agent at that point, leads the task in an incorrect direction, and directly contributes to a wrong outcome or prevents the task from being solved correctly. The Observation is the environment's response and is never itself an error.

Your task is to review and critique the trajectory step by step. Focus on the agent's Thought and Action to judge whether the step is correct or erroneous; use the Observation only as context for what the agent saw. Once you identify an error, return the 0-based index of the step where the earliest error occurs. If all Thoughts and Actions are correct, return $-1$.

You must put your final answer (i.e., the index) in \verb|\boxed{}|.
\end{quote}
The answer $-1$ is recorded as ``no mistake''; a 0-based index $i$ is the step $m=i+1$ of the paper's 1-based notation.

\input{appendix_tables}
\input{appendix_tables_arm}

\end{document}

%% file: numbers.tex
\newcommand{\nPool}{170}
\newcommand{\nTasks}{150}
\newcommand{\nBaseRuns}{1,200}
\newcommand{\nRunsPerTask}{8}
\newcommand{\stepBudget}{100}
\newcommand{\policyId}{Qwen/Qwen3.6-27B-FP8}
\newcommand{\vllmVersion}{0.30.0}
\newcommand{\scaffoldVersion}{2.4.6}
\newcommand{\swebenchVersion}{5.0.2}
\newcommand{\datasetRevision}{78f471bf}
\newcommand{\taskSeed}{20260928}
\newcommand{\samplingTemperature}{0.70}
\newcommand{\samplingTopP}{0.80}
\newcommand{\lengthMean}{61.2}
\newcommand{\lengthMedian}{58}
\newcommand{\maxModelLen}{131,072}

\newcommand{\passRate}{0.68}

\newcommand{\nBand}{21}
\newcommand{\bandShare}{0.14}
\newcommand{\capHitShare}{0.08}

\newcommand{\nTasksAllPass}{77}
\newcommand{\nTasksNonePass}{28}
\newcommand{\nCandidates}{84}
\newcommand{\nSelected}{58}
\newcommand{\nTarget}{60}
\newcommand{\nTasksSelected}{21}
\newcommand{\nDecoys}{10}

\newcommand{\maxPerTask}{3}
\newcommand{\replayedAll}{94}
\newcommand{\replayableAll}{70}
\newcommand{\replayRateAll}{74\%}

\newcommand{\nBackfill}{6}
\newcommand{\nStudy}{47}

\newcommand{\SolNoMistakeFailed}{1/47}
\newcommand{\SolFlaggedDecoys}{8/10}
\newcommand{\SolPositionMedian}{0.23}
\newcommand{\SolSelfExact}{15/20}

\newcommand{\FiveFiveNoMistakeFailed}{11/47}
\newcommand{\FiveFiveFlaggedDecoys}{4/10}
\newcommand{\FiveFivePositionMedian}{0.39}
\newcommand{\FiveFiveSelfExact}{6/20}

\newcommand{\OthreeNoMistakeFailed}{13/47}
\newcommand{\OthreeFlaggedDecoys}{3/10}
\newcommand{\OthreePositionMedian}{0.47}
\newcommand{\OthreeSelfExact}{5/20}

\newcommand{\AstraNoMistakeFailed}{1/47}
\newcommand{\AstraFlaggedDecoys}{8/10}
\newcommand{\AstraPositionMedian}{0.23}
\newcommand{\AstraSelfExact}{19/20}

\newcommand{\LunaNoMistakeFailed}{9/47}
\newcommand{\LunaFlaggedDecoys}{3/10}
\newcommand{\LunaPositionMedian}{0.44}
\newcommand{\LunaSelfExact}{8/20}

\newcommand{\nJudged}{57}
\newcommand{\consensusExact}{3/57}

\newcommand{\pairSolVsFiveFiveExact}{16/57}
\newcommand{\pairSolVsFiveFiveWithinThree}{18/57}
\newcommand{\pairSolVsOthreeExact}{9/57}
\newcommand{\pairSolVsOthreeWithinThree}{12/57}
\newcommand{\pairSolVsAstraExact}{52/57}
\newcommand{\pairSolVsAstraWithinThree}{53/57}
\newcommand{\pairSolVsLunaExact}{14/57}
\newcommand{\pairSolVsLunaWithinThree}{20/57}
\newcommand{\pairFiveFiveVsOthreeExact}{21/57}
\newcommand{\pairFiveFiveVsOthreeWithinThree}{24/57}
\newcommand{\pairFiveFiveVsAstraExact}{15/57}
\newcommand{\pairFiveFiveVsAstraWithinThree}{17/57}
\newcommand{\pairFiveFiveVsLunaExact}{24/57}
\newcommand{\pairFiveFiveVsLunaWithinThree}{30/57}
\newcommand{\pairOthreeVsAstraExact}{9/57}
\newcommand{\pairOthreeVsAstraWithinThree}{12/57}
\newcommand{\pairOthreeVsLunaExact}{21/57}
\newcommand{\pairOthreeVsLunaWithinThree}{25/57}

\newcommand{\evalCacheUnique}{805}
\newcommand{\evalCacheRuns}{1,200}

\newcommand{\nInterventions}{327}
\newcommand{\nBaseExcluded}{4}
\newcommand{\nContExcludedSoFar}{178}
\newcommand{\Kscreen}{8}
\newcommand{\Kconfirm}{32}
\newcommand{\Klabel}{16}
\newcommand{\Kpnr}{24}
\newcommand{\epsPnr}{0.15}
\newcommand{\maxBisect}{3}
\newcommand{\epsZero}{1/8}
\newcommand{\alphaLevel}{0.05}
\newcommand{\nMaskRules}{26}
\newcommand{\nSynthetic}{2,000}
\newcommand{\secondDropMin}{0.25}
\newcommand{\KlabelRaised}{32}
\newcommand{\labelRaiseDrop}{0.20}
\newcommand{\Kcheck}{8}
\newcommand{\deltaScreen}{0.25}

\newcommand{\prelimResolved}{47}
\newcommand{\prelimContinuations}{15,576}
\newcommand{\prelimMeanContPerTraj}{331}

\newcommand{\prelimPnrCertified}{47}
\newcommand{\prelimLabels}{162}
\newcommand{\prelimBefore}{67}
\newcommand{\prelimAt}{7}
\newcommand{\prelimAfter}{45}
\newcommand{\prelimPostPnr}{43}
\newcommand{\prelimNone}{26}
\newcommand{\prelimProbes}{95}
\newcommand{\prelimProbesZero}{64}
\newcommand{\prelimProbesSmall}{27}
\newcommand{\prelimProbesBig}{4}

\newcommand{\prelimZeroShare}{67\%}
\newcommand{\argmaxConfirmed}{13}

\newcommand{\labelProbeCost}{55}
\newcommand{\retestMedian}{0.00}
\newcommand{\retestN}{30}
\newcommand{\retestMax}{0.38}
\newcommand{\sanityPass}{5/5}

\newcommand{\restartNeverSuccess}{0.44}

\newcommand{\restartOracleSuccess}{0.57}
\newcommand{\restartOracleGain}{+0.13}

\newcommand{\restartPnrSuccess}{0.53}

\newcommand{\restartTstarSuccess}{0.56}

\newcommand{\restartSolGain}{-0.04}

\newcommand{\restartFiveFiveGain}{-0.08}

\newcommand{\restartOthreeGain}{-0.03}

\newcommand{\restartOthreeMeanT}{44}

\newcommand{\restartAstraGain}{-0.07}

\newcommand{\restartFixedTenGain}{-0.07}

\newcommand{\restartJudgeGainMin}{-0.08}
\newcommand{\restartJudgeGainMax}{-0.03}
\newcommand{\decoyLossSol}{0.46}
\newcommand{\decoyFalseRestartsSol}{8/10}

\newcommand{\restartMargin}{0.10}

\newcommand{\cheapFullN}{276}
\newcommand{\cheapFullExact}{28\%}
\newcommand{\cheapFullWithinThree}{47\%}
\newcommand{\cheapFullCaptured}{0.44}

\newcommand{\cheapScreenBisectN}{147}
\newcommand{\cheapScreenBisectExact}{16\%}

\newcommand{\cheapSparseN}{230}

\newcommand{\cheapJudgeSolN}{255}

\newcommand{\cheapJudgeSolVerified}{40/940}

\newcommand{\cheapReps}{20}
\newcommand{\cheapNTraj}{47}

\newcommand{\rqOneTraj}{47}

\newcommand{\rqOneHOne}{81\%}
\newcommand{\rqOneHOneK}{132/162}
\newcommand{\rqOneHOneCI}{75--87\%}
\newcommand{\rqOneHOneClusterCI}{70--92\%}
\newcommand{\rqOneOn}{7/188}
\newcommand{\rqOneZero}{66\%}

\newcommand{\rqOneNearTstar}{17/138}
\newcommand{\rqOneNearTsym}{3/138}
\newcommand{\rqOneCliffHOne}{91\%}
\newcommand{\rqOneErosionHOne}{77\%}
\newcommand{\rqOneCliffN}{53}
\newcommand{\rqOneErosionN}{109}
\newcommand{\rqOneSilent}{7/47}

\newcommand{\rqOneHumanZero}{5/10}

\newcommand{\flakeRechecked}{46}
\newcommand{\flakeDisagree}{0}

\newcommand{\rqOneCliffPostAll}{29}
\newcommand{\rqOneCliffLabels}{53}
\newcommand{\rqOneErosionPre}{46}

\newcommand{\rqOneErosionLabels}{109}
\newcommand{\rankVcAstra}{first}
\newcommand{\rankVcSol}{second}

\newcommand{\rankVcLuna}{fourth}
\newcommand{\rankVcFirstEdit}{fifth}

\newcommand{\rankLaLuna}{first}

\newcommand{\rankLaSol}{third}
\newcommand{\rankLaAstra}{fourth}

\newcommand{\onlineTraj}{30}
\newcommand{\onlineRuns}{240}
\newcommand{\onlineObserved}{0.39}
\newcommand{\onlineObservedCI}{0.33--0.45}
\newcommand{\onlinePredicted}{0.45}
\newcommand{\onlineContinue}{0.26}
\newcommand{\onlineDiff}{-0.06}
\newcommand{\onlineDiffCI}{-0.15 to +0.04}
\newcommand{\onlineMeanT}{7}

\newcommand{\sensTraj}{29}

\newcommand{\sensHOne}{85\%}

\newcommand{\sensRestartOracleGain}{+0.17}

\newcommand{\sensRestartSolGain}{-0.01}

\newcommand{\sensRestartFiveFiveGain}{-0.04}

\newcommand{\sensRestartOthreeGain}{+0.01}

\newcommand{\sensRestartAstraGain}{-0.04}

\newcommand{\humanN}{40}

\newcommand{\humanFlaggedDecoys}{1/10}
\newcommand{\humanNoMistakeFailed}{7/30}
\newcommand{\humanVsOpusExact}{15/30}

\newcommand{\auditTrajErrors}{2}
\newcommand{\auditStaleRuns}{15}

\newcommand{\nRecoveries}{2}

\newcommand{\capHitOverall}{31\%}
\newcommand{\capHitN}{15,576}
\newcommand{\capHitMax}{35\%}

\newcommand{\capHitLtwo}{62\%}
\newcommand{\capHitLzeroOne}{6\%}
\newcommand{\gateCLocalized}{47/47}
\newcommand{\gateCReplay}{73\%}

\newcommand{\tabSolPre}{23}
\newcommand{\tabSolOn}{3}
\newcommand{\tabSolPost}{12}
\newcommand{\tabSolPostPnr}{8}
\newcommand{\tabSolNone}{1}
\newcommand{\tabSolZero}{30/46}
\newcommand{\tabSolHOne}{78\%}
\newcommand{\tabFiveFivePre}{11}
\newcommand{\tabFiveFiveOn}{1}
\newcommand{\tabFiveFivePost}{13}
\newcommand{\tabFiveFivePostPnr}{11}
\newcommand{\tabFiveFiveNone}{11}
\newcommand{\tabFiveFiveZero}{21/36}
\newcommand{\tabFiveFiveHOne}{81\%}
\newcommand{\tabOthreePre}{9}
\newcommand{\tabOthreeOn}{0}
\newcommand{\tabOthreePost}{8}
\newcommand{\tabOthreePostPnr}{17}
\newcommand{\tabOthreeNone}{13}
\newcommand{\tabOthreeZero}{26/34}
\newcommand{\tabOthreeHOne}{91\%}
\newcommand{\tabAstraPre}{24}
\newcommand{\tabAstraOn}{3}
\newcommand{\tabAstraPost}{12}
\newcommand{\tabAstraPostPnr}{7}
\newcommand{\tabAstraNone}{1}
\newcommand{\tabAstraZero}{30/46}
\newcommand{\tabAstraHOne}{78\%}
\newcommand{\tabLunaPre}{16}
\newcommand{\tabLunaOn}{1}
\newcommand{\tabLunaPost}{12}
\newcommand{\tabLunaPostPnr}{9}
\newcommand{\tabLunaNone}{9}
\newcommand{\tabLunaZero}{12/19}
\newcommand{\tabLunaHOne}{63\%}
\newcommand{\tabOpusPre}{3}
\newcommand{\tabOpusOn}{1}
\newcommand{\tabOpusPost}{8}
\newcommand{\tabOpusPostPnr}{7}
\newcommand{\tabOpusNone}{11}
\newcommand{\tabOpusZero}{9/12}
\newcommand{\tabOpusHOne}{89\%}
\newcommand{\tabHumanPre}{3}
\newcommand{\tabHumanOn}{2}
\newcommand{\tabHumanPost}{8}
\newcommand{\tabHumanPostPnr}{10}
\newcommand{\tabHumanNone}{7}
\newcommand{\tabHumanZero}{5/10}
\newcommand{\tabHumanHOne}{83\%}
\newcommand{\tabPoolPre}{67}
\newcommand{\tabPoolOn}{7}
\newcommand{\tabPoolPost}{45}
\newcommand{\tabPoolPostPnr}{43}
\newcommand{\tabPoolNone}{26}
\newcommand{\tabPoolZero}{107/162}
\newcommand{\tabPoolHOne}{81\%}
\newcommand{\catTstarEdit}{6\%}

\newcommand{\catLabelEdit}{31\%}

\newcommand{\catTstarRead}{81\%}
\newcommand{\catAllRead}{75\%}

\newcommand{\rqTwoN}{47}
\newcommand{\rqTwoCliff}{17}
\newcommand{\rqTwoErosion}{30}
\newcommand{\rqTwoCMedian}{0.39}
\newcommand{\rqTwoTstarFracMedian}{0.28}
\newcommand{\rqTwoBeforeEdit}{35/47}

\newcommand{\rqTwoPnrGapMedian}{6}

\newcommand{\bcN}{10}
\newcommand{\bcWithinThree}{6/10}
\newcommand{\bcSameShape}{8/10}

\newcommand{\vcSolOne}{0.16}

\newcommand{\vcFiveFiveOne}{0.09}

\newcommand{\vcFirstEditOne}{0.07}

\newcommand{\vcRandomOne}{0.04}

\newcommand{\rqThreeTau}{0.56}
\newcommand{\rqThreeP}{0.04}

\newcommand{\rqThreeConsensus}{22/47}
\newcommand{\opusN}{40}
\newcommand{\opusNFailed}{30}
\newcommand{\opusNDecoys}{10}
\newcommand{\opusNoMistakeFailed}{11/30}
\newcommand{\opusFlaggedDecoys}{0/10}

\newcommand{\caseMplTstar}{6}

\newcommand{\caseMplC}{0.95}
\newcommand{\caseMplShape}{cliff}
\newcommand{\caseMplVFive}{9/32}
\newcommand{\caseMplVSix}{0/32}

\newcommand{\caseMplVSeventeen}{1/16}

\newcommand{\caseMplVFortyNine}{0/16}
\newcommand{\caseReqTstar}{25}
\newcommand{\caseReqPnr}{29}

\newcommand{\caseReqShape}{cliff}

\newcommand{\caseReqVTwentyFour}{6/16}
\newcommand{\caseReqVTwentyFive}{2/24}

\newcommand{\caseReqVThirtyFive}{1/16}

\newcommand{\caseSymPnr}{28}
\newcommand{\caseSymC}{0.20}
\newcommand{\caseSymShape}{erosion}
\newcommand{\caseSymVFive}{6/8}

\newcommand{\caseSymVSeven}{20/32}

\newcommand{\caseSymVThirtyFour}{0/16}

\newcommand{\caseDjTstar}{11}

\newcommand{\caseDjShape}{erosion}

\newcommand{\caseDjVTen}{7/32}
\newcommand{\caseDjVEleven}{1/32}
\newcommand{\caseDjVSixteen}{0/24}

%% file: numbers_arm.tex
\newcommand{\armLengthMean}{85.1}
\newcommand{\armLengthMedian}{99}

\newcommand{\armPassRate}{0.34}

\newcommand{\armNBand}{44}

\newcommand{\armCapHitShare}{0.48}

\newcommand{\armNTasksAllPass}{14}
\newcommand{\armNTasksNonePass}{65}

\newcommand{\armNSelected}{60}

\newcommand{\armNTasksSelected}{33}
\newcommand{\armNDecoys}{10}

\newcommand{\armReplayRateAll}{72\%}

\newcommand{\armNBackfill}{21}
\newcommand{\armNStudy}{60}

\newcommand{\armSolNoMistakeFailed}{1/60}
\newcommand{\armSolFlaggedDecoys}{10/10}
\newcommand{\armSolPositionMedian}{0.19}
\newcommand{\armSolSelfExact}{15/20}

\newcommand{\armFiveFiveNoMistakeFailed}{6/60}
\newcommand{\armFiveFiveFlaggedDecoys}{3/10}
\newcommand{\armFiveFivePositionMedian}{0.28}
\newcommand{\armFiveFiveSelfExact}{10/20}

\newcommand{\armOthreeNoMistakeFailed}{15/60}
\newcommand{\armOthreeFlaggedDecoys}{3/10}
\newcommand{\armOthreePositionMedian}{0.41}
\newcommand{\armOthreeSelfExact}{5/20}

\newcommand{\armAstraNoMistakeFailed}{1/60}
\newcommand{\armAstraFlaggedDecoys}{10/10}
\newcommand{\armAstraPositionMedian}{0.15}
\newcommand{\armAstraSelfExact}{18/20}

\newcommand{\armLunaNoMistakeFailed}{5/60}
\newcommand{\armLunaFlaggedDecoys}{7/10}
\newcommand{\armLunaPositionMedian}{0.23}
\newcommand{\armLunaSelfExact}{10/20}

\newcommand{\armPairSolVsAstraExact}{57/70}

\newcommand{\armPrelimResolved}{60}
\newcommand{\armPrelimContinuations}{20,720}

\newcommand{\armNBandCapShare}{43/60}

\newcommand{\armRestartOracleGain}{+0.05}

\newcommand{\armRestartTstarGain}{+0.09}

\newcommand{\armRestartJudgeGainMin}{-0.13}
\newcommand{\armRestartJudgeGainMax}{-0.03}

\newcommand{\armRqOneTraj}{60}
\newcommand{\armRqOneLabels}{217}
\newcommand{\armRqOneHOne}{76\%}
\newcommand{\armRqOneHOneK}{166/217}

\newcommand{\armRqOneOn}{14/240}
\newcommand{\armRqOneZero}{63\%}

\newcommand{\armRqOneBig}{12/217}

\newcommand{\armCapHitOverall}{58\%}

\newcommand{\armCapHitLtwo}{69\%}

\newcommand{\armRqTwoCliff}{20}
\newcommand{\armRqTwoErosion}{40}

\newcommand{\armRqTwoTstarFracMedian}{0.27}

\newcommand{\armBcN}{10}

%% file: appendix_tables.tex
\section{Frozen estimator parameters}
\label{app:params}

Table~\ref{tab:params} lists every estimator and schedule parameter. They were fixed before the first base run and are the same for base runs and continuations; the schedule of Table~\ref{tab:alg} reads them and nothing else. The values trade cost against resolution: $K_{\text{screen}}=8$ makes a screening point cheap but noisy (a drop of $0.25$ is one standard error at $V\approx0.5$), $K_{\text{confirm}}=32$ brings the standard error of a confirmed drop to about $0.08$, $K_{\text{pnr}}=24$ with $\epsilon=0.15$ certifies a point of no return from $0$ successes (one-sided Wilson upper bound $0.10$) or $1$ success ($0.19$, not certified), and $K_{\text{label}}=16$ is the smallest sample at which a drop of $0.3$ at a label is distinguishable from none. Conclusion: the curves are coarse where nothing happens and precise where the schedule found a drop or a label, by design; no quantity in the paper is read from a point the schedule did not raise to the $K$ its definition requires.

\begin{table}[htbp]
\centering
{\footnotesize\setlength{\tabcolsep}{3pt}
\begin{tabular}[t]{p{0.56\columnwidth}p{0.36\columnwidth}}
\toprule
Parameter & Value \\
\midrule
Budget $B$ (steps) & 100 \\
Base runs per task & 8 \\
Selection band (successes of 8) & 2--6 \\
Trajectories per task, at most & 3 \\
Decoys / recovery pool & 10 / 20 \\
Anchors & every 5 to step 50, every 10 after, $T-1$ \\
$K_{\text{screen}}$, $\delta_{\text{screen}}$ & 8, 0.25 \\
Bisected intervals, at most & 3 \\
$K_{\text{confirm}}$, second-drop minimum & 32, 0.25 \\
$K_{\text{label}}$ (raised to, when drop $\ge$) & 16 (32, 0.2) \\
Labels probed per trajectory, at most & 4 \\
$K_{\text{pnr}}$, $\epsilon$ & 24, 0.15 \\
Cliff threshold on $C$ & 0.5 \\
$\alpha$ & 0.05 \\
Wilson $z$ (two-sided, one-sided) & 1.960, 1.645 \\
Batch size (trajectories); pipeline threshold & 10; 3 \\
Sanity / retest / budget-control trajectories & 5 / 10 / 10 \\
Bootstrap resamples & 10000 \\
\bottomrule
\end{tabular}}
\caption{Estimator and schedule parameters (\texttt{configs/thresholds.yaml}).}
\label{tab:params}
\end{table}\vspace{-5pt}

\FloatBarrier
\section{The base-run pool}
\label{app:pool}

Table~\ref{tab:pool} summarizes the base runs of the main policy (left) and the distribution of tasks by number of successes (right). The distribution is bimodal: 77 tasks succeed in all eight runs and 28 in none, so only 21 tasks fall in the band of 2--6 successes where a failed run is informative; pass@8 is 0.51 against pass@1 0.68. Two consequences follow. The study set is bounded by the band (21 tasks here) because failures outside it are uninformative: on a doomed task ($\le1$ success) $V(h_0)$ is already near zero and no step can lose much, and on a saturated task a failed run is a rare draw whose continuations mostly succeed, so the curve is flat near one until late. The band is therefore where cause and symptom can be told apart, and every trajectory in the study comes from it. Second, the base runs also give $R(x,b)$ (Definition~5): with eight runs per task, $\hat R$ is a step function with increments of $1/8$, which is the resolution of the offline restart evaluation and one reason for the online check (Appendix~\ref{app:online}). 8\% of base runs hit the step cap and 105 submitted an empty patch (counted as failures without a harness run). Conclusion: the informative failures are those of the 21 band tasks of 150, so a study of where failures are caused must select by pass rate first; the 129 tasks outside the band contribute to $R(x,b)$ and to the decoys' calibration but not to the curves.

\begin{table}[htbp]
\centering
{\footnotesize\setlength{\tabcolsep}{3pt}
\begin{tabular}[t]{lr}
\toprule
Quantity & Value \\
\midrule
Tasks, runs & 150, 1200 \\
pass@1 & 0.683 \\
pass@2 / @4 / @8 & 0.63 / 0.58 / 0.51 \\
Doomed (0--1 of 8) & 0.24 \\
Saturated (7--8 of 8) & 0.62 \\
Band (2--6 of 8) & 0.14 (21 tasks) \\
Cap hits & 8.2\% \\
Empty patches & 105 \\
Evaluation errors & 5 \\
\bottomrule
\end{tabular}}\hfill
{\footnotesize\setlength{\tabcolsep}{3pt}
\begin{tabular}[t]{cr}
\toprule
Successes & Tasks \\
\midrule
0 & 28 \\
1 & 8 \\
2 & 6 \\
3 & 2 \\
4 & 4 \\
5 & 4 \\
6 & 5 \\
7 & 16 \\
8 & 77 \\
\bottomrule
\end{tabular}}
\caption{Base-run pool of the main policy (left) and tasks by number of successes out of 8 (right).}
\label{tab:pool}
\end{table}\vspace{-5pt}

\FloatBarrier
\section{All trajectories}
\label{app:trajectories}

Table~\ref{tab:alltraj} gives, for each of the 47 trajectories of the frozen set, the quantities the curve yields: Stage~2 stratum (L0 and L1: the base run submitted a patch, in the lower or upper length tercile; L2: the base run hit the step cap), length $T$, value at $h_0$ (isotonic fit), shape and concentration $C$, the decisive step $t^*$ with its interval, the certified point of no return, the first repository edit, the first explicit failure signal, and the number of continuations. Both endpoints of the interval are estimated at $K\ge32$ in 13 trajectories ($\dagger$). Tables~\ref{tab:alllabels} and \ref{tab:alllabels-2} list every source's first-mistake label next to $t^*$ and $t_{\text{pnr}}$; ``none'' is a ``no mistake'' answer and $\cdot$ means the source did not label that trajectory (GPT-6.1 Sol, GPT-6 Astra, GPT-5.5 and OpenAI o3 are the main judges, GPT-6 Luna the contrast model, Claude Opus 5.5 the annotator-protocol model on its subset). Reading the table: the set has 17 L0, 12 L1 and 18 L2 trajectories, 40 to 100 steps long (median 75); the fitted value at $h_0$ runs from 0.05 to 1.00 (median 0.50), around a base pass rate that lies in the band $[0.25,0.75]$ by construction (the fitted value differs from it by the sampling noise of eight continuations and by the monotone fit). The decisive step sits at a median of 28\% of the trajectory; 4 trajectories lose the task within the first five steps, and none only at the last step: every failure is decided before the agent stops. The point of no return follows $t^*$ by a median of 6 steps (upper quartile 30); where the gap is long the curve is an erosion whose value leaks slowly after the largest drop. 40 of 47 trajectories show an explicit failure signal after the first edit; on the others the agent fails silently, usually by submitting a patch that does not resolve the issue. In Tables~\ref{tab:alllabels} and \ref{tab:alllabels-2} the ``no mistake'' answers cluster by lineage: GPT-6.1 Sol 1, GPT-6 Astra 1, GPT-5.5 11, OpenAI o3 13, GPT-6 Luna (contrast) 9, Claude Opus 5.5 (annotator) 11, Human annotator 7. Conclusion: the table is the paper's raw material; every count in \S\ref{sec:results} can be recomputed from it, and a reader who wants to check a claim about one trajectory (for instance that the judges' labels fall after $t_{\text{pnr}}$) can do so by comparing the label columns with the $t^*$ and $t_{\text{pnr}}$ columns.

\begin{table*}[tp]
\centering
{\scriptsize\setlength{\tabcolsep}{3pt}
\begin{tabular}[t]{lccccccccccc}
\toprule
Trajectory & Str. & $T$ & $\hat V(h_0)$ & Shape & $C$ & $t^*$ & $(a,b]$ & $t_{\text{pnr}}$ & $t_{\text{edit}}$ & $t_{\text{sym}}$ & $n$ \\
\midrule
django-11555 s2 & L1 & 88 & 0.48 & erosion & 0.30 & 35 & (34,35] & 71 & 44 & 45 & 376 \\
django-11555 s5 & L2 & 100 & 0.40 & erosion & 0.39 & 86 & (85,90] & 93 & 52 & 54 & 440 \\
django-11555 s7 & L1 & 99 & 0.53 & erosion & 0.23 & 6 & (5,10] & 82 & 40 & 41 & 392 \\
django-12325 s1 & L1 & 67 & 0.07 & erosion & 0.40 & 11 & (10,11] & 15 & 30 & 39 & 248 \\
django-12325 s2 & L1 & 90 & 0.06 & cliff & 0.60 & 2 & (1,2] & 2 & 38 & 46 & 296 \\
django-12325 s6 & L1 & 67 & 0.05 & cliff & 1.00 & 16 & (15,20] & 0 & 31 & 36 & 200 \\
django-13807 s2 & L1 & 71 & 0.44 & cliff & 0.71 & 1 & (0,5]$^\dagger$ & 15 & 18 & 23 & 248 \\
django-13807 s3 & L0 & 41 & 0.50 & cliff & 0.69 & 13 & (12,13]$^\dagger$ & 13 & 13 & 19 & 232 \\
django-13807 s7 & L1 & 68 & 0.25 & erosion & 0.40 & 6 & (5,7] & 18 & 16 & 46 & 264 \\
django-15128 s4 & L2 & 100 & 0.50 & erosion & 0.25 & 17 & (16,17] & 63 & 65 & 78 & 424 \\
django-15128 s5 & L1 & 84 & 0.50 & erosion & 0.38 & 9 & (8,9] & 40 & 38 & 42 & 360 \\
django-15128 s6 & L1 & 92 & 0.62 & erosion & 0.25 & 51 & (50,60] & 81 & 53 & -- & 392 \\
django-16100 s1 & L0 & 63 & 0.46 & erosion & 0.32 & 39 & (38,39] & 60 & 21 & 34 & 416 \\
django-16100 s2 & L0 & 58 & 0.36 & cliff & 1.00 & 50 & (49,50] & 50 & 24 & 34 & 344 \\
django-16100 s6 & L1 & 68 & 0.32 & erosion & 0.39 & 48 & (47,48] & 48 & 21 & 28 & 392 \\
django-16938 s2 & L2 & 100 & 0.50 & erosion & 0.35 & 11 & (10,11]$^\dagger$ & 50 & 35 & 39 & 320 \\
django-16938 s4 & L1 & 65 & 0.56 & erosion & 0.33 & 11 & (10,11]$^\dagger$ & 11 & 19 & 20 & 248 \\
django-16938 s5 & L2 & 100 & 0.50 & cliff & 0.71 & 95 & (94,96] & 97 & 68 & -- & 376 \\
matplotlib-22865 s2 & L0 & 54 & 0.39 & erosion & 0.32 & 6 & (5,10] & 30 & 32 & -- & 264 \\
matplotlib-22865 s3 & L0 & 40 & 0.50 & erosion & 0.39 & 19 & (18,19] & 22 & 25 & -- & 256 \\
matplotlib-22865 s7 & L0 & 51 & 0.31 & cliff & 0.60 & 12 & (11,12] & 20 & 31 & 35 & 248 \\
matplotlib-25960 s0 & L0 & 58 & 0.12 & cliff & 0.50 & 22 & (21,25] & 0 & 29 & 38 & 248 \\
matplotlib-25960 s3 & L0 & 64 & 0.12 & erosion & 0.48 & 11 & (10,15] & 25 & 21 & 29 & 264 \\
matplotlib-25960 s5 & L0 & 57 & 0.18 & cliff & 0.95 & 6 & (5,6]$^\dagger$ & 6 & 23 & 31 & 256 \\
\bottomrule
\end{tabular}}
\caption{Curve quantities of every trajectory of the frozen set (1 of 2); $\dagger$: both endpoints of the interval at $K\ge32$.}
\label{tab:alltraj}
\end{table*}\vspace{-5pt}
\begin{table*}[tp]
\centering
{\scriptsize\setlength{\tabcolsep}{3pt}
\begin{tabular}[t]{lccccccccccc}
\toprule
Trajectory & Str. & $T$ & $\hat V(h_0)$ & Shape & $C$ & $t^*$ & $(a,b]$ & $t_{\text{pnr}}$ & $t_{\text{edit}}$ & $t_{\text{sym}}$ & $n$ \\
\midrule
seaborn-3069 s2 & L2 & 100 & 0.50 & cliff & 0.53 & 14 & (13,14]$^\dagger$ & 91 & 44 & 56 & 368 \\
seaborn-3069 s3 & L2 & 100 & 0.43 & erosion & 0.42 & 2 & (1,2] & 30 & 59 & 82 & 384 \\
seaborn-3069 s5 & L2 & 100 & 0.54 & erosion & 0.39 & 62 & (61,62] & 62 & 29 & 30 & 344 \\
requests-6028 s0 & L0 & 64 & 0.12 & cliff & 0.50 & 26 & (25,26] & 26 & 38 & -- & 248 \\
requests-6028 s1 & L0 & 57 & 0.24 & erosion & 0.35 & 13 & (12,13]$^\dagger$ & 15 & 23 & 39 & 240 \\
requests-6028 s4 & L0 & 61 & 0.29 & cliff & 0.65 & 25 & (24,25] & 29 & 35 & -- & 360 \\
scikit-learn-10908 s3 & L0 & 47 & 1.00 & cliff & 0.65 & 19 & (18,19]$^\dagger$ & 19 & 22 & 27 & 256 \\
sphinx-8265 s0 & L0 & 64 & 0.50 & cliff & 0.65 & 16 & (15,16]$^\dagger$ & 22 & 23 & 38 & 280 \\
sphinx-8265 s5 & L1 & 75 & 1.00 & erosion & 0.25 & 18 & (17,18] & 21 & 26 & 41 & 320 \\
sphinx-8593 s2 & L2 & 100 & 0.88 & erosion & 0.20 & 51 & (50,51] & 99 & 48 & 90 & 424 \\
sphinx-8593 s6 & L2 & 100 & 0.65 & erosion & 0.33 & 98 & (97,98] & 76 & 36 & 59 & 352 \\
sphinx-9591 s2 & L2 & 100 & 0.50 & cliff & 0.50 & 48 & (47,48]$^\dagger$ & 67 & 50 & 52 & 424 \\
sphinx-9591 s4 & L2 & 100 & 0.38 & erosion & 0.33 & 28 & (27,28] & 70 & 48 & 58 & 448 \\
sphinx-9591 s5 & L2 & 100 & 0.54 & erosion & 0.31 & 8 & (7,10] & 47 & 76 & 99 & 352 \\
sympy-13877 s1 & L0 & 58 & 0.62 & erosion & 0.20 & 6 & (5,6] & 28 & 34 & -- & 280 \\
sympy-13877 s5 & L0 & 44 & 0.79 & erosion & 0.35 & 32 & (31,32]$^\dagger$ & 32 & 20 & 36 & 328 \\
sympy-15875 s1 & L0 & 57 & 0.57 & erosion & 0.49 & 15 & (14,15]$^\dagger$ & 16 & 17 & 42 & 304 \\
sympy-15875 s4 & L2 & 100 & 0.54 & cliff & 0.50 & 35 & (34,35] & 88 & 45 & 60 & 432 \\
sympy-19040 s0 & L2 & 100 & 0.38 & erosion & 0.33 & 1 & (0,2] & 4 & 74 & 84 & 296 \\
sympy-19040 s3 & L2 & 100 & 0.60 & cliff & 0.53 & 42 & (41,42]$^\dagger$ & 52 & 46 & 89 & 344 \\
sympy-19040 s7 & L2 & 100 & 0.15 & erosion & 0.45 & 66 & (65,66] & 31 & 54 & 58 & 368 \\
sympy-19783 s1 & L2 & 100 & 0.82 & erosion & 0.30 & 97 & (96,97] & 97 & 41 & 82 & 496 \\
sympy-19783 s5 & L2 & 100 & 0.75 & erosion & 0.21 & 46 & (45,50] & 90 & 29 & 89 & 424 \\
\bottomrule
\end{tabular}}
\caption{Curve quantities of every trajectory of the frozen set (2 of 2); $\dagger$: both endpoints of the interval at $K\ge32$.}
\label{tab:alltraj-2}
\end{table*}\vspace{-5pt}
\begin{table*}[tp]
\centering
{\scriptsize\setlength{\tabcolsep}{3pt}
\begin{tabular}[t]{lccccccccc}
\toprule
 &  &  & GPT-6.1 & GPT-6 & GPT-5.5 & OpenAI & GPT-6 & Claude & Human \\ Trajectory & $t^*$ & $t_{\text{pnr}}$ & Sol & Astra &  & o3 & Luna & Opus 5.5 &  \\
\midrule
django-11555 s2 & 35 & 71 & 13 & 13 & 56 & 44 & 44 & $\cdot$ & $\cdot$ \\
django-11555 s5 & 86 & 93 & 12 & 12 & 58 & none & 52 & none & none \\
django-11555 s7 & 6 & 82 & 13 & 13 & 82 & 82 & 65 & $\cdot$ & $\cdot$ \\
django-12325 s1 & 11 & 15 & 27 & 27 & none & 28 & none & none & none \\
django-12325 s2 & 2 & 2 & 38 & 38 & none & 22 & none & none & 38 \\
django-12325 s6 & 16 & 0 & 28 & 28 & none & none & none & $\cdot$ & $\cdot$ \\
django-13807 s2 & 1 & 15 & 10 & 10 & none & 18 & none & $\cdot$ & $\cdot$ \\
django-13807 s3 & 13 & 13 & 10 & 10 & none & none & none & none & 13 \\
django-13807 s7 & 6 & 18 & 8 & 8 & none & none & none & none & 11 \\
django-15128 s4 & 17 & 63 & 11 & 11 & 40 & 65 & 39 & $\cdot$ & $\cdot$ \\
django-15128 s5 & 9 & 40 & 10 & 10 & 30 & 76 & 38 & 36 & 36 \\
django-15128 s6 & 51 & 81 & 10 & 10 & 65 & 10 & 15 & 53 & 53 \\
django-16100 s1 & 39 & 60 & 17 & 17 & 17 & 21 & 19 & $\cdot$ & $\cdot$ \\
django-16100 s2 & 50 & 50 & 2 & 2 & none & 24 & none & $\cdot$ & $\cdot$ \\
django-16100 s6 & 48 & 48 & 20 & 20 & 21 & 21 & 40 & $\cdot$ & $\cdot$ \\
django-16938 s2 & 11 & 50 & 11 & 11 & 99 & none & 11 & $\cdot$ & $\cdot$ \\
django-16938 s4 & 11 & 11 & 11 & 11 & 16 & none & none & 17 & 17 \\
django-16938 s5 & 95 & 97 & 8 & 8 & none & none & 11 & none & none \\
matplotlib-22865 s2 & 6 & 30 & 16 & 16 & 28 & none & 16 & 30 & 30 \\
matplotlib-22865 s3 & 19 & 22 & 20 & 20 & 20 & 18 & 18 & $\cdot$ & $\cdot$ \\
matplotlib-22865 s7 & 12 & 20 & 21 & 21 & 23 & 21 & 21 & 23 & 23 \\
matplotlib-25960 s0 & 22 & 0 & 29 & 17 & 21 & 29 & 28 & 17 & 28 \\
matplotlib-25960 s3 & 11 & 25 & 16 & 21 & 16 & 21 & 21 & 21 & 16 \\
matplotlib-25960 s5 & 6 & 6 & 17 & 17 & 17 & 49 & 32 & 17 & 20 \\
\bottomrule
\end{tabular}}
\caption{First-mistake labels of every source next to $t^*$ and $t_{\text{pnr}}$, first half of the set (``none'' = no mistake; $\cdot$ = not labelled by that source).}
\label{tab:alllabels}
\end{table*}\vspace{-5pt}
\begin{table*}[tp]
\centering
{\scriptsize\setlength{\tabcolsep}{3pt}
\begin{tabular}[t]{lccccccccc}
\toprule
 &  &  & GPT-6.1 & GPT-6 & GPT-5.5 & OpenAI & GPT-6 & Claude & Human \\ Trajectory & $t^*$ & $t_{\text{pnr}}$ & Sol & Astra &  & o3 & Luna & Opus 5.5 &  \\
\midrule
seaborn-3069 s2 & 14 & 91 & 14 & 14 & 14 & 45 & 45 & $\cdot$ & $\cdot$ \\
seaborn-3069 s3 & 2 & 30 & 16 & 16 & 96 & 59 & 59 & 96 & 96 \\
seaborn-3069 s5 & 62 & 62 & 16 & 16 & 29 & 29 & 56 & $\cdot$ & $\cdot$ \\
requests-6028 s0 & 26 & 26 & 18 & 18 & 18 & none & 18 & $\cdot$ & $\cdot$ \\
requests-6028 s1 & 13 & 15 & 14 & 14 & 14 & 21 & 14 & 15 & 14 \\
requests-6028 s4 & 25 & 29 & 24 & 24 & 35 & 35 & 35 & 30 & 29 \\
scikit-learn-10908 s3 & 19 & 19 & none & none & none & none & none & $\cdot$ & $\cdot$ \\
sphinx-8265 s0 & 16 & 22 & 19 & 19 & 19 & none & 21 & 19 & 19 \\
sphinx-8265 s5 & 18 & 21 & 19 & 19 & 26 & 73 & 26 & 19 & 19 \\
sphinx-8593 s2 & 51 & 99 & 48 & 48 & 99 & 48 & 65 & $\cdot$ & $\cdot$ \\
sphinx-8593 s6 & 98 & 76 & 76 & 76 & none & none & 76 & none & 76 \\
sphinx-9591 s2 & 48 & 67 & 7 & 7 & 83 & 83 & 10 & $\cdot$ & $\cdot$ \\
sphinx-9591 s4 & 28 & 70 & 8 & 8 & 48 & 48 & 47 & none & none \\
sphinx-9591 s5 & 8 & 47 & 7 & 7 & none & 76 & 75 & none & none \\
sympy-13877 s1 & 6 & 28 & 7 & 7 & 34 & 34 & 34 & 24 & 30 \\
sympy-13877 s5 & 32 & 32 & 16 & 16 & 16 & 20 & 20 & 16 & 15 \\
sympy-15875 s1 & 15 & 16 & 17 & 17 & 27 & 27 & 12 & none & 23 \\
sympy-15875 s4 & 35 & 88 & 12 & 12 & 55 & 58 & 44 & 58 & none \\
sympy-19040 s0 & 1 & 4 & 62 & 62 & 62 & none & 62 & 93 & 95 \\
sympy-19040 s3 & 42 & 52 & 11 & 11 & 35 & 46 & 15 & 35 & 35 \\
sympy-19040 s7 & 66 & 31 & 24 & 24 & 24 & 91 & 24 & 83 & 89 \\
sympy-19783 s1 & 97 & 97 & 24 & 22 & 39 & 75 & 39 & none & none \\
sympy-19783 s5 & 46 & 90 & 26 & 26 & 36 & 76 & 26 & $\cdot$ & $\cdot$ \\
\bottomrule
\end{tabular}}
\caption{First-mistake labels of every source next to $t^*$ and $t_{\text{pnr}}$, second half of the set.}
\label{tab:alllabels-2}
\end{table*}\vspace{-5pt}

\FloatBarrier
\section{Schedule statistics}
\label{app:schedule}

Table~\ref{tab:stages} breaks the 15,576 continuations of the frozen curves down by schedule stage. A curve has 24.1 points on average (range 15--35) and costs 331 continuations (median 344, range 200--496); screening and the label probes dominate, confirmation and the point of no return add the rest. Control continuations (Appendix~\ref{app:controls}) are not in the curves. The label probes are the price of the direct measurement in \S\ref{sec:rq1}: every distinct labeled step costs two points at $K\ge16$, and because the four judges rarely agree, most trajectories carry three or four distinct labels. The confirmation and PNR stages are cheaper than their $K$ suggests because they reuse the screening and bisection continuations at the same histories (continuations are cumulative per history). The cost of a curve is set by its length and by how many drops screening finds, not by the outcome: cliffs and erosions cost about the same. Conclusion: a curve without label probes costs about 277 continuations, and the probes, which exist only to test the judges, add 55; a controller that needs the restart point alone would spend on screening and the point of no return and skip confirmation (\S\ref{sec:cheap}).

\begin{table}[htbp]
\centering
{\footnotesize\setlength{\tabcolsep}{3pt}
\begin{tabular}[t]{lrrr}
\toprule
Stage & Cont. & Share & Per traj. \\
\midrule
screen & 5,240 & 34\% & 111 \\
confirm & 3,384 & 22\% & 72 \\
pnr & 2,728 & 18\% & 58 \\
label & 2,576 & 17\% & 55 \\
bisect & 1,648 & 11\% & 35 \\
\midrule all & 15,576 & 100\% & 331 \\
\bottomrule
\end{tabular}}
\caption{Continuations by schedule stage over the frozen set.}
\label{tab:stages}
\end{table}\vspace{-5pt}

\FloatBarrier
\section{Replay}
\label{app:replay}

Table~\ref{tab:replay} reports replay (Stage~3) by role and by stratum (left) and the mismatches it found (right: by type, by the kind of command at the first mismatch), and Table~\ref{tab:replay-levels} the comparison level at which observations matched. 70 of 94 trajectories replay exactly; the L1-submit stratum replays worst (60\%). Non-replayable trajectories print genuinely non-deterministic values; 1,036 anchor snapshots were built, and 6 back-fills in 1 round(s) left 47 of 64 selected trajectories for Stage~4. The first mismatch occurs at step 17 on median (quartiles 9, 38; maximum 52), and almost always in an observation rather than in the filesystem fingerprint: the commands involved print values that cannot be masked by context, such as random tokens, savepoint names, \texttt{git stash} output or the ordering of a hash-based structure. We drop rather than patch such trajectories, because a continuation from a state that differs from the recorded one would measure a different history. Most matches are exact; the sorted-lines and character-multiset levels absorb reorderings of directory listings and test output whose content is identical. The L2 stratum replays best, plausibly because capped runs rarely reach the commands that print non-deterministic values. Conclusion: exact replay is attainable for about three quarters of real SWE-bench trajectories with a normalization fixed in advance, the failures are explainable and are reported rather than repaired, and because back-fills are drawn from the same strata in a fixed order, the study set remains a stratified sample of the band's failures.

\begin{table}[htbp]
\centering
{\scriptsize\setlength{\tabcolsep}{3pt}
\begin{tabular}[t]{lrrr}
\toprule
Group & $n$ & Repl. & Rate \\
\midrule
selected & 64 & 47 & 73\% \\
decoy & 10 & 8 & 80\% \\
recovery & 20 & 15 & 75\% \\
\midrule L0-submit & 24 & 17 & 71\% \\
\midrule L1-submit & 20 & 12 & 60\% \\
\midrule L2-cap & 20 & 18 & 90\% \\
\midrule all & 94 & 70 & 74\% \\
\bottomrule
\end{tabular}}\hfill
{\scriptsize\setlength{\tabcolsep}{3pt}
\begin{tabular}[t]{p{0.30\columnwidth}r}
\toprule
Mismatch & $n$ \\
\midrule
observation & 22 \\
returncode & 1 \\
docker\_diff\_sha256 & 1 \\
\midrule \\
python script & 7 \\
test run & 7 \\
git stash & 5 \\
write file / heredoc & 5 \\
\bottomrule
\end{tabular}}
\caption{Replay by role and stratum (left); mismatch types and command kinds at the first mismatch (right).}
\label{tab:replay}
\end{table}\vspace{-5pt}
\begin{table}[htbp]
\centering
{\scriptsize\setlength{\tabcolsep}{3pt}
\begin{tabular}[t]{ccc}
\toprule
sorted\_lines & exact & char\_multiset \\
\midrule
32 & 28 & 10 \\
\bottomrule
\end{tabular}}
\caption{Comparison level at which the observations of replayable trajectories matched (steps).}
\label{tab:replay-levels}
\end{table}\vspace{-5pt}

\FloatBarrier
\section{Judge agreement in detail}
\label{app:judges}

Table~\ref{tab:judge-matrix} gives the pairwise agreement of every judge pair on the 47 failed trajectories as exact / within 1 / within 3 steps, and Table~\ref{tab:judge-detail} each judge's number of labels, answers that did not follow the required format (a single step index, or a sentinel for ``no mistake''), ``no mistake'' answers on failed runs, flagged decoys, label-position quartiles ($m/T$), self-consistency over three passes on 20 trajectories (exact / within 1 / within 3) and API spend. A consensus of the four main judges is rare: unanimous on the exact step in 3 of 57 trajectories, at least two within one step in 56. Total judge spend US\$53. The matrix has a block structure: the two GPT-6 judges agree with each other far more often than either agrees with GPT-5.5 or OpenAI o3, and GPT-5.5, OpenAI o3 and the small contrast model form a looser second group that answers ``no mistake'' more often and labels later in the trajectory (position quartiles in Table~\ref{tab:judge-detail}). Self-consistency separates the groups as well: the GPT-6 judges return the same answer in three passes on most trajectories, GPT-5.5 and OpenAI o3 on a minority, so part of the cross-lineage disagreement is within-judge sampling noise rather than a stable difference of opinion. The spend column shows the cost of one pass over the study set per judge; the three self-consistency passes on the 20-trajectory subset are included. Conclusion: the judges do not provide four independent opinions but two lineages with different biases, one that labels early and flags decoys, one that labels late and abstains; any ground truth built by aggregating them inherits the balance between the two, and inter-judge agreement overstates reliability unless computed across lineages. The names used throughout the paper map to the following API model identifiers, fixed for the whole study: GPT-6.1 Sol: \texttt{gpt-6.1-sol}, GPT-5.5: \texttt{gpt-5.5-2026-04-23}, OpenAI o3: \texttt{o3-2025-04-16}, GPT-6 Astra: \texttt{gpt-6-astra}, GPT-6 Luna: \texttt{gpt-6-luna}; the annotator-protocol model is Claude Opus 5.5 (\texttt{claude-opus-5-5}). The two policies are \texttt{Qwen/Qwen3.6-27B-FP8} (main) and \texttt{Qwen/Qwen3.5-9B} (second policy), at the revisions recorded in the released manifest.

\begin{table}[htbp]
\centering
{\scriptsize\setlength{\tabcolsep}{3pt}
\begin{tabular}[t]{lccccc}
\toprule
 & GPT-6.1 & GPT-5.5 & OpenAI & GPT-6 & GPT-6 \\  & Sol &  & o3 & Astra & Luna \\
\midrule
GPT-6.1 Sol & -- & 12/13/14 & 5/7/8 & 44/44/45 & 10/11/16 \\
GPT-5.5 & 12/13/14 & -- & 14/14/17 & 11/12/13 & 16/18/22 \\
OpenAI o3 & 5/7/8 & 14/14/17 & -- & 5/7/8 & 14/17/18 \\
GPT-6 Astra & 44/44/45 & 11/12/13 & 5/7/8 & -- & 11/11/16 \\
GPT-6 Luna & 10/11/16 & 16/18/22 & 14/17/18 & 11/11/16 & -- \\
\bottomrule
\end{tabular}}
\caption{Pairwise agreement on the 47 failed trajectories: exact / within 1 / within 3 steps.}
\label{tab:judge-matrix}
\end{table}\vspace{-5pt}
\begin{table}[htbp]
\centering
{\scriptsize\setlength{\tabcolsep}{3pt}
\begin{tabular}[t]{lccccc}
\toprule
 & GPT-6.1 & GPT-5.5 & OpenAI & GPT-6 & GPT-6 \\  & Sol &  & o3 & Astra & Luna \\
\midrule
labels (pass 1) & 57 & 57 & 57 & 57 & 57 \\
unparsed answers & 0 & 0 & 0 & 0 & 0 \\
``no mistake'' (failed) & 1/47 & 11/47 & 13/47 & 1/47 & 9/47 \\
flagged decoys & 8/10 & 4/10 & 3/10 & 8/10 & 3/10 \\
position $m/T$, Q1 & 0.11 & 0.29 & 0.39 & 0.11 & 0.27 \\
position $m/T$, median & 0.23 & 0.39 & 0.47 & 0.23 & 0.44 \\
position $m/T$, Q3 & 0.30 & 0.58 & 0.75 & 0.29 & 0.56 \\
self-consistency, exact & 15/20 & 6/20 & 5/20 & 19/20 & 8/20 \\
self-consistency, $\pm1$ & 15/20 & 7/20 & 5/20 & 19/20 & 9/20 \\
self-consistency, $\pm3$ & 16/20 & 7/20 & 5/20 & 19/20 & 10/20 \\
spend (US\$) & 4.1 & 21.5 & 6.0 & 21.1 & 0.3 \\
\bottomrule
\end{tabular}}
\caption{Judges: labels, answers not in the required format, ``no mistake'' on failed runs, flagged decoys, label-position quartiles, self-consistency over three passes on 20 trajectories, spend.}
\label{tab:judge-detail}
\end{table}\vspace{-5pt}

\FloatBarrier
\section{Controls}
\label{app:controls}

Three controls run on the first trajectories of the processing order. Table~\ref{tab:sanity}: on the first 5 trajectories the continuations from the empty history are compared with the task's base runs (Fisher's exact test, two-sided). Table~\ref{tab:retest}: on the first 10 trajectories, screening is repeated at three anchors with fresh seeds; the median absolute gap at $K=8$ is 0.00 (maximum 0.38). Table~\ref{tab:controls}: on the first 10 trajectories the screening anchors are run again with the full budget $B$ instead of $B-t$; the decisive step from screening agrees within three steps in 6 of 10 (sh.: shape, cli/ero; final: the study's $t^*$). The sanity rows show that resuming the agent from the empty history reproduces the base pass rate (no row is rejected at $\alpha=0.05$), so the continuation machinery itself does not shift the success probability. The retest rows show that at $K=8$ two independent screenings of the same history agree exactly in 21 of 30 anchors and differ by more than $0.25$ in 1; a gap of $3/8$ between two draws of eight is within binomial noise at $K=8$, which is why the schedule confirms at $K=32$ before naming a decisive step. The budget-control rows show the one systematic effect of the protocol: with the full budget the decisive step never moves earlier and moves later in 4 of 10 trajectories, in 2 cases to the last step, where a run that was still viable under $B$ runs out of steps under $B-t$. We therefore read a late $t^*$ under the protocol's $B-t$ as ``lost under the remaining budget'', which is the quantity a controller faces, and not as a judgment about the agent's reasoning alone. Conclusion: the method reproduces the base pass rate, repeats itself within binomial noise, and has one known bias, a late decisive step on long runs, whose direction and size are measured here and whose effect on the main claims is bounded by the analysis without the L2 stratum (Appendix~\ref{app:rq1}).

\begin{table}[htbp]
\centering
{\scriptsize\setlength{\tabcolsep}{3pt}
\begin{tabular}[t]{lrr}
\toprule
Trajectory & At $h_0$ & $p(x)$ \\
\midrule
django-15128 s5 & 4/8 & 0.50 \\
django-16938 s4 & 4/8 & 0.50 \\
requests-6028 s1 & 0/8 & 0.25 \\
sympy-15875 s4 & 4/8 & 0.75 \\
sympy-19040 s3 & 2/8 & 0.25 \\
\bottomrule
\end{tabular}}
\caption{$t=0$ sanity: successes of the continuations from the empty history against the task's base pass rate.}
\label{tab:sanity}
\end{table}\vspace{-5pt}
\begin{table}[htbp]
\centering
{\scriptsize\setlength{\tabcolsep}{3pt}
\begin{tabular}[t]{lrrrr}
\toprule
Trajectory & $t$ & $\hat V_1$ & $\hat V_2$ & $|\Delta|$ \\
\midrule
django-15128 s5 & 0 & 0.50 & 0.50 & 0.00 \\
django-15128 s5 & 35 & 0.25 & 0.50 & 0.25 \\
django-15128 s5 & 83 & 0.00 & 0.00 & 0.00 \\
django-16938 s4 & 0 & 0.50 & 0.50 & 0.00 \\
django-16938 s4 & 30 & 0.00 & 0.00 & 0.00 \\
django-16938 s4 & 64 & 0.00 & 0.00 & 0.00 \\
matplotlib-25960 s5 & 0 & 0.00 & 0.00 & 0.00 \\
matplotlib-25960 s5 & 30 & 0.00 & 0.00 & 0.00 \\
matplotlib-25960 s5 & 56 & 0.00 & 0.00 & 0.00 \\
requests-6028 s1 & 0 & 0.00 & 0.12 & 0.12 \\
requests-6028 s1 & 30 & 0.00 & 0.00 & 0.00 \\
requests-6028 s1 & 56 & 0.00 & 0.00 & 0.00 \\
requests-6028 s4 & 0 & 0.12 & 0.12 & 0.00 \\
requests-6028 s4 & 30 & 0.00 & 0.00 & 0.00 \\
requests-6028 s4 & 60 & 0.00 & 0.00 & 0.00 \\
sympy-13877 s1 & 0 & 0.50 & 0.75 & 0.25 \\
sympy-13877 s1 & 30 & 0.00 & 0.12 & 0.12 \\
sympy-13877 s1 & 57 & 0.00 & 0.00 & 0.00 \\
sympy-13877 s5 & 0 & 0.62 & 0.38 & 0.25 \\
sympy-13877 s5 & 25 & 0.25 & 0.25 & 0.00 \\
sympy-13877 s5 & 43 & 0.00 & 0.00 & 0.00 \\
sympy-15875 s4 & 0 & 0.50 & 0.38 & 0.12 \\
sympy-15875 s4 & 40 & 0.25 & 0.38 & 0.12 \\
sympy-15875 s4 & 99 & 0.00 & 0.00 & 0.00 \\
sympy-19040 s3 & 0 & 0.25 & 0.25 & 0.00 \\
sympy-19040 s3 & 40 & 0.62 & 0.62 & 0.00 \\
sympy-19040 s3 & 99 & 0.00 & 0.00 & 0.00 \\
sympy-19783 s1 & 0 & 0.75 & 0.38 & 0.38 \\
sympy-19783 s1 & 40 & 0.88 & 0.75 & 0.12 \\
sympy-19783 s1 & 99 & 0.00 & 0.00 & 0.00 \\
\bottomrule
\end{tabular}}
\caption{Test--retest: screening estimate at three anchors from the main schedule ($\hat V_1$) and from fresh seeds ($\hat V_2$), $K=8$ each.}
\label{tab:retest}
\end{table}\vspace{-5pt}
\begin{table}[htbp]
\centering
{\scriptsize\setlength{\tabcolsep}{3pt}
\begin{tabular}[t]{lrrlrlr}
\toprule
Trajectory & $T$ & $t^*_{B-t}$ & sh. & $t^*_{B}$ & sh. & final \\
\midrule
django-15128 s5 & 84 & 6 & cli & 21 & ero & 9 \\
django-16938 s4 & 65 & 6 & cli & 6 & cli & 11 \\
matplotlib-25960 s5 & 57 & 6 & cli & 41 & cli & 6 \\
requests-6028 s1 & 57 & 11 & cli & 11 & cli & 13 \\
requests-6028 s4 & 61 & 21 & cli & 21 & cli & 25 \\
sympy-13877 s1 & 58 & 16 & cli & 16 & cli & 6 \\
sympy-13877 s5 & 44 & 16 & cli & 16 & cli & 32 \\
sympy-15875 s4 & 100 & 71 & cli & 100 & cli & 35 \\
sympy-19040 s3 & 100 & 41 & cli & 41 & cli & 42 \\
sympy-19783 s1 & 100 & 46 & ero & 100 & cli & 97 \\
\bottomrule
\end{tabular}}
\caption{Budget control: screening with $B-t$ and with the full budget $B$.}
\label{tab:controls}
\end{table}\vspace{-5pt}

\FloatBarrier
\section{Labels against the curves, in full}
\label{app:rq1}

Table~\ref{tab:rq1-full} extends Table~\ref{tab:rq1} to every source and every quantity, one column per source (the four main judges, GPT-6 Luna as the contrast model, Claude Opus 5.5 under the annotator protocol on its subset, and the four main judges pooled). The upper block counts labels by position relative to the decisive interval and the point of no return; the lower block gives the direct probes at the labeled step (zero-drop: $V(h_{m-1})-V(h_m)\le0$; significant: one-sided Fisher $p<\alpha$), the H1 share (post-cause or zero-drop) with its Wilson interval, and, on runs with an explicit failure signal, labels within one step of $t^*$ and of $t_{\text{sym}}$. By shape, the pooled H1 share is 91\% on cliffs (53 labels) and 77\% on erosion (109); 7 of 47 trajectories are silent (no failure signal after the first edit). Three readings of the table. First, the H1 share is high for every source, from 63\% to 91\%, and the sources differ mainly in how they miss: the GPT-6 judges miss early (``pre'': the labeled step reads as a mistake but the curve has not moved) and the GPT-5.5/OpenAI o3 lineage misses late (``post-PNR'': the task was certified lost before the label). Second, the zero-drop row is the strongest evidence, because it does not depend on where $t^*$ is: at the labeled step itself, probed with $K\ge{}$16 on both sides, the success probability does not fall for about two thirds of the labels of every source, and a drop of $0.2$ or more is found at a few labels only. Third, the labels are not the first symptom either: on symptomatic runs almost none lie within one step of $t_{\text{sym}}$, and only a minority within one step of $t^*$; what judges mark is the step where a wrong idea is stated or acted upon, which the curve usually dates earlier. Conclusion: on this policy the first-mistake label is not a measurement of where the task was lost. The result holds for every judge, for both lineages, for the annotator-protocol model, for cliffs and erosions, and without the L2 stratum (Table~\ref{tab:rq1-noL2}); what varies between sources is the direction of the miss, not its existence.

\begin{table*}[tp]
\centering
{\footnotesize\setlength{\tabcolsep}{3pt}
\begin{tabular}[t]{lcccccccc}
\toprule
 & GPT-6.1 & GPT-5.5 & OpenAI & GPT-6 & GPT-6 & Claude & Human & Pooled \\  & Sol &  & o3 & Astra & Luna & Opus 5.5 &  &  \\
\midrule
trajectories & 47 & 47 & 47 & 47 & 47 & 30 & 30 & 47 \\
labels & 46 & 36 & 34 & 46 & 38 & 19 & 23 & 162 \\
\midrule \\
pre & 23 & 11 & 9 & 24 & 16 & 3 & 3 & 67 \\
on & 3 & 1 & 0 & 3 & 1 & 1 & 2 & 7 \\
post & 12 & 13 & 8 & 12 & 12 & 8 & 8 & 45 \\
post-PNR & 8 & 11 & 17 & 7 & 9 & 7 & 10 & 43 \\
no mistake & 1 & 11 & 13 & 1 & 9 & 11 & 7 & 26 \\
\midrule \\
zero-drop & 30/46 & 21/36 & 26/34 & 30/46 & 12/19 & 9/12 & 5/10 & 107/162 \\
drop $\ge0.2$ & 3/46 & 2/36 & 1/34 & 3/46 & 1/19 & 1/12 & 2/10 & 9/162 \\
significant & 5/46 & 2/36 & 1/34 & 5/46 & 2/19 & 1/12 & 3/10 & 13/162 \\
H1 & 78\% & 81\% & 91\% & 78\% & 63\% & 89\% & 83\% & 81\% \\
H1 95\% CI & 64--88 & 65--90 & 77--97 & 64--88 & 47--77 & 69--97 & 63--93 & 75--87 \\
near $t^*$ & 7/39 & 3/30 & 0/30 & 7/39 & 2/31 & 1/15 & 3/19 & 17/138 \\
near $t_{\text{sym}}$ & 0/39 & 1/30 & 2/30 & 0/39 & 2/31 & 0/15 & 0/19 & 3/138 \\
\bottomrule
\end{tabular}}
\caption{Labels against the curves on the frozen set (47 trajectories), every source and quantity.}
\label{tab:rq1-full}
\end{table*}
Table~\ref{tab:rq1-noL2} repeats the analysis without the L2 stratum (base runs that hit the cap), on 29 trajectories: the pooled H1 share is 85\% (83/98) against 81\% on the full set.

\begin{table*}[tp]
\centering
{\footnotesize\setlength{\tabcolsep}{3pt}
\begin{tabular}[t]{lcccccccc}
\toprule
 & GPT-6.1 & GPT-5.5 & OpenAI & GPT-6 & GPT-6 & Claude & Human & Pooled \\  & Sol &  & o3 & Astra & Luna & Opus 5.5 &  &  \\
\midrule
trajectories & 29 & 29 & 29 & 29 & 29 & 19 & 19 & 29 \\
labels & 28 & 21 & 21 & 28 & 20 & 14 & 18 & 98 \\
\midrule \\
pre & 9 & 5 & 6 & 10 & 7 & 2 & 1 & 30 \\
on & 1 & 0 & 0 & 1 & 0 & 1 & 2 & 2 \\
post & 11 & 9 & 3 & 11 & 7 & 7 & 8 & 34 \\
post-PNR & 7 & 7 & 12 & 6 & 6 & 4 & 7 & 32 \\
no mistake & 1 & 8 & 8 & 1 & 9 & 5 & 1 & 18 \\
\midrule \\
zero-drop & 17/28 & 11/21 & 15/21 & 17/28 & 8/11 & 6/9 & 3/7 & 60/98 \\
drop $\ge0.2$ & 1/28 & 0/21 & 0/21 & 1/28 & 0/11 & 1/9 & 2/7 & 2/98 \\
significant & 2/28 & 0/21 & 0/21 & 2/28 & 0/11 & 1/9 & 2/7 & 4/98 \\
H1 & 82\% & 86\% & 90\% & 82\% & 75\% & 86\% & 83\% & 85\% \\
H1 95\% CI & 64--92 & 65--95 & 71--97 & 64--92 & 53--89 & 60--96 & 61--94 & 76--90 \\
near $t^*$ & 4/22 & 2/15 & 0/17 & 4/22 & 1/14 & 1/10 & 3/14 & 10/76 \\
near $t_{\text{sym}}$ & 0/22 & 0/15 & 1/17 & 0/22 & 2/14 & 0/10 & 0/14 & 1/76 \\
\bottomrule
\end{tabular}}
\caption{The same without the L2 stratum (29 trajectories whose base run submitted).}
\label{tab:rq1-noL2}
\end{table*}

\FloatBarrier
\section{Restart policies, in full}
\label{app:restart}

Table~\ref{tab:restart-full} lists every restart policy evaluated offline on the 47 frozen curves (margin $\mu=0.1$): how many trajectories it restarts, the mean restart step, its expected success on the failed runs alone and on a random run of the task (successes included), and the gain over never restarting. Table~\ref{tab:decoys} gives the judge policies' behaviour on the successful decoys, where a label is a restart of a run that would have succeeded. Tables~\ref{tab:restart-traj} and \ref{tab:restart-traj-judges} give, per trajectory, the restart step and the expected success $R(x,B-t)$ of the value-based policies and of each judge's label. The value-based policies restart early (mean step 7 at the restart point, 29 at $t^*$) and almost always: on a failed run the only way to recover value is a fresh start while enough budget remains. The judge policies restart on 46/47 (GPT-6.1 Sol) to 34/47 (OpenAI o3) trajectories, and late (mean step 39 for GPT-5.5, 44 for OpenAI o3), so that the fresh run has little budget left; the fixed policies show the same trade-off along a single axis, restarting everything at step 10 being too eager (it forfeits successful runs that pass step 10) and at step 50 too late. The decoy table quantifies the eagerness of the judge policies on successful runs: a label on a decoy is a restart that forfeits a success, and the GPT-6 judges, which label most decoys, lose most there. The per-trajectory tables show where the judge policies lose: the GPT-6 labels fall within three steps of $t^*$ on 30 of 94 judge--trajectory pairs, and there the restart value is close to the value-based policy's; elsewhere, and for the GPT-5.5/OpenAI o3 lineage on most trajectories, the label falls long after $t^*$, often after the point of no return, where $R(x,B-t)$ has decayed to a few percent. Conclusion: the gain of a restart policy is decided by two things, how early it restarts failing runs and how rarely it restarts succeeding ones; the measured curve gives both, a certified point of no return gives the second with a known error rate at some cost in the first, and a judge's label gives neither, since it comes late on failed runs and fires on successful ones.

\begin{table}[htbp]
\centering
{\scriptsize\setlength{\tabcolsep}{2pt}
\begin{tabular}[t]{lrrrrr}
\toprule
Policy & Restarts & $\bar t$ & Failed & Random & Gain \\
\midrule
never & 0/47 & -- & 0.00 & 0.44 & +0.00 \\
restart point $t_r$ & 30/47 & 7 & 0.29 & 0.57 & +0.13 \\
$t_{\text{pnr}}$ & 47/47 & 43 & 0.20 & 0.53 & +0.09 \\
$t^*$ & 47/47 & 29 & 0.26 & 0.56 & +0.12 \\
label, GPT-6.1 Sol & 46/47 & 19 & 0.32 & 0.40 & -0.04 \\
label, GPT-5.5 & 36/47 & 39 & 0.16 & 0.36 & -0.08 \\
label, OpenAI o3 & 34/47 & 44 & 0.12 & 0.41 & -0.03 \\
label, GPT-6 Astra & 46/47 & 19 & 0.32 & 0.37 & -0.07 \\
fixed $t=10$ & 47/47 & 10 & 0.37 & 0.37 & -0.07 \\
fixed $t=20$ & 47/47 & 20 & 0.34 & 0.34 & -0.10 \\
fixed $t=30$ & 47/47 & 30 & 0.27 & 0.27 & -0.17 \\
fixed $t=50$ & 43/47 & 50 & 0.02 & 0.05 & -0.39 \\
\bottomrule
\end{tabular}}
\caption{Restart policies on the frozen set (47 trajectories). $\bar t$: mean restart step; Failed / Random: expected success on failed runs / on a random run.}
\label{tab:restart-full}
\end{table}
\begin{table}[htbp]
\centering
{\scriptsize\setlength{\tabcolsep}{3pt}
\begin{tabular}[t]{lrrr}
\toprule
Judge & Decoys & Restarted & Loss \\
\midrule
GPT-6.1 Sol & 10 & 8 & 0.46 \\
GPT-5.5 & 10 & 4 & 0.36 \\
OpenAI o3 & 10 & 3 & 0.23 \\
GPT-6 Astra & 10 & 8 & 0.53 \\
\bottomrule
\end{tabular}}
\caption{Judge policies on successful decoys: labelled decoys are restarted; expected loss $1-R(x,B-m)$ per decoy.}
\label{tab:decoys}
\end{table}

\begin{table}[htbp]
\centering
{\scriptsize\setlength{\tabcolsep}{2pt}
\begin{tabular}[t]{lrrrrr}
\toprule
Policy & Restarts & $\bar t$ & Failed & Random & Gain \\
\midrule
never & 0/29 & -- & 0.00 & 0.43 & +0.00 \\
restart point $t_r$ & 24/29 & 8 & 0.36 & 0.60 & +0.17 \\
$t_{\text{pnr}}$ & 29/29 & 28 & 0.31 & 0.57 & +0.14 \\
$t^*$ & 29/29 & 19 & 0.34 & 0.59 & +0.16 \\
label, GPT-6.1 Sol & 28/29 & 17 & 0.38 & 0.42 & -0.01 \\
label, GPT-5.5 & 21/29 & 29 & 0.23 & 0.39 & -0.04 \\
label, OpenAI o3 & 21/29 & 33 & 0.20 & 0.44 & +0.01 \\
label, GPT-6 Astra & 28/29 & 17 & 0.38 & 0.40 & -0.04 \\
fixed $t=10$ & 29/29 & 10 & 0.41 & 0.41 & -0.02 \\
fixed $t=20$ & 29/29 & 20 & 0.40 & 0.40 & -0.03 \\
fixed $t=30$ & 29/29 & 30 & 0.32 & 0.32 & -0.11 \\
fixed $t=50$ & 25/29 & 50 & 0.03 & 0.07 & -0.36 \\
\bottomrule
\end{tabular}}
\caption{Restart policies without the L2 stratum (29 trajectories).}
\label{tab:restart-noL2}
\end{table}

Table~\ref{tab:restart-noL2} repeats the evaluation without the L2 stratum (29 trajectories): the gain at the restart point is +0.17 (full set +0.13) and the judge policies gain -0.04 to +0.01.

\begin{table}[htbp]
\centering
{\scriptsize\setlength{\tabcolsep}{3pt}
\begin{tabular}[t]{lrrrr}
\toprule
Trajectory & $p(x)$ & $t_r$ & $t_{\text{pnr}}$ & $t^*$ \\
\midrule
django-11555 s2 & 0.50 & -- & 71/0.00 & 35/0.00 \\
django-11555 s5 & 0.50 & -- & 93/0.00 & 86/0.00 \\
django-11555 s7 & 0.50 & -- & 82/0.00 & 6/0.50 \\
django-12325 s1 & 0.25 & 0/0.25 & 15/0.25 & 11/0.25 \\
django-12325 s2 & 0.25 & 0/0.25 & 2/0.25 & 2/0.25 \\
django-12325 s6 & 0.25 & 0/0.25 & 0/0.25 & 16/0.25 \\
django-13807 s2 & 0.62 & 0/0.62 & 15/0.62 & 1/0.62 \\
django-13807 s3 & 0.62 & 0/0.62 & 13/0.62 & 13/0.62 \\
django-13807 s7 & 0.62 & 0/0.62 & 18/0.62 & 6/0.62 \\
django-15128 s4 & 0.50 & 2/0.50 & 63/0.00 & 17/0.38 \\
django-15128 s5 & 0.50 & 10/0.50 & 40/0.00 & 9/0.50 \\
django-15128 s6 & 0.50 & 6/0.50 & 81/0.00 & 51/0.00 \\
django-16100 s1 & 0.38 & -- & 60/0.00 & 39/0.12 \\
django-16100 s2 & 0.38 & -- & 50/0.00 & 50/0.00 \\
django-16100 s6 & 0.38 & 21/0.38 & 48/0.00 & 48/0.00 \\
django-16938 s2 & 0.50 & 11/0.50 & 50/0.00 & 11/0.50 \\
django-16938 s4 & 0.50 & 7/0.50 & 11/0.50 & 11/0.50 \\
django-16938 s5 & 0.50 & -- & 97/0.00 & 95/0.00 \\
matplotlib-22865 s2 & 0.25 & 27/0.25 & 30/0.25 & 6/0.25 \\
matplotlib-22865 s3 & 0.25 & 19/0.25 & 22/0.25 & 19/0.25 \\
matplotlib-22865 s7 & 0.25 & 12/0.25 & 20/0.25 & 12/0.25 \\
matplotlib-25960 s0 & 0.25 & 0/0.25 & 0/0.25 & 22/0.25 \\
matplotlib-25960 s3 & 0.25 & 0/0.25 & 25/0.25 & 11/0.25 \\
matplotlib-25960 s5 & 0.25 & 6/0.25 & 6/0.25 & 6/0.25 \\
seaborn-3069 s2 & 0.38 & -- & 91/0.00 & 14/0.12 \\
seaborn-3069 s3 & 0.38 & -- & 30/0.00 & 2/0.38 \\
seaborn-3069 s5 & 0.38 & -- & 62/0.00 & 62/0.00 \\
requests-6028 s0 & 0.25 & 0/0.25 & 26/0.00 & 26/0.00 \\
requests-6028 s1 & 0.25 & 14/0.25 & 15/0.25 & 13/0.25 \\
requests-6028 s4 & 0.25 & -- & 29/0.00 & 25/0.00 \\
scikit-learn-10908 s3 & 0.75 & 19/0.75 & 19/0.75 & 19/0.75 \\
sphinx-8265 s0 & 0.75 & 0/0.75 & 22/0.75 & 16/0.75 \\
sphinx-8265 s5 & 0.75 & 18/0.75 & 21/0.75 & 18/0.75 \\
sphinx-8593 s2 & 0.62 & -- & 99/0.00 & 51/0.00 \\
sphinx-8593 s6 & 0.62 & -- & 76/0.00 & 98/0.00 \\
sphinx-9591 s2 & 0.25 & -- & 67/0.00 & 48/0.00 \\
sphinx-9591 s4 & 0.25 & -- & 70/0.00 & 28/0.00 \\
sphinx-9591 s5 & 0.25 & -- & 47/0.00 & 8/0.12 \\
sympy-13877 s1 & 0.50 & 17/0.50 & 28/0.50 & 6/0.50 \\
sympy-13877 s5 & 0.50 & 20/0.50 & 32/0.50 & 32/0.50 \\
sympy-15875 s1 & 0.75 & 0/0.75 & 16/0.75 & 15/0.75 \\
sympy-15875 s4 & 0.75 & 0/0.75 & 88/0.00 & 35/0.62 \\
sympy-19040 s0 & 0.25 & 4/0.25 & 4/0.25 & 1/0.25 \\
sympy-19040 s3 & 0.25 & -- & 52/0.00 & 42/0.00 \\
sympy-19040 s7 & 0.25 & 0/0.25 & 31/0.12 & 66/0.00 \\
sympy-19783 s1 & 0.75 & -- & 97/0.00 & 97/0.00 \\
sympy-19783 s5 & 0.75 & 2/0.75 & 90/0.00 & 46/0.00 \\
\bottomrule
\end{tabular}}
\caption{Value-based policies per trajectory: restart step / expected success $R(x,B-t)$ (-- = no restart).}
\label{tab:restart-traj}
\end{table}
\begin{table}[htbp]
\centering
{\scriptsize\setlength{\tabcolsep}{3pt}
\begin{tabular}[t]{lrrrr}
\toprule
 & GPT-6.1 & GPT-6 & GPT-5.5 & OpenAI \\ Trajectory & Sol & Astra &  & o3 \\
\midrule
django-11555 s2 & 13/0.25 & 13/0.25 & 56/0.00 & 44/0.00 \\
django-11555 s5 & 12/0.25 & 12/0.25 & 58/0.00 & -- \\
django-11555 s7 & 13/0.25 & 13/0.25 & 82/0.00 & 82/0.00 \\
django-12325 s1 & 27/0.25 & 27/0.25 & -- & 28/0.25 \\
django-12325 s2 & 38/0.12 & 38/0.12 & -- & 22/0.25 \\
django-12325 s6 & 28/0.25 & 28/0.25 & -- & -- \\
django-13807 s2 & 10/0.62 & 10/0.62 & -- & 18/0.62 \\
django-13807 s3 & 10/0.62 & 10/0.62 & -- & -- \\
django-13807 s7 & 8/0.62 & 8/0.62 & -- & -- \\
django-15128 s4 & 11/0.50 & 11/0.50 & 40/0.00 & 65/0.00 \\
django-15128 s5 & 10/0.50 & 10/0.50 & 30/0.25 & 76/0.00 \\
django-15128 s6 & 10/0.50 & 10/0.50 & 65/0.00 & 10/0.50 \\
django-16100 s1 & 17/0.38 & 17/0.38 & 17/0.38 & 21/0.38 \\
django-16100 s2 & 2/0.38 & 2/0.38 & -- & 24/0.38 \\
django-16100 s6 & 20/0.38 & 20/0.38 & 21/0.38 & 21/0.38 \\
django-16938 s2 & 11/0.50 & 11/0.50 & 99/0.00 & -- \\
django-16938 s4 & 11/0.50 & 11/0.50 & 16/0.50 & -- \\
django-16938 s5 & 8/0.50 & 8/0.50 & -- & -- \\
matplotlib-22865 s2 & 16/0.25 & 16/0.25 & 28/0.25 & -- \\
matplotlib-22865 s3 & 20/0.25 & 20/0.25 & 20/0.25 & 18/0.25 \\
matplotlib-22865 s7 & 21/0.25 & 21/0.25 & 23/0.25 & 21/0.25 \\
matplotlib-25960 s0 & 29/0.25 & 17/0.25 & 21/0.25 & 29/0.25 \\
matplotlib-25960 s3 & 16/0.25 & 21/0.25 & 16/0.25 & 21/0.25 \\
matplotlib-25960 s5 & 17/0.25 & 17/0.25 & 17/0.25 & 49/0.12 \\
seaborn-3069 s2 & 14/0.12 & 14/0.12 & 14/0.12 & 45/0.00 \\
seaborn-3069 s3 & 16/0.12 & 16/0.12 & 96/0.00 & 59/0.00 \\
seaborn-3069 s5 & 16/0.12 & 16/0.12 & 29/0.00 & 29/0.00 \\
requests-6028 s0 & 18/0.25 & 18/0.25 & 18/0.25 & -- \\
requests-6028 s1 & 14/0.25 & 14/0.25 & 14/0.25 & 21/0.12 \\
requests-6028 s4 & 24/0.12 & 24/0.12 & 35/0.00 & 35/0.00 \\
scikit-learn-10908 s3 & -- & -- & -- & -- \\
sphinx-8265 s0 & 19/0.75 & 19/0.75 & 19/0.75 & -- \\
sphinx-8265 s5 & 19/0.75 & 19/0.75 & 26/0.62 & 73/0.00 \\
sphinx-8593 s2 & 48/0.00 & 48/0.00 & 99/0.00 & 48/0.00 \\
sphinx-8593 s6 & 76/0.00 & 76/0.00 & -- & -- \\
sphinx-9591 s2 & 7/0.12 & 7/0.12 & 83/0.00 & 83/0.00 \\
sphinx-9591 s4 & 8/0.12 & 8/0.12 & 48/0.00 & 48/0.00 \\
sphinx-9591 s5 & 7/0.12 & 7/0.12 & -- & 76/0.00 \\
sympy-13877 s1 & 7/0.50 & 7/0.50 & 34/0.50 & 34/0.50 \\
sympy-13877 s5 & 16/0.50 & 16/0.50 & 16/0.50 & 20/0.50 \\
sympy-15875 s1 & 17/0.75 & 17/0.75 & 27/0.75 & 27/0.75 \\
sympy-15875 s4 & 12/0.75 & 12/0.75 & 55/0.25 & 58/0.12 \\
sympy-19040 s0 & 62/0.00 & 62/0.00 & 62/0.00 & -- \\
sympy-19040 s3 & 11/0.25 & 11/0.25 & 35/0.12 & 46/0.00 \\
sympy-19040 s7 & 24/0.25 & 24/0.25 & 24/0.25 & 91/0.00 \\
sympy-19783 s1 & 24/0.12 & 22/0.25 & 39/0.00 & 75/0.00 \\
sympy-19783 s5 & 26/0.12 & 26/0.12 & 36/0.12 & 76/0.00 \\
\bottomrule
\end{tabular}}
\caption{Judge-label policies per trajectory: restart step / expected success (-- = no mistake, no restart).}
\label{tab:restart-traj-judges}
\end{table}

\FloatBarrier
\section{Online restart check}
\label{app:online}

Table~\ref{tab:online} lists the online check of the restart evaluation: for each of the 30 frozen trajectories with a restart point $t_r$ (17 have none), eight fresh runs of the task with budget $B-t_r$ and new seeds, evaluated by the harness, next to the value of continuing $V(h_{t_r})$ and the restart value $R(x,B-t_r)$ predicted from the base runs. Over the 240 runs the observed success is 0.39 (Wilson 0.33--0.45; mean over trajectories 0.39) against a predicted 0.45 and a continuing value of 0.26; the prediction is optimistic by 0.06 (task-cluster bootstrap 95\% interval -0.15 to +0.04), and restarting beats continuing on the same trajectories by a wide margin. Mean restart step 7. Per trajectory, the observed success is below the prediction on 14 of 30 trajectories and above it on 9, and above the value of continuing on 20; 29 of the 240 fresh runs hit the step cap. Two sources explain the gap between observed and predicted. The prediction $\hat R(x,B-t_r)$ is a step function from eight base runs, so it is coarse and, since the restart points are early (small $t_r$), it is close to $p(x)$; the fresh runs are drawn at a different time from the same server and carry the ordinary task-level variance of a pass rate estimated from eight runs. The rows with $t_r=0$ make the second source visible: there the check replicates the base pass rate itself, and the fresh runs come out below it on 9 of 13 such trajectories and above it on 0. This is selection, not drift: a restart point at $t=0$ exists only when $\hat R(x,B)-\hat V(h_0)\ge\mu$, that is, when the eight base runs came out high relative to the screening at $h_0$, so the tasks with early restart points are those whose base pass rate was an upward fluctuation, and a fresh draw regresses toward the mean. The offline evaluation inherits this optimism in $\hat R$. The conclusion the main text draws from the offline evaluation, that restarting at the restart point gains and restarting at a judge's label does not, rests on the difference between $R(x,B-t)$ and $V(h_t)$, which the fresh runs confirm; a modest optimism of $\hat R$ reduces the size of the gain, not its sign. Conclusion: restarting at the measured restart point is worth about 0.13 in success probability on the affected trajectories when the restart is actually executed, against 0.19 predicted offline; the offline evaluation should be read as an upper bound of that size, and a controller that estimates $R$ online, from its own fresh runs, will see the smaller number.

\begin{table}[htbp]
\centering
{\scriptsize\setlength{\tabcolsep}{3pt}
\begin{tabular}[t]{lrrrr}
\toprule
Trajectory & $t_r$ & $V$ & $R$ pred. & Obs. \\
\midrule
django-12325 s1 & 0 & 0.07 & 0.25 & 0/8 \\
django-12325 s2 & 0 & 0.06 & 0.25 & 0/8 \\
django-12325 s6 & 0 & 0.05 & 0.25 & 0/8 \\
django-13807 s2 & 0 & 0.44 & 0.62 & 5/8 \\
django-13807 s3 & 0 & 0.50 & 0.62 & 2/8 \\
django-13807 s7 & 0 & 0.25 & 0.62 & 3/8 \\
django-15128 s4 & 2 & 0.38 & 0.50 & 7/8 \\
django-15128 s5 & 10 & 0.34 & 0.50 & 6/8 \\
django-15128 s6 & 6 & 0.31 & 0.50 & 5/8 \\
django-16100 s6 & 21 & 0.19 & 0.38 & 3/8 \\
django-16938 s2 & 11 & 0.33 & 0.50 & 3/8 \\
django-16938 s4 & 7 & 0.38 & 0.50 & 5/8 \\
matplotlib-22865 s2 & 27 & 0.12 & 0.25 & 3/8 \\
matplotlib-22865 s3 & 19 & 0.07 & 0.25 & 1/8 \\
matplotlib-22865 s7 & 12 & 0.12 & 0.25 & 3/8 \\
matplotlib-25960 s0 & 0 & 0.12 & 0.25 & 0/8 \\
matplotlib-25960 s3 & 0 & 0.12 & 0.25 & 1/8 \\
matplotlib-25960 s5 & 6 & 0.01 & 0.25 & 1/8 \\
requests-6028 s0 & 0 & 0.12 & 0.25 & 1/8 \\
requests-6028 s1 & 14 & 0.08 & 0.25 & 0/8 \\
scikit-learn-10908 s3 & 19 & 0.00 & 0.75 & 7/8 \\
sphinx-8265 s0 & 0 & 0.50 & 0.75 & 6/8 \\
sphinx-8265 s5 & 18 & 0.53 & 0.75 & 4/8 \\
sympy-13877 s1 & 17 & 0.38 & 0.50 & 5/8 \\
sympy-13877 s5 & 20 & 0.38 & 0.50 & 5/8 \\
sympy-15875 s1 & 0 & 0.57 & 0.75 & 6/8 \\
sympy-15875 s4 & 0 & 0.54 & 0.75 & 1/8 \\
sympy-19040 s0 & 4 & 0.04 & 0.25 & 2/8 \\
sympy-19040 s7 & 0 & 0.15 & 0.25 & 2/8 \\
sympy-19783 s5 & 2 & 0.57 & 0.75 & 6/8 \\
\bottomrule
\end{tabular}}
\caption{Online restart check: observed successes of eight fresh runs with budget $B-t_r$, next to $V(h_{t_r})$ and the predicted $R(x,B-t_r)$.}
\label{tab:online}
\end{table}\vspace{-5pt}

\FloatBarrier
\section{Cheaper schedules, in full}
\label{app:cheap}

Table~\ref{tab:cheap-full} reports the parametric bootstrap of cheaper schedules (20 replications per trajectory on 47 curves): continuations used, agreement of the estimated decisive step with the fit's largest drop (exact, within 1, within 3), the share of replications that find a drop at all, and the share of the true drop captured within $\pm10$ steps. Even the full schedule is exact in 28\% of replications. Judge-guided schedules probe the judge's label first and fall back to a search when the drop there is too small; Table~\ref{tab:cheap-paths} counts the paths taken, and Table~\ref{tab:cheap-shape} splits the comparison by failure shape. The rows order themselves by cost: the full schedule and the judge-guided schedules cost the most, screening with bisection but no confirmation about half, screening at $K=4$ and sparse anchors less again. Accuracy does not follow cost: the sparse schedule matches the full one within noise, because the largest drop is rarely between two anchors ten steps apart, while dropping confirmation loses exact agreement because the $K=8$ estimates at the bracketing points are too noisy to order neighbouring drops. The value-captured column is the quantity a controller cares about and it is far more forgiving than exact agreement: a schedule that misses $t^*$ by a few steps still captures most of the drop. The judge-guided schedules take the ``label verified'' path in a small minority of replications; in the rest they fall back to a local search around the label and then to the full schedule, which is why their cost equals the full schedule's plus the probe. By shape, cliffs are located far more reliably than erosions under every schedule, which is the expected behaviour of an estimator that looks for one large drop. Conclusion: the decisive step is an expensive and, on erosions, ill-defined target, while the drop itself can be captured at a fraction of the cost; a practical localizer should be evaluated on value captured and budgeted for screening and certification rather than for exact localization, and should not route its search through a judge's label.

\begin{table}[htbp]
\centering
{\scriptsize\setlength{\tabcolsep}{3pt}
\begin{tabular}[t]{lrrrrrr}
\toprule
Schedule & $n$ & Exact & $\pm1$ & $\pm3$ & Found & Capt. \\
\midrule
full schedule & 276 & 28\% & 36\% & 47\% & 64\% & 0.44 \\
screening + bisection & 147 & 16\% & 25\% & 34\% & 82\% & 0.30 \\
screening at $K=4$ & 203 & 16\% & 24\% & 34\% & 84\% & 0.30 \\
sparse anchors & 230 & 28\% & 38\% & 50\% & 57\% & 0.45 \\
judge-guided, GPT-6.1 Sol & 255 & 27\% & 36\% & 47\% & 65\% & 0.44 \\
judge-guided, GPT-5.5 & 255 & 27\% & 37\% & 47\% & 66\% & 0.42 \\
judge-guided, OpenAI o3 & 267 & 26\% & 35\% & 45\% & 65\% & 0.42 \\
judge-guided, GPT-6 Astra & 256 & 26\% & 35\% & 48\% & 63\% & 0.42 \\
\bottomrule
\end{tabular}}
\caption{Cheaper schedules: continuations, agreement with the full curve's decisive step, drops found, value captured.}
\label{tab:cheap-full}
\end{table}\vspace{-5pt}
\begin{table}[htbp]
\centering
{\scriptsize\setlength{\tabcolsep}{3pt}
\begin{tabular}[t]{p{0.3\columnwidth}p{0.55\columnwidth}}
\toprule
Schedule & Paths \\
\midrule
judge-guided, GPT-6.1 Sol & fallback full 471, label verified 40, local search 409, no label: full 20 \\
judge-guided, GPT-5.5 & fallback full 346, label verified 17, local search 357, no label: full 220 \\
judge-guided, OpenAI o3 & fallback full 364, label verified 7, local search 309, no label: full 260 \\
judge-guided, GPT-6 Astra & fallback full 475, label verified 47, local search 398, no label: full 20 \\
\bottomrule
\end{tabular}}
\caption{Paths taken by the judge-guided schedules (replications).}
\label{tab:cheap-paths}
\end{table}\vspace{-5pt}
\begin{table}[htbp]
\centering
{\scriptsize\setlength{\tabcolsep}{3pt}
\begin{tabular}[t]{llrrr}
\toprule
Shape & Schedule & Exact & $\pm3$ & Capt. \\
\midrule
cli & full schedule & 39\% & 53\% & 0.45 \\
cli & screening + bisection & 23\% & 40\% & 0.29 \\
cli & screening at $K=4$ & 27\% & 40\% & 0.35 \\
cli & sparse anchors & 44\% & 58\% & 0.50 \\
cli & judge-guided, GPT-6.1 Sol & 44\% & 60\% & 0.52 \\
cli & judge-guided, GPT-5.5 & 45\% & 58\% & 0.52 \\
cli & judge-guided, OpenAI o3 & 41\% & 58\% & 0.47 \\
cli & judge-guided, GPT-6 Astra & 42\% & 61\% & 0.48 \\
ero & full schedule & 22\% & 44\% & 0.44 \\
ero & screening + bisection & 11\% & 30\% & 0.30 \\
ero & screening at $K=4$ & 9\% & 31\% & 0.27 \\
ero & sparse anchors & 19\% & 45\% & 0.42 \\
ero & judge-guided, GPT-6.1 Sol & 18\% & 39\% & 0.40 \\
ero & judge-guided, GPT-5.5 & 16\% & 40\% & 0.36 \\
ero & judge-guided, OpenAI o3 & 18\% & 37\% & 0.39 \\
ero & judge-guided, GPT-6 Astra & 17\% & 40\% & 0.39 \\
\bottomrule
\end{tabular}}
\caption{Cheaper schedules by failure shape.}
\label{tab:cheap-shape}
\end{table}\vspace{-5pt}

\FloatBarrier
\section{Ranking localizers, in full}
\label{app:rq3}

Table~\ref{tab:rq3-full} scores every localizer by value captured within $k$ steps (VC$_k$), by agreement with the labels of the judges of the other lineages within $k$ steps (A$_k$; a lineage consensus exists on only 22 trajectories), and by agreement with $t^*$ ($t^*_k$). Kendall's $\tau$ between the rankings by VC$_1$ and by A$_1$ over the 9 localizers scored on all trajectories is 0.56 (permutation $p=0.04$); by VC$_1$ the order is GPT-6 Astra, GPT-6.1 Sol, GPT-5.5, GPT-6 Luna (contrast), first edit, OpenAI o3, step 1, before first symptom, first verification; by A$_1$ it is GPT-6 Luna (contrast), GPT-5.5, GPT-6.1 Sol, GPT-6 Astra, OpenAI o3, first edit, before first symptom, step 1, first verification. Three features of the table. The absolute values of VC$_k$ are low for every localizer, judges and heuristics alike: on erosion curves the drop is spread over many steps and no single step can capture much of it, and on cliffs the judges' labels usually fall a few steps after the drop, so widening the window to $k=3$ roughly doubles the value captured without changing the order. The heuristics are informative baselines: ``first edit'' captures as much value as a mid-ranked judge because the decisive step is usually at or before the first edit (\S\ref{sec:shapes}), while ``step 1'' and ``before first symptom'' capture almost nothing, the first because the drop is rarely at the very start and the second because the symptom comes long after the cause. Agreement with labels (A$_k$) rewards the judges that label like the other judges: the contrast model and GPT-5.5 rank first and second by A$_1$ and in the middle by VC$_1$, while the GPT-6 judges, which capture the most value, rank below them by agreement because their early labels do not coincide with the later labels of the other lineage. Agreement with $t^*$ ($t^*_k$) is low throughout and is reported for reference only; it inherits the instability of a single decisive step on erosions (Appendix~\ref{app:cheap}). Conclusion: a benchmark that ranks localizers by agreement with labels and one that ranks them by value captured would pick different winners, and the second criterion is the one that corresponds to what a controller gains; the measured curves make the second criterion computable for any proposed localizer, including ones not studied here.

\begin{table}[htbp]
\centering
{\scriptsize\setlength{\tabcolsep}{2.5pt}
\begin{tabular}[t]{lrrrrrrrrr}
\toprule
Localizer & $n$ & VC$_0$ & VC$_1$ & VC$_3$ & A$_0$ & A$_1$ & A$_3$ & $t^*_0$ & $t^*_3$ \\
\midrule
GPT-6 Astra & 47 & 0.07 & 0.16 & 0.26 & 0.17 & 0.20 & 0.22 & 0.06 & 0.32 \\
GPT-6.1 Sol & 47 & 0.07 & 0.16 & 0.23 & 0.18 & 0.21 & 0.23 & 0.06 & 0.32 \\
Human annotator & 30 & 0.08 & 0.14 & 0.22 & 0.18 & 0.23 & 0.33 & 0.03 & 0.17 \\
GPT-5.5 & 47 & 0.04 & 0.09 & 0.13 & 0.26 & 0.28 & 0.31 & 0.02 & 0.11 \\
\twoline{Claude Opus 5.5}{(annotator)} & 30 & 0.03 & 0.08 & 0.19 & 0.25 & 0.28 & 0.33 & 0.00 & 0.13 \\
\twoline{GPT-6 Luna}{(contrast)} & 47 & 0.02 & 0.07 & 0.12 & 0.27 & 0.30 & 0.38 & 0.02 & 0.09 \\
first edit & 47 & 0.05 & 0.07 & 0.15 & 0.16 & 0.18 & 0.24 & 0.02 & 0.13 \\
OpenAI o3 & 47 & 0.02 & 0.05 & 0.08 & 0.17 & 0.20 & 0.23 & 0.00 & 0.04 \\
step 1 & 47 & 0.01 & 0.04 & 0.05 & 0.00 & 0.01 & 0.01 & 0.04 & 0.09 \\
random step & 47 & 0.01 & 0.04 & 0.10 & -- & -- & -- & -- & -- \\
before first symptom & 47 & 0.00 & 0.01 & 0.07 & 0.04 & 0.04 & 0.05 & 0.00 & 0.04 \\
first verification & 47 & 0.01 & 0.01 & 0.05 & 0.01 & 0.01 & 0.02 & 0.04 & 0.09 \\
\bottomrule
\end{tabular}}
\caption{Localizers by value captured (VC$_k$), label agreement (A$_k$) and agreement with $t^*$; sorted by VC$_1$.}
\label{tab:rq3-full}
\end{table}\vspace{-5pt}

\FloatBarrier
\section{Synthetic validation of the estimator}
\label{app:synthetic}

Before any real curve was built, the estimator and schedule were run on 2,000 synthetic curves of four families with known $t^*$ and shape (seed 20260928; Table~\ref{tab:synthetic}): shape classification accuracy, share classified as cliff, share with a detected drop, with an onset, with a certified point of no return, and continuations used. Overall shape accuracy is 92\% (with the literal $C$ it would be 49\%, which motivated the isotonic estimator); flat curves are called cliffs in 3\% of cases; on cliffs of size $\ge0.4$ ($n=881$) the median $|\hat t^*-t^*|$ is 0 and 94\% are within one step; a curve costs 321 continuations on average (median 328, p90 408). The families behave as designed. Cliffs and plateau-cliffs (a flat high value followed by one drop) are detected and classified reliably, with an onset found in most and a certified point of no return in all; erosions are classified correctly as erosions precisely because no single drop is large enough to be ``detected'', and their point of no return is still certified since the value does reach zero; flat curves (a constant value, no failure to localize) produce a certified point of no return in none and a detected drop in a few percent, which is the estimator's false-positive rate under sampling noise. The share of flat curves classified as cliffs is not an error rate: $C$ is undefined for a curve with no drop and the label is reported for completeness. The synthetic cost per curve matches the real cost in Appendix~\ref{app:schedule}, so the schedule's budget was set from these simulations before the first continuation was run. Conclusion: the estimator finds and classifies the drop it is designed for, certifies the point of no return whenever the value truly reaches zero, and does not invent drops on flat curves beyond its nominal false-positive rate; its known weakness, locating a single step on an erosion, is a property of the target, not of the sample size, and is reported as such in the main text.

\begin{table}[htbp]
\centering
{\scriptsize\setlength{\tabcolsep}{3pt}
\begin{tabular}[t]{lrrrrrrr}
\toprule
Family & $n$ & Acc. & Cliff & Drop & Onset & PNR & Cont. \\
\midrule
cliff & 500 & 79\% & 79\% & 72\% & 81\% & 100\% & 327 \\
erosion & 500 & 99\% & 1\% & 5\% & 27\% & 100\% & 349 \\
flat & 500 & -- & 71\% & 3\% & 19\% & 0\% & 292 \\
plateau\_cliff & 500 & 97\% & 97\% & 87\% & 91\% & 100\% & 316 \\
\bottomrule
\end{tabular}}
\caption{Estimator validation on synthetic curves with known $t^*$ and shape.}
\label{tab:synthetic}
\end{table}\vspace{-5pt}

\FloatBarrier
\section{The annotators: Claude Opus 5.5 and a human}
\label{app:annotator}

Claude Opus 5.5 labelled the annotation set (30 failed trajectories, 10 decoys) under the annotator protocol (\S\ref{sec:setup}). Table~\ref{tab:annotator} gives its agreement with each judge on the common trajectories. It answers ``no mistake'' on 11 failed trajectories and flags 0 decoys; against the curves of the 30 resolved trajectories of the subset its labels fall after it 8, inside it 1, no mistake 11, after the point of no return 7, before the decisive interval 3. The protocol differs from the judges' in three ways that matter for reading the table: the model sees the trajectory as static pages rather than as one prompt, it is asked to record alternative candidates and a confidence, and it labels in a random order with no outcome shown, exactly as a human annotator would. Under this protocol it flags no decoy at all and answers ``no mistake'' on more than a third of the failed trajectories, the opposite of the GPT-6 judges' behaviour and closest to GPT-5.5, with which it also agrees most often. Against the curves it is no closer to the drop than the judges: its labels fall after the point of no return as often as inside the decisive interval, and the value does not move at most of the labeled steps it was possible to probe. The subset is small, so the annotator is reported beside the judges rather than ranked with them; it serves as a check that the main result is not an artefact of one prompt or one vendor. Conclusion: a model that is careful enough never to flag a successful run still places its labels where the agent's own success probability does not move; calibration on decoys and localization of the cause are separate properties, and the second is the one the paper measures. A human annotator with a software-engineering background, blind to outcomes, curves and the judges' labels, labelled the same 40 trajectories from the same static pages under the same rubric (one annotator; no inter-annotator statistics are possible). Table~\ref{tab:human} gives the agreement of the human labels with every other source. The annotator flags 1 of the 10 decoys and answers ``no mistake'' on 7 of the 30 failed trajectories, and agrees most with Claude Opus 5.5 (15/30 exact); against the curves, 2 of the 23 labels fall inside the decisive interval, 10 after the point of no return, and $V$ does not drop at 5 of the 10 probed labels (H1 83\%). Conclusion: a careful human reader, like the annotator-protocol model, avoids flagging successful runs, and like every other source places most labels where the agent's own success probability had already fallen or does not move.

\begin{table}[htbp]
\centering
{\scriptsize\setlength{\tabcolsep}{3pt}
\begin{tabular}[t]{lrrrr}
\toprule
Judge & $n$ & Exact & $\pm1$ & $\pm3$ \\
\midrule
GPT-6.1 Sol & 40 & 6 & 7 & 8 \\
GPT-5.5 & 40 & 19 & 21 & 23 \\
OpenAI o3 & 40 & 14 & 14 & 15 \\
GPT-6 Astra & 40 & 8 & 9 & 10 \\
GPT-6 Luna (contrast) & 40 & 12 & 13 & 16 \\
\bottomrule
\end{tabular}}
\caption{Claude Opus 5.5 (annotator protocol): agreement with each judge, exact / within 1 / within 3 steps.}
\label{tab:annotator}
\end{table}\vspace{-5pt}
\begin{table}[htbp]
\centering
{\scriptsize\setlength{\tabcolsep}{3pt}
\begin{tabular}[t]{lrrrr}
\toprule
Source & $n$ & Exact & $\pm1$ & $\pm3$ \\
\midrule
GPT-6.1 Sol & 30 & 6 & 8 & 12 \\
GPT-6 Astra & 30 & 5 & 6 & 10 \\
GPT-5.5 & 30 & 9 & 8 & 10 \\
OpenAI o3 & 30 & 2 & 1 & 3 \\
GPT-6 Luna (contrast) & 30 & 4 & 3 & 6 \\
Claude Opus 5.5 (annotator) & 30 & 15 & 12 & 14 \\
\bottomrule
\end{tabular}}
\caption{Human annotator: agreement with each judge and with Claude Opus 5.5, exact / within 1 / within 3 steps.}
\label{tab:human}
\end{table}\vspace{-5pt}

\section{Disclosure of AI Use}
The authors have made use of AI tools for coding assistance and writing refinement.

\FloatBarrier
\clearpage
\section{All V-curves}
\label{app:gallery}

\begin{figure}[H]
\centering
\includegraphics[width=0.96\textwidth]{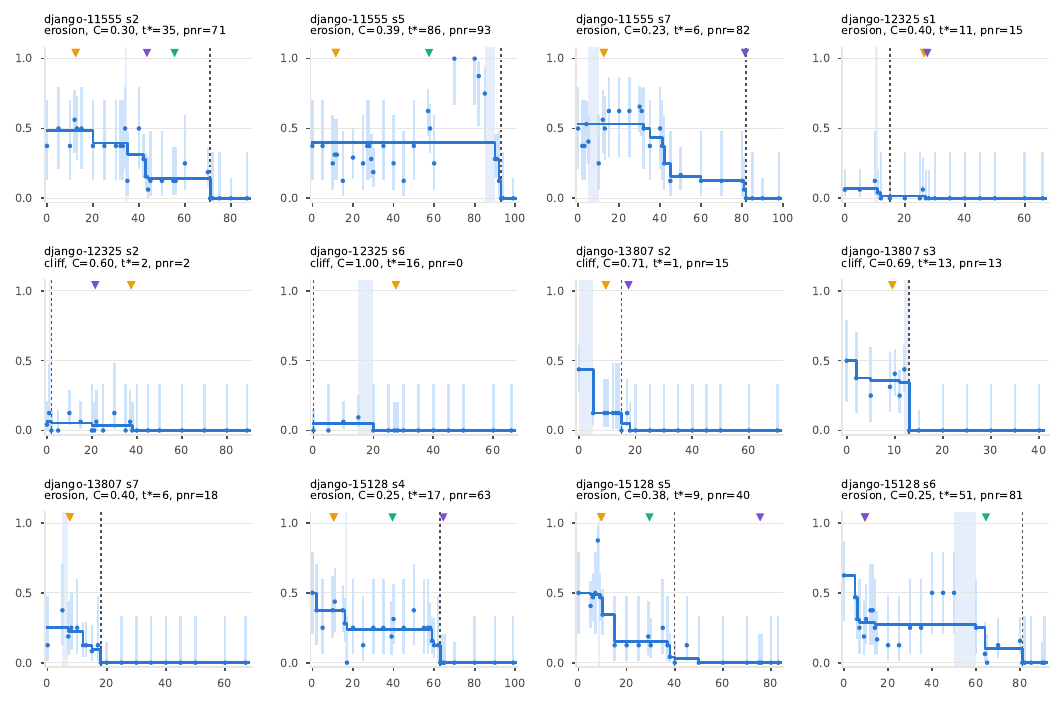}
\caption{V-curves of the frozen set (1 of 4); one diagnostic note per curve follows.}
\label{fig:gallery1}
\end{figure}

\begin{figure}[H]
\centering
\includegraphics[width=0.96\textwidth]{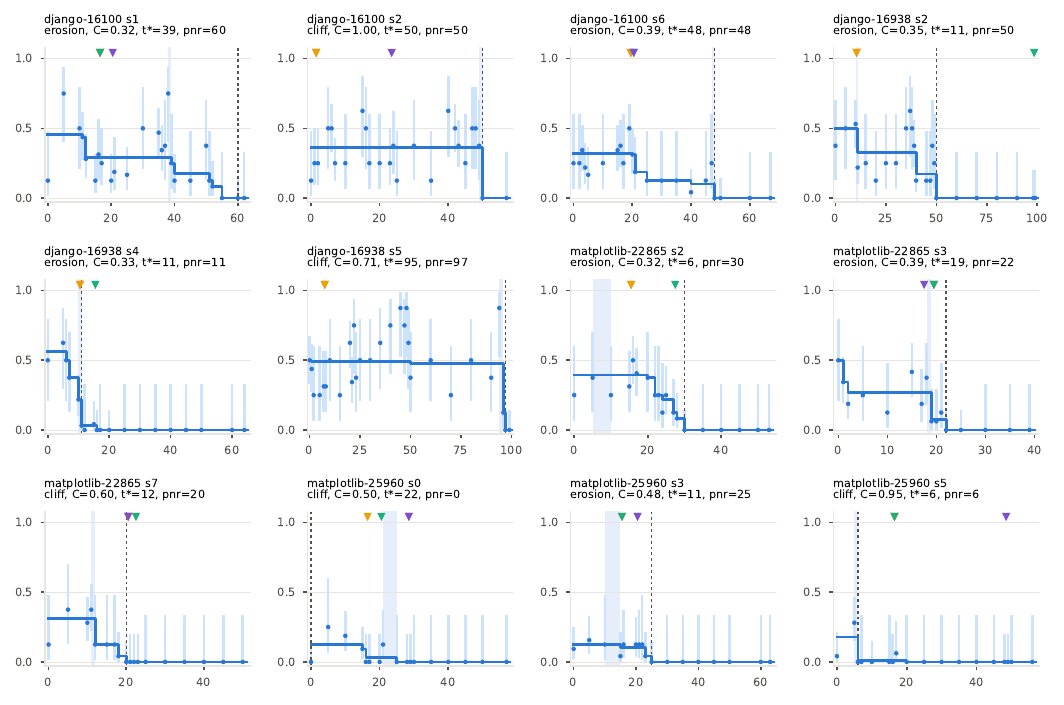}
\caption{V-curves of the frozen set (2 of 4); one diagnostic note per curve follows.}
\label{fig:gallery2}
\end{figure}

\begin{figure}[H]
\centering
\includegraphics[width=0.96\textwidth]{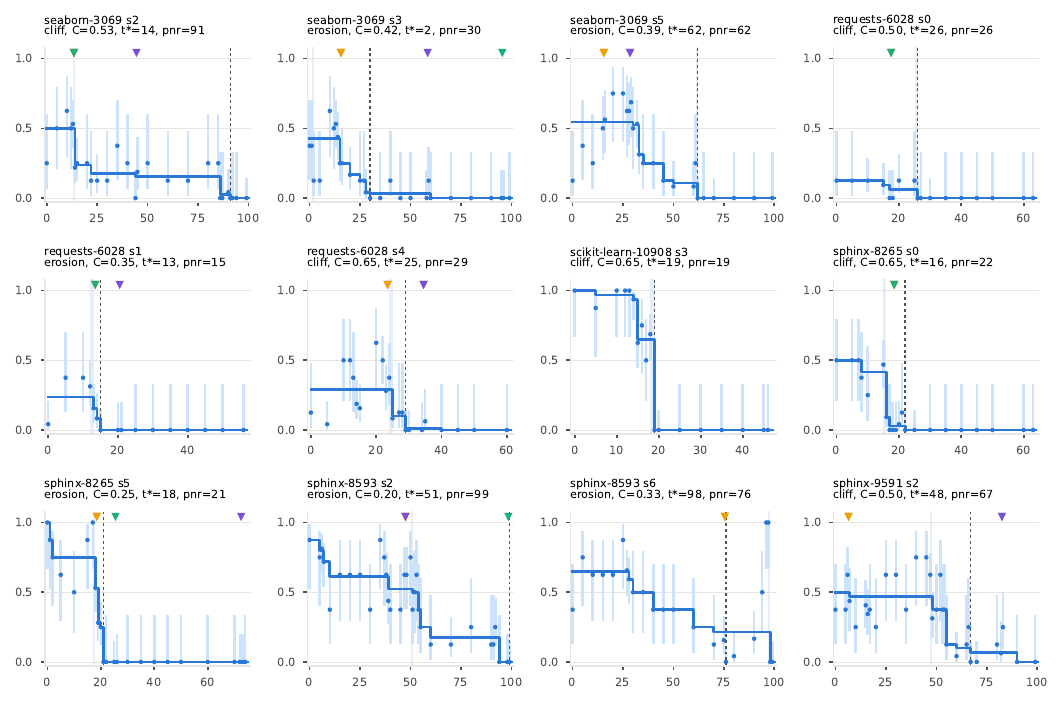}
\caption{V-curves of the frozen set (3 of 4); one diagnostic note per curve follows.}
\label{fig:gallery3}
\end{figure}

\begin{figure}[H]
\centering
\includegraphics[width=0.96\textwidth]{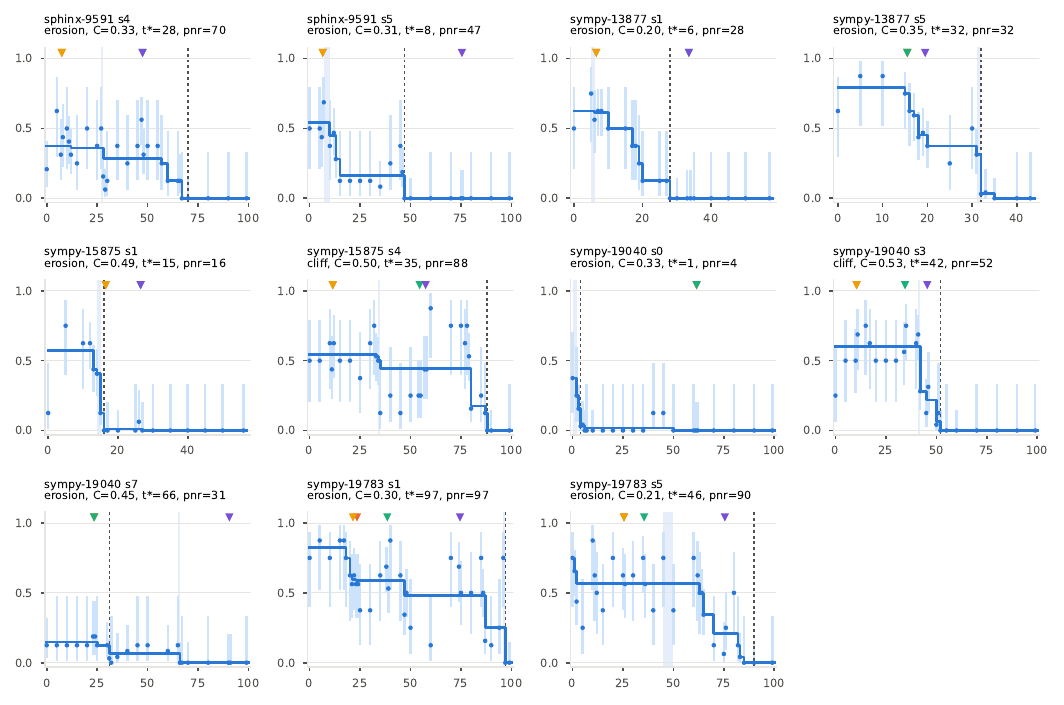}
\caption{V-curves of the frozen set (4 of 4); one diagnostic note per curve follows.}
\label{fig:gallery4}
\end{figure}

Figures~\ref{fig:gallery1}--\ref{fig:gallery4} show every curve of the frozen set in processing order, with the quantities of Table~\ref{tab:alltraj} in each title: the isotonic fit (line), the raw estimates with their Wilson 95\% bars (dots), the decisive interval (shaded), the certified point of no return (dotted) and the main judges' labels (triangles above the axis: GPT-6.1 Sol orange, GPT-6 Astra yellow, GPT-5.5 green, OpenAI o3 purple). A few patterns are visible across the pages. Cliffs show a single step from a high plateau to zero (their decisive step sits at a median of 35\% of the run) with the labels clustered a few steps to its right; erosions show a staircase whose steps are each within the noise of a $K=8$ point, so their decisive interval is often wide (9 of the 13 wide intervals are erosions) and their point of no return lies a median of 13 steps after $t^*$ against 2 on cliffs. Curves that start low (a task with $p(x)$ near $0.25$) leave little to localize and their drops are correspondingly small. On some curves two judges' triangles of different lineages sit at the same step; these are the trajectories on which the lineages agree, and they are not confined to cliffs: the shared step is usually the one at which the wrong idea is acted upon, which both lineages recognize, whether or not the value had already moved. Conclusion: the gallery is the evidence behind Table~\ref{tab:rq1} in its rawest form; a reader can verify on any curve that the triangles rarely sit on the shaded interval: a main judge's label lies inside it on 3 of the 47 curves (1 cliffs, 2 erosions).

Across the 47 curves: 17 cliffs and 30 erosions; in 26 the value is flat (within $10\%$ of $\hat V(h_0)$) until the decisive interval, so everything before $t^*$ was harmless, while in 16 at least $30\%$ of the value had already leaked away before the largest drop, the erosion pattern; 13 decisive intervals are wider than one step (bisection stopped where neighbouring drops were indistinguishable at $K=8$); no curve loses the task only at its last step: every failure is decided before the agent stops; 33 show a rise of at least $0.25$ between consecutive raw points, which at $K\le32$ is usually within noise; in 2 of them the two 95\% intervals do not overlap, a genuine recovery that the monotone fit flattens and that is named in the notes below; the point of no return coincides with $t^*$ in 11. On the labels: 3 curves have at least one main-judge label inside the decisive interval, 10 have every label after it, and 5 have every label before it.

How to read a curve for diagnosis: the height at $t=0$ is the task's pass rate under this policy, so a curve that starts low has little to lose and its drops are small by construction; a flat stretch followed by one step to zero (a cliff) means that one decision ended the run, and the steps on the shaded interval are where to look in the transcript; a staircase (an erosion) means the agent spent its chances over many steps, and the decisive interval is only the largest of several losses; a drop that coincides with the first edit means the fix itself was wrong, a drop before it means the diagnosis was wrong before any code changed; the dotted line is the earliest step at which the agent's prospects were certifiably below $\epsilon$, so any label to its right marks something that could no longer matter; a curve whose dotted line is far to the right of its shaded interval is one on which the agent kept a small chance for a long time, typically because a later retry could still have worked. The triangles above the axis are the judges' labels; the question the paper asks is simply whether they sit on the shaded interval, and on most curves they do not.

\clearpage
\onecolumn
\ifdefined\nolinenumbers\nolinenumbers\fi
\paragraph{Figure~\ref{fig:gallery1}.}
\begin{multicols}{2}\scriptsize\begin{itemize}\setlength{\itemsep}{1pt}
\item \textbf{django-11555 s2} (erosion, $C=0.30$; $\hat V(h_0)=0.48$, $T=88$, L1): the decisive step 35 (resolved to one step) comes before the first edit (edit at 44); point of no return at 71 (36 steps later); the raw estimates rise by 0.38 from step 35 (1/8) to 40 (4/8), within the noise of the two samples; first failure signal at 45. Labels: GPT-6.1 Sol 13, GPT-6 Astra 13, GPT-5.5 56, OpenAI o3 44 (2 before it; 2 after it).
\item \textbf{django-11555 s5} (erosion, $C=0.39$; $\hat V(h_0)=0.40$, $T=100$, L2): the decisive step 86 (in an interval of 5 steps) comes after the first edit (edit at 52), with the value flat until then; point of no return at 93 (7 steps later); the raw estimates rise by 0.75 from step 60 (2/8) to 70 (8/8), a genuine recovery (the 95\% intervals do not overlap) that the monotone fit flattens; first failure signal at 54. Labels: GPT-6.1 Sol 12, GPT-6 Astra 12, GPT-5.5 58, OpenAI o3 none (3 before it; 1 ``no mistake'').
\item \textbf{django-11555 s7} (erosion, $C=0.23$; $\hat V(h_0)=0.53$, $T=99$, L1): the decisive step 6 (in an interval of 5 steps) comes before the first edit (edit at 40), with the value flat until then; point of no return at 82 (76 steps later); the raw estimates rise by 0.31 from step 10 (2/8) to 12 (9/16), within the noise of the two samples; first failure signal at 41. Labels: GPT-6.1 Sol 13, GPT-6 Astra 13, GPT-5.5 82, OpenAI o3 82 (4 after it).
\item \textbf{django-12325 s1} (erosion, $C=0.40$; $\hat V(h_0)=0.07$, $T=67$, L1): the decisive step 11 (resolved to one step) comes before the first edit (edit at 30), with the value flat until then; point of no return at 15 (4 steps later); first failure signal at 39. Labels: GPT-6.1 Sol 27, GPT-6 Astra 27, GPT-5.5 none, OpenAI o3 28 (3 after the point of no return; 1 ``no mistake'').
\item \textbf{django-12325 s2} (cliff, $C=0.60$; $\hat V(h_0)=0.06$, $T=90$, L1): the decisive step 2 (resolved to one step) comes before the first edit (edit at 38), with the value flat until then; point of no return at 2 (same step); first failure signal at 46. Labels: GPT-6.1 Sol 38, GPT-6 Astra 38, GPT-5.5 none, OpenAI o3 22 (3 after the point of no return; 1 ``no mistake'').
\item \textbf{django-12325 s6} (cliff, $C=1.00$; $\hat V(h_0)=0.05$, $T=67$, L1): the decisive step 16 (in an interval of 5 steps) comes before the first edit (edit at 31), with the value flat until then; point of no return at 0 (-16 steps later); first failure signal at 36. Labels: GPT-6.1 Sol 28, GPT-6 Astra 28, GPT-5.5 none, OpenAI o3 none (2 after the point of no return; 2 ``no mistake'').
\item \textbf{django-13807 s2} (cliff, $C=0.71$; $\hat V(h_0)=0.44$, $T=71$, L1): the decisive step 1 (in an interval of 5 steps) comes before the first edit (edit at 18), with the value flat until then; point of no return at 15 (14 steps later); first failure signal at 23. Labels: GPT-6.1 Sol 10, GPT-6 Astra 10, GPT-5.5 none, OpenAI o3 18 (2 after it; 1 after the point of no return; 1 ``no mistake'').
\item \textbf{django-13807 s3} (cliff, $C=0.69$; $\hat V(h_0)=0.50$, $T=41$, L0): the decisive step 13 (resolved to one step) comes at the first edit (edit at 13), with 31\% of the value already gone before it; point of no return at 13 (same step); first failure signal at 19. Labels: GPT-6.1 Sol 10, GPT-6 Astra 10, GPT-5.5 none, OpenAI o3 none (2 before it; 2 ``no mistake'').
\item \textbf{django-13807 s7} (erosion, $C=0.40$; $\hat V(h_0)=0.25$, $T=68$, L1): the decisive step 6 (in an interval of 2 steps) comes before the first edit (edit at 16), with the value flat until then; point of no return at 18 (12 steps later); the raw estimates rise by 0.25 from step 0 (1/8) to 5 (3/8), within the noise of the two samples; first failure signal at 46. Labels: GPT-6.1 Sol 8, GPT-6 Astra 8, GPT-5.5 none, OpenAI o3 none (2 after it; 2 ``no mistake'').
\item \textbf{django-15128 s4} (erosion, $C=0.25$; $\hat V(h_0)=0.50$, $T=100$, L2): the decisive step 17 (resolved to one step) comes before the first edit (edit at 65), with 44\% of the value already gone before it; point of no return at 63 (46 steps later); the raw estimates rise by 0.25 from step 17 (0/8) to 20 (2/8), within the noise of the two samples; first failure signal at 78. Labels: GPT-6.1 Sol 11, GPT-6 Astra 11, GPT-5.5 40, OpenAI o3 65 (2 before it; 1 after it; 1 after the point of no return).
\item \textbf{django-15128 s5} (erosion, $C=0.38$; $\hat V(h_0)=0.50$, $T=84$, L1): the decisive step 9 (resolved to one step) comes before the first edit (edit at 38), with the value flat until then; point of no return at 40 (31 steps later); the raw estimates rise by 0.38 from step 7 (4/8) to 8 (7/8), within the noise of the two samples; first failure signal at 42. Labels: GPT-6.1 Sol 10, GPT-6 Astra 10, GPT-5.5 30, OpenAI o3 76 (3 after it; 1 after the point of no return).
\item \textbf{django-15128 s6} (erosion, $C=0.25$; $\hat V(h_0)=0.62$, $T=92$, L1): the decisive step 51 (in an interval of 10 steps) comes before the first edit (edit at 53), with 56\% of the value already gone before it; point of no return at 81 (30 steps later); the raw estimates rise by 0.25 from step 35 (2/8) to 40 (4/8), within the noise of the two samples; no explicit failure signal. Labels: GPT-6.1 Sol 10, GPT-6 Astra 10, GPT-5.5 65, OpenAI o3 10 (3 before it; 1 after it).
\end{itemize}\end{multicols}
\clearpage
\paragraph{Figure~\ref{fig:gallery2}.}
\begin{multicols}{2}\scriptsize\begin{itemize}\setlength{\itemsep}{1pt}
\item \textbf{django-16100 s1} (erosion, $C=0.32$; $\hat V(h_0)=0.46$, $T=63$, L0): the decisive step 39 (resolved to one step) comes after the first edit (edit at 21), with 36\% of the value already gone before it; point of no return at 60 (21 steps later); the raw estimates rise by 0.62 from step 0 (1/8) to 5 (6/8), within the noise of the two samples; first failure signal at 34. Labels: GPT-6.1 Sol 17, GPT-6 Astra 17, GPT-5.5 17, OpenAI o3 21 (4 before it).
\item \textbf{django-16100 s2} (cliff, $C=1.00$; $\hat V(h_0)=0.36$, $T=58$, L0): the decisive step 50 (resolved to one step) comes after the first edit (edit at 24), with the value flat until then; point of no return at 50 (same step); the raw estimates rise by 0.50 from step 35 (1/8) to 40 (5/8), within the noise of the two samples; first failure signal at 34. Labels: GPT-6.1 Sol 2, GPT-6 Astra 2, GPT-5.5 none, OpenAI o3 24 (3 before it; 1 ``no mistake'').
\item \textbf{django-16100 s6} (erosion, $C=0.39$; $\hat V(h_0)=0.32$, $T=68$, L1): the decisive step 48 (resolved to one step) comes after the first edit (edit at 21), with 69\% of the value already gone before it; point of no return at 48 (same step); the raw estimates rise by 0.25 from step 17 (2/8) to 19 (16/32), within the noise of the two samples; first failure signal at 28. Labels: GPT-6.1 Sol 20, GPT-6 Astra 20, GPT-5.5 21, OpenAI o3 21 (4 before it).
\item \textbf{django-16938 s2} (erosion, $C=0.35$; $\hat V(h_0)=0.50$, $T=100$, L2): the decisive step 11 (resolved to one step) comes before the first edit (edit at 35), with the value flat until then; point of no return at 50 (39 steps later); the raw estimates rise by 0.25 from step 30 (2/8) to 35 (4/8), within the noise of the two samples; first failure signal at 39. Labels: GPT-6.1 Sol 11, GPT-6 Astra 11, GPT-5.5 99, OpenAI o3 none (2 inside the decisive interval; 1 after the point of no return; 1 ``no mistake'').
\item \textbf{django-16938 s4} (erosion, $C=0.33$; $\hat V(h_0)=0.56$, $T=65$, L1): the decisive step 11 (resolved to one step) comes before the first edit (edit at 19), with 61\% of the value already gone before it; point of no return at 11 (same step); first failure signal at 20. Labels: GPT-6.1 Sol 11, GPT-6 Astra 11, GPT-5.5 16, OpenAI o3 none (2 inside the decisive interval; 1 after the point of no return; 1 ``no mistake'').
\item \textbf{django-16938 s5} (cliff, $C=0.71$; $\hat V(h_0)=0.50$, $T=100$, L2): the decisive step 95 (in an interval of 2 steps) comes after the first edit (edit at 68), with the value flat until then; point of no return at 97 (2 steps later); the raw estimates rise by 0.50 from step 90 (3/8) to 94 (7/8), within the noise of the two samples; no explicit failure signal. Labels: GPT-6.1 Sol 8, GPT-6 Astra 8, GPT-5.5 none, OpenAI o3 none (2 before it; 2 ``no mistake'').
\item \textbf{matplotlib-22865 s2} (erosion, $C=0.32$; $\hat V(h_0)=0.39$, $T=54$, L0): the decisive step 6 (in an interval of 5 steps) comes before the first edit (edit at 32), with the value flat until then; point of no return at 30 (24 steps later); no explicit failure signal. Labels: GPT-6.1 Sol 16, GPT-6 Astra 16, GPT-5.5 28, OpenAI o3 none (3 after it; 1 ``no mistake'').
\item \textbf{matplotlib-22865 s3} (erosion, $C=0.39$; $\hat V(h_0)=0.50$, $T=40$, L0): the decisive step 19 (resolved to one step) comes before the first edit (edit at 25), with 46\% of the value already gone before it; point of no return at 22 (3 steps later); the raw estimates rise by 0.29 from step 10 (1/8) to 15 (10/24), within the noise of the two samples; no explicit failure signal. Labels: GPT-6.1 Sol 20, GPT-6 Astra 20, GPT-5.5 20, OpenAI o3 18 (1 before it; 3 after it).
\item \textbf{matplotlib-22865 s7} (cliff, $C=0.60$; $\hat V(h_0)=0.31$, $T=51$, L0): the decisive step 12 (resolved to one step) comes before the first edit (edit at 31), with the value flat until then; point of no return at 20 (8 steps later); the raw estimates rise by 0.25 from step 0 (1/8) to 5 (3/8), within the noise of the two samples; first failure signal at 35. Labels: GPT-6.1 Sol 21, GPT-6 Astra 21, GPT-5.5 23, OpenAI o3 21 (4 after the point of no return).
\item \textbf{matplotlib-25960 s0} (cliff, $C=0.50$; $\hat V(h_0)=0.12$, $T=58$, L0): the decisive step 22 (in an interval of 4 steps) comes before the first edit (edit at 29), with 75\% of the value already gone before it; point of no return at 0 (-22 steps later); the raw estimates rise by 0.25 from step 0 (0/24) to 5 (2/8), within the noise of the two samples; first failure signal at 38. Labels: GPT-6.1 Sol 29, GPT-6 Astra 17, GPT-5.5 21, OpenAI o3 29 (2 before it; 2 after the point of no return).
\item \textbf{matplotlib-25960 s3} (erosion, $C=0.48$; $\hat V(h_0)=0.12$, $T=64$, L0): the decisive step 11 (in an interval of 5 steps) comes before the first edit (edit at 21), with the value flat until then; point of no return at 25 (14 steps later); first failure signal at 29. Labels: GPT-6.1 Sol 16, GPT-6 Astra 21, GPT-5.5 16, OpenAI o3 21 (4 after it).
\item \textbf{matplotlib-25960 s5} (cliff, $C=0.95$; $\hat V(h_0)=0.18$, $T=57$, L0): the decisive step 6 (resolved to one step) comes before the first edit (edit at 23), with the value flat until then; point of no return at 6 (same step); first failure signal at 31. Labels: GPT-6.1 Sol 17, GPT-6 Astra 17, GPT-5.5 17, OpenAI o3 49 (4 after the point of no return).
\end{itemize}\end{multicols}
\clearpage
\paragraph{Figure~\ref{fig:gallery3}.}
\begin{multicols}{2}\scriptsize\begin{itemize}\setlength{\itemsep}{1pt}
\item \textbf{seaborn-3069 s2} (cliff, $C=0.53$; $\hat V(h_0)=0.50$, $T=100$, L2): the decisive step 14 (resolved to one step) comes before the first edit (edit at 44), with the value flat until then; point of no return at 91 (77 steps later); the raw estimates rise by 0.25 from step 0 (2/8) to 5 (4/8), within the noise of the two samples; first failure signal at 56. Labels: GPT-6.1 Sol 14, GPT-6 Astra 14, GPT-5.5 14, OpenAI o3 45 (3 inside the decisive interval; 1 after it).
\item \textbf{seaborn-3069 s3} (erosion, $C=0.42$; $\hat V(h_0)=0.43$, $T=100$, L2): the decisive step 2 (resolved to one step) comes before the first edit (edit at 59), with the value flat until then; point of no return at 30 (28 steps later); the raw estimates rise by 0.50 from step 5 (1/8) to 10 (5/8), within the noise of the two samples; first failure signal at 82. Labels: GPT-6.1 Sol 16, GPT-6 Astra 16, GPT-5.5 96, OpenAI o3 59 (2 after it; 2 after the point of no return).
\item \textbf{seaborn-3069 s5} (erosion, $C=0.39$; $\hat V(h_0)=0.54$, $T=100$, L2): the decisive step 62 (resolved to one step) comes after the first edit (edit at 29), with 80\% of the value already gone before it; point of no return at 62 (same step); the raw estimates rise by 0.25 from step 0 (1/8) to 5 (3/8), within the noise of the two samples; first failure signal at 30. Labels: GPT-6.1 Sol 16, GPT-6 Astra 16, GPT-5.5 29, OpenAI o3 29 (4 before it).
\item \textbf{requests-6028 s0} (cliff, $C=0.50$; $\hat V(h_0)=0.12$, $T=64$, L0): the decisive step 26 (resolved to one step) comes before the first edit (edit at 38), with 50\% of the value already gone before it; point of no return at 26 (same step); no explicit failure signal. Labels: GPT-6.1 Sol 18, GPT-6 Astra 18, GPT-5.5 18, OpenAI o3 none (3 before it; 1 ``no mistake'').
\item \textbf{requests-6028 s1} (erosion, $C=0.35$; $\hat V(h_0)=0.24$, $T=57$, L0): the decisive step 13 (resolved to one step) comes before the first edit (edit at 23), with the value flat until then; point of no return at 15 (2 steps later); the raw estimates rise by 0.33 from step 0 (1/24) to 5 (3/8), within the noise of the two samples; first failure signal at 39. Labels: GPT-6.1 Sol 14, GPT-6 Astra 14, GPT-5.5 14, OpenAI o3 21 (3 after it; 1 after the point of no return).
\item \textbf{requests-6028 s4} (cliff, $C=0.65$; $\hat V(h_0)=0.29$, $T=61$, L0): the decisive step 25 (resolved to one step) comes before the first edit (edit at 35), with the value flat until then; point of no return at 29 (4 steps later); the raw estimates rise by 0.46 from step 5 (1/24) to 10 (4/8), a genuine recovery (the 95\% intervals do not overlap) that the monotone fit flattens; no explicit failure signal. Labels: GPT-6.1 Sol 24, GPT-6 Astra 24, GPT-5.5 35, OpenAI o3 35 (2 before it; 2 after the point of no return).
\item \textbf{scikit-learn-10908 s3} (cliff, $C=0.65$; $\hat V(h_0)=1.00$, $T=47$, L0): the decisive step 19 (resolved to one step) comes before the first edit (edit at 22), with 35\% of the value already gone before it; point of no return at 19 (same step); first failure signal at 27. Labels: GPT-6.1 Sol none, GPT-6 Astra none, GPT-5.5 none, OpenAI o3 none (4 ``no mistake'').
\item \textbf{sphinx-8265 s0} (cliff, $C=0.65$; $\hat V(h_0)=0.50$, $T=64$, L0): the decisive step 16 (resolved to one step) comes before the first edit (edit at 23); point of no return at 22 (6 steps later); first failure signal at 38. Labels: GPT-6.1 Sol 19, GPT-6 Astra 19, GPT-5.5 19, OpenAI o3 none (3 after it; 1 ``no mistake'').
\item \textbf{sphinx-8265 s5} (erosion, $C=0.25$; $\hat V(h_0)=1.00$, $T=75$, L1): the decisive step 18 (resolved to one step) comes before the first edit (edit at 26); point of no return at 21 (3 steps later); the raw estimates rise by 0.38 from step 10 (4/8) to 15 (7/8), within the noise of the two samples; first failure signal at 41. Labels: GPT-6.1 Sol 19, GPT-6 Astra 19, GPT-5.5 26, OpenAI o3 73 (2 after it; 2 after the point of no return).
\item \textbf{sphinx-8593 s2} (erosion, $C=0.20$; $\hat V(h_0)=0.88$, $T=100$, L2): the decisive step 51 (resolved to one step) comes after the first edit (edit at 48), with 40\% of the value already gone before it; point of no return at 99 (48 steps later); the raw estimates rise by 0.50 from step 30 (3/8) to 35 (7/8), within the noise of the two samples; first failure signal at 90. Labels: GPT-6.1 Sol 48, GPT-6 Astra 48, GPT-5.5 99, OpenAI o3 48 (3 before it; 1 after it).
\item \textbf{sphinx-8593 s6} (erosion, $C=0.33$; $\hat V(h_0)=0.65$, $T=100$, L2): the decisive step 98 (resolved to one step) comes after the first edit (edit at 36), with 67\% of the value already gone before it; point of no return at 76 (-22 steps later); the raw estimates rise by 0.50 from step 94 (4/8) to 96 (8/8), within the noise of the two samples; first failure signal at 59. Labels: GPT-6.1 Sol 76, GPT-6 Astra 76, GPT-5.5 none, OpenAI o3 none (2 before it; 2 ``no mistake'').
\item \textbf{sphinx-9591 s2} (cliff, $C=0.50$; $\hat V(h_0)=0.50$, $T=100$, L2): the decisive step 48 (resolved to one step) comes before the first edit (edit at 50), with the value flat until then; point of no return at 67 (19 steps later); the raw estimates rise by 0.38 from step 20 (2/8) to 25 (5/8), within the noise of the two samples; first failure signal at 52. Labels: GPT-6.1 Sol 7, GPT-6 Astra 7, GPT-5.5 83, OpenAI o3 83 (2 before it; 2 after the point of no return).
\end{itemize}\end{multicols}
\clearpage
\paragraph{Figure~\ref{fig:gallery4}.}
\begin{multicols}{2}\scriptsize\begin{itemize}\setlength{\itemsep}{1pt}
\item \textbf{sphinx-9591 s4} (erosion, $C=0.33$; $\hat V(h_0)=0.38$, $T=100$, L2): the decisive step 28 (resolved to one step) comes before the first edit (edit at 48), with the value flat until then; point of no return at 70 (42 steps later); the raw estimates rise by 0.42 from step 0 (5/24) to 5 (5/8), within the noise of the two samples; first failure signal at 58. Labels: GPT-6.1 Sol 8, GPT-6 Astra 8, GPT-5.5 48, OpenAI o3 48 (2 before it; 2 after it).
\item \textbf{sphinx-9591 s5} (erosion, $C=0.31$; $\hat V(h_0)=0.54$, $T=100$, L2): the decisive step 8 (in an interval of 3 steps) comes before the first edit (edit at 76), with the value flat until then; point of no return at 47 (39 steps later); the raw estimates rise by 0.25 from step 6 (7/16) to 7 (11/16), within the noise of the two samples; first failure signal at 99. Labels: GPT-6.1 Sol 7, GPT-6 Astra 7, GPT-5.5 none, OpenAI o3 76 (2 before it; 1 after the point of no return; 1 ``no mistake'').
\item \textbf{sympy-13877 s1} (erosion, $C=0.20$; $\hat V(h_0)=0.62$, $T=58$, L0): the decisive step 6 (resolved to one step) comes before the first edit (edit at 34), with the value flat until then; point of no return at 28 (22 steps later); the raw estimates rise by 0.25 from step 0 (4/8) to 5 (6/8), within the noise of the two samples; no explicit failure signal. Labels: GPT-6.1 Sol 7, GPT-6 Astra 7, GPT-5.5 34, OpenAI o3 34 (2 after it; 2 after the point of no return).
\item \textbf{sympy-13877 s5} (erosion, $C=0.35$; $\hat V(h_0)=0.79$, $T=44$, L0): the decisive step 32 (resolved to one step) comes after the first edit (edit at 20), with 61\% of the value already gone before it; point of no return at 32 (same step); the raw estimates rise by 0.25 from step 0 (5/8) to 5 (7/8), within the noise of the two samples; first failure signal at 36. Labels: GPT-6.1 Sol 16, GPT-6 Astra 16, GPT-5.5 16, OpenAI o3 20 (4 before it).
\item \textbf{sympy-15875 s1} (erosion, $C=0.49$; $\hat V(h_0)=0.57$, $T=57$, L0): the decisive step 15 (resolved to one step) comes before the first edit (edit at 17); point of no return at 16 (1 steps later); the raw estimates rise by 0.62 from step 0 (1/8) to 5 (6/8), within the noise of the two samples; first failure signal at 42. Labels: GPT-6.1 Sol 17, GPT-6 Astra 17, GPT-5.5 27, OpenAI o3 27 (4 after the point of no return).
\item \textbf{sympy-15875 s4} (cliff, $C=0.50$; $\hat V(h_0)=0.54$, $T=100$, L2): the decisive step 35 (resolved to one step) comes before the first edit (edit at 45), with the value flat until then; point of no return at 88 (53 steps later); the raw estimates rise by 0.44 from step 58 (7/16) to 60 (7/8), within the noise of the two samples; first failure signal at 60. Labels: GPT-6.1 Sol 12, GPT-6 Astra 12, GPT-5.5 55, OpenAI o3 58 (2 before it; 2 after it).
\item \textbf{sympy-19040 s0} (erosion, $C=0.33$; $\hat V(h_0)=0.38$, $T=100$, L2): the decisive step 1 (in an interval of 2 steps) comes before the first edit (edit at 74), with the value flat until then; point of no return at 4 (3 steps later); first failure signal at 84. Labels: GPT-6.1 Sol 62, GPT-6 Astra 62, GPT-5.5 62, OpenAI o3 none (3 after the point of no return; 1 ``no mistake'').
\item \textbf{sympy-19040 s3} (cliff, $C=0.53$; $\hat V(h_0)=0.60$, $T=100$, L2): the decisive step 42 (resolved to one step) comes before the first edit (edit at 46), with the value flat until then; point of no return at 52 (10 steps later); the raw estimates rise by 0.25 from step 0 (2/8) to 5 (4/8), within the noise of the two samples; first failure signal at 89. Labels: GPT-6.1 Sol 11, GPT-6 Astra 11, GPT-5.5 35, OpenAI o3 46 (3 before it; 1 after it).
\item \textbf{sympy-19040 s7} (erosion, $C=0.45$; $\hat V(h_0)=0.15$, $T=100$, L2): the decisive step 66 (resolved to one step) comes after the first edit (edit at 54), with 55\% of the value already gone before it; point of no return at 31 (-35 steps later); first failure signal at 58. Labels: GPT-6.1 Sol 24, GPT-6 Astra 24, GPT-5.5 24, OpenAI o3 91 (3 before it; 1 after the point of no return).
\item \textbf{sympy-19783 s1} (erosion, $C=0.30$; $\hat V(h_0)=0.82$, $T=100$, L2): the decisive step 97 (resolved to one step) comes after the first edit (edit at 41), with 70\% of the value already gone before it; point of no return at 97 (same step); the raw estimates rise by 0.62 from step 60 (1/8) to 70 (6/8), within the noise of the two samples; first failure signal at 82. Labels: GPT-6.1 Sol 24, GPT-6 Astra 22, GPT-5.5 39, OpenAI o3 75 (4 before it).
\item \textbf{sympy-19783 s5} (erosion, $C=0.21$; $\hat V(h_0)=0.75$, $T=100$, L2): the decisive step 46 (in an interval of 5 steps) comes after the first edit (edit at 29); point of no return at 90 (44 steps later); the raw estimates rise by 0.62 from step 5 (2/8) to 10 (7/8), within the noise of the two samples; first failure signal at 89. Labels: GPT-6.1 Sol 26, GPT-6 Astra 26, GPT-5.5 36, OpenAI o3 76 (3 before it; 1 after it).
\end{itemize}\end{multicols}
\ifdefined\linenumbers\linenumbers\fi
\twocolumn

%% file: appendix_tables_arm.tex
\clearpage
\section{The second policy, Qwen3.5-9B}
\label{arm:app:all}

The tables and curves of the preceding sections, for the second policy's frozen set; every quantity is defined as for the second policy, and the main-policy table that each one extends is named in its caption. This policy fails the cap-hit check (\S\ref{sec:checks}): its curves are reported as a replication of the label results, and its restart evaluation is to be read with that caveat.

\subsection{The base-run pool}
\label{arm:app:pool}

Table~\ref{arm:tab:pool} summarizes the base runs of the second policy (left) and the distribution of tasks by number of successes (right). The distribution is bimodal: 14 tasks succeed in all eight runs and 65 in none, so only 44 tasks fall in the band of 2--6 successes where a failed run is informative; pass@8 is 0.09 against pass@1 0.34. Two consequences follow. The study set is bounded by the band (44 tasks here) because failures outside it are uninformative: on a doomed task ($\le1$ success) $V(h_0)$ is already near zero and no step can lose much, and on a saturated task a failed run is a rare draw whose continuations mostly succeed, so the curve is flat near one until late. The band is therefore where cause and symptom can be told apart, and every trajectory in the study comes from it. Second, the base runs also give $R(x,b)$ (Definition~5): with eight runs per task, $\hat R$ is a step function with increments of $1/8$, which is the resolution of the offline restart evaluation and one reason for the online check (Appendix~\ref{app:online}). 48\% of base runs hit the step cap and 603 submitted an empty patch (counted as failures without a harness run). Conclusion: the informative failures are those of the 44 band tasks of 150, so a study of where failures are caused must select by pass rate first; the 106 tasks outside the band contribute to $R(x,b)$ and to the decoys' calibration but not to the curves.

\begin{table}[htbp]
\centering
{\footnotesize\setlength{\tabcolsep}{3pt}
\begin{tabular}[t]{lr}
\toprule
Quantity & Value \\
\midrule
Tasks, runs & 150, 1200 \\
pass@1 & 0.340 \\
pass@2 / @4 / @8 & 0.25 / 0.16 / 0.09 \\
Doomed (0--1 of 8) & 0.52 \\
Saturated (7--8 of 8) & 0.19 \\
Band (2--6 of 8) & 0.29 (44 tasks) \\
Cap hits & 48.5\% \\
Empty patches & 603 \\
Evaluation errors & 48 \\
\bottomrule
\end{tabular}}\hfill
{\footnotesize\setlength{\tabcolsep}{3pt}
\begin{tabular}[t]{cr}
\toprule
Successes & Tasks \\
\midrule
0 & 65 \\
1 & 13 \\
2 & 9 \\
3 & 6 \\
4 & 7 \\
5 & 11 \\
6 & 11 \\
7 & 14 \\
8 & 14 \\
\bottomrule
\end{tabular}}
\caption{Base-run pool of the second policy (left) and tasks by number of successes out of 8 (right).}
\label{arm:tab:pool}
\end{table}\vspace{-5pt}

\FloatBarrier
\subsection{All trajectories}
\label{arm:app:trajectories}

Table~\ref{arm:tab:alltraj} gives, for each of the 60 trajectories of the second policy's frozen set, the quantities the curve yields: Stage~2 stratum (L0 and L1: the base run submitted a patch, in the lower or upper length tercile; L2: the base run hit the step cap), length $T$, value at $h_0$ (isotonic fit), shape and concentration $C$, the decisive step $t^*$ with its interval, the certified point of no return, the first repository edit, the first explicit failure signal, and the number of continuations. Both endpoints of the interval are estimated at $K\ge32$ in 23 trajectories ($\dagger$). Tables~\ref{arm:tab:alllabels} and \ref{arm:tab:alllabels-2} list every source's first-mistake label next to $t^*$ and $t_{\text{pnr}}$; ``none'' is a ``no mistake'' answer and $\cdot$ means the source did not label that trajectory (GPT-6.1 Sol, GPT-6 Astra, GPT-5.5 and OpenAI o3 are the main judges, GPT-6 Luna the contrast model, Claude Opus 5.5 the annotator-protocol model on its subset). Reading the table: the set has 16 L0, 0 L1 and 44 L2 trajectories, 38 to 100 steps long (median 100); the fitted value at $h_0$ runs from 0.06 to 1.00 (median 0.58), around a base pass rate that lies in the band $[0.25,0.75]$ by construction (the fitted value differs from it by the sampling noise of eight continuations and by the monotone fit). The decisive step sits at a median of 27\% of the trajectory; 9 trajectories lose the task within the first five steps and 9 only at the last step, where the agent submits or runs out of budget with the value still above zero. The point of no return follows $t^*$ by a median of 19 steps (upper quartile 38); where the gap is long the curve is an erosion whose value leaks slowly after the largest drop. 36 of 60 trajectories show an explicit failure signal after the first edit; on the others the agent fails silently, usually by submitting a patch that does not resolve the issue. In Tables~\ref{arm:tab:alllabels} and \ref{arm:tab:alllabels-2} the ``no mistake'' answers cluster by lineage: GPT-6.1 Sol 1, GPT-6 Astra 1, GPT-5.5 6, OpenAI o3 15, GPT-6 Luna (contrast) 5. Conclusion: the table is the paper's raw material; every count in \S\ref{sec:results} can be recomputed from it, and a reader who wants to check a claim about one trajectory (for instance that the judges' labels fall after $t_{\text{pnr}}$) can do so by comparing the label columns with the $t^*$ and $t_{\text{pnr}}$ columns.

\begin{table*}[tp]
\centering
{\scriptsize\setlength{\tabcolsep}{3pt}
\begin{tabular}[t]{lccccccccccc}
\toprule
Trajectory & Str. & $T$ & $\hat V(h_0)$ & Shape & $C$ & $t^*$ & $(a,b]$ & $t_{\text{pnr}}$ & $t_{\text{edit}}$ & $t_{\text{sym}}$ & $n$ \\
\midrule
django-11333 s6 & L2 & 100 & 0.38 & erosion & 0.38 & 7 & (6,7]$^\dagger$ & 36 & 22 & 27 & 288 \\
django-11451 s0 & L0 & 81 & 0.88 & erosion & 0.14 & 1 & (0,5] & 71 & 19 & 22 & 352 \\
django-11451 s2 & L0 & 76 & 0.88 & cliff & 0.75 & 76 & (75,76]$^\dagger$ & 76 & 21 & 27 & 264 \\
django-11999 s0 & L2 & 100 & 0.56 & erosion & 0.26 & 71 & (70,71] & 60 & 28 & 35 & 448 \\
django-11999 s1 & L2 & 100 & 0.36 & erosion & 0.35 & 40 & (39,40]$^\dagger$ & 20 & 43 & 53 & 376 \\
django-11999 s6 & L2 & 100 & 0.44 & erosion & 0.38 & 21 & (20,21] & 48 & -- & -- & 416 \\
django-12713 s3 & L2 & 100 & 0.84 & erosion & 0.22 & 34 & (33,35] & 35 & -- & -- & 296 \\
django-13028 s2 & L2 & 100 & 0.25 & erosion & 0.25 & 11 & (10,12] & 25 & -- & -- & 384 \\
django-13028 s5 & L0 & 87 & 0.50 & cliff & 1.00 & 87 & (86,87]$^\dagger$ & 87 & 39 & 58 & 296 \\
django-13028 s6 & L2 & 100 & 0.06 & cliff & 0.57 & 12 & (11,12] & 10 & -- & -- & 288 \\
django-13089 s5 & L0 & 79 & 0.94 & cliff & 0.57 & 79 & (78,79]$^\dagger$ & 79 & 23 & 58 & 232 \\
django-13551 s4 & L0 & 71 & 0.69 & erosion & 0.33 & 23 & (22,23] & 42 & 42 & -- & 296 \\
django-13821 s6 & L2 & 100 & 0.62 & erosion & 0.28 & 37 & (36,37] & 56 & -- & -- & 408 \\
django-13821 s7 & L2 & 100 & 0.82 & erosion & 0.15 & 71 & (70,74] & 89 & 55 & 83 & 336 \\
django-13837 s0 & L2 & 100 & 0.59 & cliff & 0.50 & 1 & (0,1]$^\dagger$ & 20 & -- & -- & 384 \\
django-13837 s2 & L2 & 100 & 0.43 & erosion & 0.29 & 26 & (25,27] & 64 & 52 & 64 & 456 \\
django-13933 s0 & L2 & 100 & 0.81 & erosion & 0.31 & 11 & (10,12] & 75 & 32 & 39 & 392 \\
django-13933 s6 & L2 & 100 & 1.00 & erosion & 0.14 & 11 & (10,12] & 61 & 63 & 77 & 384 \\
django-16116 s0 & L0 & 79 & 0.68 & cliff & 0.60 & 79 & (78,79]$^\dagger$ & 79 & 37 & 41 & 280 \\
django-16116 s4 & L2 & 100 & 0.50 & erosion & 0.40 & 63 & (62,65] & 90 & 43 & 61 & 416 \\
django-16139 s0 & L2 & 100 & 0.67 & erosion & 0.37 & 83 & (82,83] & 91 & 51 & 59 & 352 \\
django-16139 s1 & L2 & 100 & 0.60 & erosion & 0.32 & 5 & (4,5] & 66 & 87 & 94 & 432 \\
django-16139 s3 & L2 & 100 & 0.50 & erosion & 0.35 & 24 & (23,24] & 80 & -- & -- & 376 \\
django-16899 s0 & L2 & 100 & 0.68 & cliff & 0.56 & 60 & (59,60]$^\dagger$ & 70 & 35 & 53 & 400 \\
\bottomrule
\end{tabular}}
\caption{Curve quantities of every trajectory of the second policy's frozen set (1 of 3); $\dagger$: both endpoints of the interval at $K\ge32$.}
\label{arm:tab:alltraj}
\end{table*}\vspace{-5pt}
\begin{table*}[tp]
\centering
{\scriptsize\setlength{\tabcolsep}{3pt}
\begin{tabular}[t]{lccccccccccc}
\toprule
Trajectory & Str. & $T$ & $\hat V(h_0)$ & Shape & $C$ & $t^*$ & $(a,b]$ & $t_{\text{pnr}}$ & $t_{\text{edit}}$ & $t_{\text{sym}}$ & $n$ \\
\midrule
django-16899 s3 & L0 & 86 & 0.72 & cliff & 0.70 & 39 & (38,39]$^\dagger$ & 39 & 39 & 45 & 368 \\
django-16899 s7 & L0 & 81 & 0.88 & cliff & 0.75 & 35 & (34,35]$^\dagger$ & 35 & 37 & -- & 352 \\
xarray-3305 s1 & L2 & 100 & 0.81 & erosion & 0.15 & 6 & (5,6] & 49 & 48 & 63 & 464 \\
xarray-3305 s5 & L2 & 100 & 0.81 & erosion & 0.15 & 26 & (25,26]$^\dagger$ & 45 & 44 & 51 & 384 \\
xarray-3305 s7 & L2 & 100 & 0.56 & erosion & 0.30 & 6 & (5,9] & 46 & 34 & 37 & 256 \\
xarray-4695 s3 & L2 & 100 & 0.16 & cliff & 0.53 & 34 & (33,35] & 35 & 56 & -- & 312 \\
xarray-4695 s4 & L2 & 100 & 0.11 & cliff & 0.79 & 32 & (31,32] & 35 & -- & -- & 272 \\
xarray-7393 s0 & L2 & 100 & 0.35 & erosion & 0.36 & 7 & (6,7] & 36 & 82 & 86 & 408 \\
xarray-7393 s4 & L2 & 100 & 0.64 & erosion & 0.21 & 46 & (45,49] & 96 & 50 & 53 & 376 \\
pytest-5631 s0 & L2 & 100 & 0.75 & cliff & 0.58 & 1 & (0,2] & 25 & -- & -- & 320 \\
pytest-5631 s6 & L0 & 80 & 0.56 & erosion & 0.31 & 1 & (0,1]$^\dagger$ & 40 & 46 & 50 & 392 \\
pytest-5631 s7 & L2 & 100 & 0.62 & erosion & 0.30 & 1 & (0,1] & 51 & 73 & 84 & 416 \\
pytest-6202 s3 & L2 & 100 & 0.38 & erosion & 0.24 & 11 & (10,15] & 25 & 83 & 86 & 296 \\
pytest-6202 s7 & L2 & 100 & 0.47 & erosion & 0.38 & 57 & (56,57] & 70 & 74 & 76 & 400 \\
pytest-7432 s0 & L2 & 100 & 0.46 & erosion & 0.40 & 17 & (16,17]$^\dagger$ & 46 & 17 & 32 & 344 \\
pytest-7521 s6 & L2 & 100 & 0.62 & erosion & 0.25 & 28 & (27,28] & 30 & 80 & 81 & 336 \\
pytest-7982 s5 & L2 & 100 & 0.72 & erosion & 0.31 & 57 & (56,57] & 71 & 40 & 42 & 408 \\
scikit-learn-13439 s2 & L0 & 57 & 0.95 & cliff & 0.72 & 57 & (56,57]$^\dagger$ & 57 & 23 & 26 & 184 \\
scikit-learn-13439 s4 & L0 & 53 & 1.00 & cliff & 0.91 & 53 & (52,53]$^\dagger$ & 53 & 20 & -- & 176 \\
scikit-learn-14983 s2 & L2 & 100 & 0.55 & erosion & 0.38 & 83 & (82,83]$^\dagger$ & 90 & 27 & -- & 352 \\
scikit-learn-15100 s6 & L0 & 38 & 1.00 & cliff & 0.75 & 38 & (37,38]$^\dagger$ & 38 & 19 & 23 & 192 \\
sphinx-11445 s3 & L0 & 90 & 0.40 & erosion & 0.35 & 33 & (32,33]$^\dagger$ & 46 & 63 & 68 & 320 \\
sphinx-11445 s6 & L2 & 100 & 0.28 & erosion & 0.30 & 3 & (2,3]$^\dagger$ & 25 & 83 & -- & 368 \\
sphinx-7757 s5 & L0 & 65 & 0.91 & cliff & 1.00 & 65 & (64,65]$^\dagger$ & 65 & 27 & -- & 184 \\
\bottomrule
\end{tabular}}
\caption{Curve quantities of every trajectory of the second policy's frozen set (2 of 3); $\dagger$: both endpoints of the interval at $K\ge32$.}
\label{arm:tab:alltraj-2}
\end{table*}\vspace{-5pt}
\begin{table*}[tp]
\centering
{\scriptsize\setlength{\tabcolsep}{3pt}
\begin{tabular}[t]{lccccccccccc}
\toprule
Trajectory & Str. & $T$ & $\hat V(h_0)$ & Shape & $C$ & $t^*$ & $(a,b]$ & $t_{\text{pnr}}$ & $t_{\text{edit}}$ & $t_{\text{sym}}$ & $n$ \\
\midrule
sphinx-8459 s0 & L2 & 100 & 0.25 & cliff & 0.50 & 1 & (0,4] & 10 & 61 & -- & 384 \\
sphinx-9320 s7 & L0 & 40 & 0.85 & cliff & 1.00 & 40 & (39,40]$^\dagger$ & 40 & 20 & -- & 128 \\
sphinx-9658 s1 & L0 & 85 & 0.26 & cliff & 0.71 & 26 & (25,26] & 60 & 42 & -- & 424 \\
sphinx-9658 s3 & L2 & 100 & 0.29 & erosion & 0.29 & 6 & (5,6]$^\dagger$ & 50 & 85 & -- & 288 \\
sphinx-9658 s6 & L2 & 100 & 0.47 & erosion & 0.48 & 36 & (35,37] & 100 & 55 & -- & 360 \\
sympy-13615 s1 & L2 & 100 & 0.38 & erosion & 0.33 & 6 & (5,6] & 61 & 30 & -- & 416 \\
sympy-13615 s3 & L2 & 100 & 0.32 & erosion & 0.39 & 19 & (18,20] & 49 & 74 & 92 & 376 \\
sympy-13615 s6 & L2 & 100 & 0.31 & erosion & 0.30 & 9 & (8,9] & 41 & 61 & 65 & 376 \\
sympy-19783 s0 & L2 & 100 & 0.50 & erosion & 0.25 & 6 & (5,7] & 30 & 33 & 35 & 336 \\
sympy-19783 s7 & L2 & 100 & 0.50 & cliff & 0.50 & 1 & (0,2] & 16 & 43 & 75 & 432 \\
sympy-22714 s0 & L2 & 100 & 0.62 & erosion & 0.25 & 10 & (9,10]$^\dagger$ & 35 & -- & -- & 392 \\
sympy-22714 s5 & L2 & 100 & 0.62 & erosion & 0.33 & 17 & (16,19] & 70 & -- & -- & 376 \\
\bottomrule
\end{tabular}}
\caption{Curve quantities of every trajectory of the second policy's frozen set (3 of 3); $\dagger$: both endpoints of the interval at $K\ge32$.}
\label{arm:tab:alltraj-3}
\end{table*}\vspace{-5pt}
\begin{table*}[tp]
\centering
{\scriptsize\setlength{\tabcolsep}{3pt}
\begin{tabular}[t]{lccccccc}
\toprule
 &  &  & GPT-6.1 & GPT-6 & GPT-5.5 & OpenAI & GPT-6 \\ Trajectory & $t^*$ & $t_{\text{pnr}}$ & Sol & Astra &  & o3 & Luna \\
\midrule
django-11333 s6 & 7 & 36 & 7 & 7 & 7 & 22 & 7 \\
django-11451 s0 & 1 & 71 & 19 & 17 & 19 & 19 & 19 \\
django-11451 s2 & 76 & 76 & 21 & 21 & 21 & 21 & 21 \\
django-11999 s0 & 71 & 60 & 28 & 15 & 99 & 99 & 28 \\
django-11999 s1 & 40 & 20 & 11 & 11 & 43 & 43 & 11 \\
django-11999 s6 & 21 & 48 & 9 & 9 & 24 & 38 & 11 \\
django-12713 s3 & 34 & 35 & 7 & 7 & 32 & 33 & 24 \\
django-13028 s2 & 11 & 25 & 2 & 2 & 32 & none & 11 \\
django-13028 s5 & 87 & 87 & 24 & 24 & 24 & none & 60 \\
django-13028 s6 & 12 & 10 & 10 & 10 & 10 & 59 & 10 \\
django-13089 s5 & 79 & 79 & 23 & 10 & 23 & 23 & 23 \\
django-13551 s4 & 23 & 42 & 42 & 42 & 42 & none & 42 \\
django-13821 s6 & 37 & 56 & 31 & 31 & 31 & 57 & 31 \\
django-13821 s7 & 71 & 89 & 75 & 75 & 75 & none & 81 \\
django-13837 s0 & 1 & 20 & 8 & 8 & 8 & 14 & 8 \\
django-13837 s2 & 26 & 64 & 15 & 14 & 28 & 60 & 32 \\
django-13933 s0 & 11 & 75 & 20 & 20 & 32 & 32 & 32 \\
django-13933 s6 & 11 & 61 & 13 & 13 & 13 & 63 & 18 \\
django-16116 s0 & 79 & 79 & 13 & 13 & 11 & none & 13 \\
django-16116 s4 & 63 & 90 & 34 & 34 & 43 & 43 & 43 \\
django-16139 s0 & 83 & 91 & 12 & 12 & none & none & 12 \\
django-16139 s1 & 5 & 66 & 38 & 38 & 84 & 86 & 38 \\
django-16139 s3 & 24 & 80 & 21 & 21 & 21 & 21 & 21 \\
django-16899 s0 & 60 & 70 & 22 & 22 & 43 & 43 & none \\
django-16899 s3 & 39 & 39 & 21 & 39 & 39 & none & none \\
django-16899 s7 & 35 & 35 & 37 & 20 & 43 & 37 & 44 \\
xarray-3305 s1 & 6 & 49 & 13 & 13 & 13 & 7 & 7 \\
xarray-3305 s5 & 26 & 45 & 26 & 26 & 44 & 44 & 44 \\
xarray-3305 s7 & 6 & 46 & 10 & 10 & 10 & 10 & 34 \\
xarray-4695 s3 & 34 & 35 & 38 & 38 & 38 & 61 & 16 \\
\bottomrule
\end{tabular}}
\caption{First-mistake labels of every source next to $t^*$ and $t_{\text{pnr}}$, first half of the set (``none'' = no mistake; $\cdot$ = not labelled by that source).}
\label{arm:tab:alllabels}
\end{table*}\vspace{-5pt}
\begin{table*}[tp]
\centering
{\scriptsize\setlength{\tabcolsep}{3pt}
\begin{tabular}[t]{lccccccc}
\toprule
 &  &  & GPT-6.1 & GPT-6 & GPT-5.5 & OpenAI & GPT-6 \\ Trajectory & $t^*$ & $t_{\text{pnr}}$ & Sol & Astra &  & o3 & Luna \\
\midrule
xarray-4695 s4 & 32 & 35 & 6 & 6 & 6 & 8 & 4 \\
xarray-7393 s0 & 7 & 36 & 34 & 20 & 82 & 82 & 34 \\
xarray-7393 s4 & 46 & 96 & 50 & 13 & 50 & 50 & 11 \\
pytest-5631 s0 & 1 & 25 & 4 & 4 & 13 & 18 & 13 \\
pytest-5631 s6 & 1 & 40 & 6 & 3 & 18 & 46 & 18 \\
pytest-5631 s7 & 1 & 51 & 6 & 6 & 6 & 72 & 6 \\
pytest-6202 s3 & 11 & 25 & 83 & 16 & 83 & 83 & 83 \\
pytest-6202 s7 & 57 & 70 & 74 & 74 & 74 & 74 & 74 \\
pytest-7432 s0 & 17 & 46 & 12 & 12 & 12 & 17 & 12 \\
pytest-7521 s6 & 28 & 30 & 17 & 17 & 80 & none & 80 \\
pytest-7982 s5 & 57 & 71 & 29 & 29 & 74 & 74 & 74 \\
scikit-learn-13439 s2 & 57 & 57 & none & none & none & none & none \\
scikit-learn-13439 s4 & 53 & 53 & 20 & 20 & 20 & 22 & none \\
scikit-learn-14983 s2 & 83 & 90 & 27 & 27 & 27 & 97 & 27 \\
scikit-learn-15100 s6 & 38 & 38 & 5 & 5 & none & none & none \\
sphinx-11445 s3 & 33 & 46 & 16 & 15 & 16 & 16 & 16 \\
sphinx-11445 s6 & 3 & 25 & 10 & 10 & none & none & 77 \\
sphinx-7757 s5 & 65 & 65 & 16 & 16 & none & 27 & 27 \\
sphinx-8459 s0 & 1 & 10 & 5 & 5 & 81 & 5 & 64 \\
sphinx-9320 s7 & 40 & 40 & 8 & 8 & 8 & none & 8 \\
sphinx-9658 s1 & 26 & 60 & 27 & 14 & 42 & none & 13 \\
sphinx-9658 s3 & 6 & 50 & 85 & 85 & none & none & 14 \\
sphinx-9658 s6 & 36 & 100 & 38 & 38 & 55 & none & 40 \\
sympy-13615 s1 & 6 & 61 & 7 & 7 & 11 & 30 & 28 \\
sympy-13615 s3 & 19 & 49 & 8 & 8 & 8 & 74 & 8 \\
sympy-13615 s6 & 9 & 41 & 9 & 9 & 9 & 61 & 9 \\
sympy-19783 s0 & 6 & 30 & 33 & 33 & 33 & 33 & 33 \\
sympy-19783 s7 & 1 & 16 & 26 & 7 & 7 & 40 & 6 \\
sympy-22714 s0 & 10 & 35 & 10 & 10 & 10 & 12 & 10 \\
sympy-22714 s5 & 17 & 70 & 16 & 16 & 20 & 16 & 14 \\
\bottomrule
\end{tabular}}
\caption{First-mistake labels of every source next to $t^*$ and $t_{\text{pnr}}$, second half of the set.}
\label{arm:tab:alllabels-2}
\end{table*}\vspace{-5pt}

\FloatBarrier
\subsection{Schedule statistics}
\label{arm:app:schedule}

Table~\ref{arm:tab:stages} breaks the 20,720 continuations of the frozen curves down by schedule stage. A curve has 25.6 points on average (range 12--33) and costs 345 continuations (median 368, range 128--464); screening and the label probes dominate, confirmation and the point of no return add the rest. Control continuations (Appendix~\ref{arm:app:controls}) are not in the curves. The label probes are the price of the direct measurement in \S\ref{sec:rq1}: every distinct labeled step costs two points at $K\ge16$, and because the four judges rarely agree, most trajectories carry three or four distinct labels. The confirmation and PNR stages are cheaper than their $K$ suggests because they reuse the screening and bisection continuations at the same histories (continuations are cumulative per history). The cost of a curve is set by its length and by how many drops screening finds, not by the outcome: cliffs and erosions cost about the same. Conclusion: a curve without label probes costs about 291 continuations, and the probes, which exist only to test the judges, add 55; a controller that needs the restart point alone would spend on screening and the point of no return and skip confirmation (\S\ref{sec:cheap}).

\begin{table}[htbp]
\centering
{\footnotesize\setlength{\tabcolsep}{3pt}
\begin{tabular}[t]{lrrr}
\toprule
Stage & Cont. & Share & Per traj. \\
\midrule
screen & 7,328 & 35\% & 122 \\
confirm & 4,608 & 22\% & 77 \\
label & 3,272 & 16\% & 55 \\
pnr & 3,128 & 15\% & 52 \\
bisect & 2,384 & 12\% & 40 \\
\midrule all & 20,720 & 100\% & 345 \\
\bottomrule
\end{tabular}}
\caption{Continuations by schedule stage over the frozen set.}
\label{arm:tab:stages}
\end{table}\vspace{-5pt}

\FloatBarrier
\subsection{Replay}
\label{arm:app:replay}

Table~\ref{arm:tab:replay} reports replay (Stage~3) by role and by stratum (left) and the mismatches it found (right: by type, by the kind of command at the first mismatch), and Table~\ref{arm:tab:replay-levels} the comparison level at which observations matched. 80 of 111 trajectories replay exactly; the L0-submit stratum replays worst (73\%). Non-replayable trajectories print genuinely non-deterministic values; 1,372 anchor snapshots were built, and 21 back-fills in 3 round(s) left 60 of 81 selected trajectories for Stage~4. The first mismatch occurs at step 26 on median (quartiles 8, 53; maximum 72), and almost always in an observation rather than in the filesystem fingerprint: the commands involved print values that cannot be masked by context, such as random tokens, savepoint names, \texttt{git stash} output or the ordering of a hash-based structure. We drop rather than patch such trajectories, because a continuation from a state that differs from the recorded one would measure a different history. Most matches are exact; the sorted-lines and character-multiset levels absorb reorderings of directory listings and test output whose content is identical. The L2 stratum replays best, plausibly because capped runs rarely reach the commands that print non-deterministic values. Conclusion: exact replay is attainable for about three quarters of real SWE-bench trajectories with a normalization fixed in advance, the failures are explainable and are reported rather than repaired, and because back-fills are drawn from the same strata in a fixed order, the study set remains a stratified sample of the band's failures.

\begin{table}[htbp]
\centering
{\scriptsize\setlength{\tabcolsep}{3pt}
\begin{tabular}[t]{lrrr}
\toprule
Group & $n$ & Repl. & Rate \\
\midrule
selected & 81 & 60 & 74\% \\
decoy & 10 & 7 & 70\% \\
recovery & 20 & 13 & 65\% \\
\midrule L0-submit & 22 & 16 & 73\% \\
\midrule L2-cap & 58 & 43 & 74\% \\
\midrule L2-submit & 1 & 1 & 100\% \\
\midrule all & 111 & 80 & 72\% \\
\bottomrule
\end{tabular}}\hfill
{\scriptsize\setlength{\tabcolsep}{3pt}
\begin{tabular}[t]{p{0.30\columnwidth}r}
\toprule
Mismatch & $n$ \\
\midrule
observation & 30 \\
docker\_diff\_sha256 & 1 \\
\midrule \\
git stash & 11 \\
python script & 3 \\
write file / heredoc & 10 \\
other & 3 \\
test run & 4 \\
\bottomrule
\end{tabular}}
\caption{Replay by role and stratum (left); mismatch types and command kinds at the first mismatch (right).}
\label{arm:tab:replay}
\end{table}\vspace{-5pt}
\begin{table}[htbp]
\centering
{\scriptsize\setlength{\tabcolsep}{3pt}
\begin{tabular}[t]{ccc}
\toprule
sorted\_lines & exact & char\_multiset \\
\midrule
31 & 45 & 4 \\
\bottomrule
\end{tabular}}
\caption{Comparison level at which the observations of replayable trajectories matched (steps).}
\label{arm:tab:replay-levels}
\end{table}\vspace{-5pt}

\FloatBarrier
\subsection{Judge agreement in detail}
\label{arm:app:judges}

Table~\ref{arm:tab:judge-matrix} gives the pairwise agreement of every judge pair on the 60 failed trajectories as exact / within 1 / within 3 steps, and Table~\ref{arm:tab:judge-detail} each judge's number of labels, answers that did not follow the required format (a single step index, or a sentinel for ``no mistake''), ``no mistake'' answers on failed runs, flagged decoys, label-position quartiles ($m/T$), self-consistency over three passes on 20 trajectories (exact / within 1 / within 3) and API spend. A consensus of the four main judges is rare: unanimous on the exact step in 8 of 70 trajectories, at least two within one step in 68. Total judge spend US\$63. The matrix has a block structure: the two GPT-6 judges agree with each other far more often than either agrees with GPT-5.5 or OpenAI o3, and GPT-5.5, OpenAI o3 and the small contrast model form a looser second group that answers ``no mistake'' more often and labels later in the trajectory (position quartiles in Table~\ref{arm:tab:judge-detail}). Self-consistency separates the groups as well: the GPT-6 judges return the same answer in three passes on most trajectories, GPT-5.5 and OpenAI o3 on a minority, so part of the cross-lineage disagreement is within-judge sampling noise rather than a stable difference of opinion. The spend column shows the cost of one pass over the study set per judge; the three self-consistency passes on the 20-trajectory subset are included. Conclusion: the judges do not provide four independent opinions but two lineages with different biases, one that labels early and flags decoys, one that labels late and abstains; any ground truth built by aggregating them inherits the balance between the two, and inter-judge agreement overstates reliability unless computed across lineages. The names used throughout the paper map to the following API model identifiers, fixed for the whole study: GPT-6.1 Sol: \texttt{gpt-6.1-sol}, GPT-5.5: \texttt{gpt-5.5-2026-04-23}, OpenAI o3: \texttt{o3-2025-04-16}, GPT-6 Astra: \texttt{gpt-6-astra}, GPT-6 Luna: \texttt{gpt-6-luna}; the annotator-protocol model is Claude Opus 5.5 (\texttt{claude-opus-5-5}). The two policies are \texttt{Qwen/Qwen3.6-27B-FP8} (main) and \texttt{Qwen/Qwen3.5-9B} (second policy), at the revisions recorded in the released manifest.

\begin{table}[htbp]
\centering
{\scriptsize\setlength{\tabcolsep}{3pt}
\begin{tabular}[t]{lccccc}
\toprule
 & GPT-6.1 & GPT-5.5 & OpenAI & GPT-6 & GPT-6 \\  & Sol &  & o3 & Astra & Luna \\
\midrule
GPT-6.1 Sol & -- & 30/30/31 & 14/14/17 & 47/49/51 & 27/27/31 \\
GPT-5.5 & 30/30/31 & -- & 23/24/28 & 27/28/30 & 29/31/33 \\
OpenAI o3 & 14/14/17 & 23/24/28 & -- & 8/9/13 & 17/17/20 \\
GPT-6 Astra & 47/49/51 & 27/28/30 & 8/9/13 & -- & 21/24/30 \\
GPT-6 Luna & 27/27/31 & 29/31/33 & 17/17/20 & 21/24/30 & -- \\
\bottomrule
\end{tabular}}
\caption{Pairwise agreement on the 60 failed trajectories: exact / within 1 / within 3 steps.}
\label{arm:tab:judge-matrix}
\end{table}\vspace{-5pt}
\begin{table}[htbp]
\centering
{\scriptsize\setlength{\tabcolsep}{3pt}
\begin{tabular}[t]{lccccc}
\toprule
 & GPT-6.1 & GPT-5.5 & OpenAI & GPT-6 & GPT-6 \\  & Sol &  & o3 & Astra & Luna \\
\midrule
labels (pass 1) & 70 & 70 & 70 & 70 & 70 \\
unparsed answers & 0 & 0 & 0 & 0 & 0 \\
``no mistake'' (failed) & 1/60 & 6/60 & 15/60 & 1/60 & 5/60 \\
flagged decoys & 10/10 & 3/10 & 3/10 & 10/10 & 7/10 \\
position $m/T$, Q1 & 0.09 & 0.12 & 0.21 & 0.09 & 0.10 \\
position $m/T$, median & 0.19 & 0.28 & 0.41 & 0.15 & 0.23 \\
position $m/T$, Q3 & 0.31 & 0.49 & 0.61 & 0.27 & 0.39 \\
self-consistency, exact & 15/20 & 10/20 & 5/20 & 18/20 & 10/20 \\
self-consistency, $\pm1$ & 16/20 & 10/20 & 6/20 & 18/20 & 10/20 \\
self-consistency, $\pm3$ & 16/20 & 10/20 & 6/20 & 18/20 & 10/20 \\
spend (US\$) & 5.5 & 21.8 & 6.9 & 28.2 & 0.3 \\
\bottomrule
\end{tabular}}
\caption{Judges: labels, answers not in the required format, ``no mistake'' on failed runs, flagged decoys, label-position quartiles, self-consistency over three passes on 20 trajectories, spend.}
\label{arm:tab:judge-detail}
\end{table}\vspace{-5pt}

\FloatBarrier
\subsection{Controls}
\label{arm:app:controls}

Three controls run on the first trajectories of the processing order. Table~\ref{arm:tab:sanity}: on the first 5 trajectories the continuations from the empty history are compared with the task's base runs (Fisher's exact test, two-sided). Table~\ref{arm:tab:retest}: on the first 10 trajectories, screening is repeated at three anchors with fresh seeds; the median absolute gap at $K=8$ is 0.06 (maximum 0.38). Table~\ref{arm:tab:controls}: on the first 10 trajectories the screening anchors are run again with the full budget $B$ instead of $B-t$; the decisive step from screening agrees within three steps in 0 of 10 (sh.: shape, cli/ero; final: the study's $t^*$). The sanity rows show that resuming the agent from the empty history reproduces the base pass rate (no row is rejected at $\alpha=0.05$), so the continuation machinery itself does not shift the success probability. The retest rows show that at $K=8$ two independent screenings of the same history agree exactly in 15 of 30 anchors and differ by more than $0.25$ in 1; a gap of $3/8$ between two draws of eight is within binomial noise at $K=8$, which is why the schedule confirms at $K=32$ before naming a decisive step. The budget-control rows show the one systematic effect of the protocol: with the full budget the decisive step never moves earlier and moves later in 10 of 10 trajectories, in 4 cases to the last step, where a run that was still viable under $B$ runs out of steps under $B-t$. We therefore read a late $t^*$ under the protocol's $B-t$ as ``lost under the remaining budget'', which is the quantity a controller faces, and not as a judgment about the agent's reasoning alone. Conclusion: the method reproduces the base pass rate, repeats itself within binomial noise, and has one known bias, a late decisive step on long runs, whose direction and size are measured here and whose effect on the main claims is bounded by the analysis without the L2 stratum (Appendix~\ref{arm:app:rq1}).

\begin{table}[htbp]
\centering
{\scriptsize\setlength{\tabcolsep}{3pt}
\begin{tabular}[t]{lrr}
\toprule
Trajectory & At $h_0$ & $p(x)$ \\
\midrule
django-13821 s7 & 5/8 & 0.75 \\
django-13837 s2 & 2/8 & 0.38 \\
xarray-3305 s5 & 6/8 & 0.50 \\
xarray-4695 s4 & 1/8 & 0.25 \\
sympy-13615 s1 & 2/8 & 0.25 \\
\bottomrule
\end{tabular}}
\caption{$t=0$ sanity: successes of the continuations from the empty history against the task's base pass rate.}
\label{arm:tab:sanity}
\end{table}\vspace{-5pt}
\begin{table}[htbp]
\centering
{\scriptsize\setlength{\tabcolsep}{3pt}
\begin{tabular}[t]{lrrrr}
\toprule
Trajectory & $t$ & $\hat V_1$ & $\hat V_2$ & $|\Delta|$ \\
\midrule
django-11999 s0 & 0 & 0.25 & 0.38 & 0.12 \\
django-11999 s0 & 40 & 0.88 & 0.62 & 0.25 \\
django-11999 s0 & 99 & 0.00 & 0.00 & 0.00 \\
django-12713 s3 & 0 & 0.50 & 0.75 & 0.25 \\
django-12713 s3 & 40 & 0.00 & 0.12 & 0.12 \\
django-12713 s3 & 99 & 0.00 & 0.00 & 0.00 \\
django-13028 s6 & 0 & 0.00 & 0.25 & 0.25 \\
django-13028 s6 & 40 & 0.00 & 0.12 & 0.12 \\
django-13028 s6 & 99 & 0.00 & 0.00 & 0.00 \\
django-13821 s7 & 0 & 0.62 & 0.75 & 0.12 \\
django-13821 s7 & 40 & 0.50 & 0.50 & 0.00 \\
django-13821 s7 & 99 & 0.00 & 0.00 & 0.00 \\
django-13837 s2 & 0 & 0.25 & 0.25 & 0.00 \\
django-13837 s2 & 40 & 0.38 & 0.12 & 0.25 \\
django-13837 s2 & 99 & 0.00 & 0.00 & 0.00 \\
django-16116 s0 & 0 & 0.50 & 0.25 & 0.25 \\
django-16116 s0 & 35 & 0.50 & 0.50 & 0.00 \\
django-16116 s0 & 78 & 0.38 & 0.75 & 0.38 \\
xarray-3305 s5 & 0 & 0.75 & 0.50 & 0.25 \\
xarray-3305 s5 & 40 & 0.00 & 0.12 & 0.12 \\
xarray-3305 s5 & 99 & 0.00 & 0.00 & 0.00 \\
xarray-4695 s4 & 0 & 0.12 & 0.12 & 0.00 \\
xarray-4695 s4 & 40 & 0.00 & 0.00 & 0.00 \\
xarray-4695 s4 & 99 & 0.00 & 0.00 & 0.00 \\
sphinx-9658 s1 & 0 & 0.25 & 0.25 & 0.00 \\
sphinx-9658 s1 & 35 & 0.38 & 0.50 & 0.12 \\
sphinx-9658 s1 & 84 & 0.00 & 0.00 & 0.00 \\
sympy-13615 s1 & 0 & 0.25 & 0.50 & 0.25 \\
sympy-13615 s1 & 40 & 0.12 & 0.00 & 0.12 \\
sympy-13615 s1 & 99 & 0.00 & 0.00 & 0.00 \\
\bottomrule
\end{tabular}}
\caption{Test--retest: screening estimate at three anchors from the main schedule ($\hat V_1$) and from fresh seeds ($\hat V_2$), $K=8$ each.}
\label{arm:tab:retest}
\end{table}\vspace{-5pt}
\begin{table}[htbp]
\centering
{\scriptsize\setlength{\tabcolsep}{3pt}
\begin{tabular}[t]{lrrlrlr}
\toprule
Trajectory & $T$ & $t^*_{B-t}$ & sh. & $t^*_{B}$ & sh. & final \\
\midrule
django-11999 s0 & 100 & 41 & cli & 91 & cli & 71 \\
django-12713 s3 & 100 & 16 & cli & 31 & ero & 34 \\
django-13028 s6 & 100 & 11 & cli & 51 & cli & 12 \\
django-13821 s7 & 100 & 81 & ero & 100 & cli & 71 \\
django-13837 s2 & 100 & 6 & cli & 16 & ero & 26 \\
django-16116 s0 & 79 & 71 & cli & 79 & cli & 79 \\
xarray-3305 s5 & 100 & 26 & ero & 100 & cli & 26 \\
xarray-4695 s4 & 100 & 1 & cli & 11 & ero & 32 \\
sphinx-9658 s1 & 85 & 11 & cli & 36 & cli & 26 \\
sympy-13615 s1 & 100 & 6 & ero & 100 & cli & 6 \\
\bottomrule
\end{tabular}}
\caption{Budget control: screening with $B-t$ and with the full budget $B$.}
\label{arm:tab:controls}
\end{table}\vspace{-5pt}

\FloatBarrier
\subsection{Labels against the curves, in full}
\label{arm:app:rq1}

Table~\ref{arm:tab:rq1-full} extends Table~\ref{tab:rq1} to every source and every quantity, one column per source (the four main judges, GPT-6 Luna as the contrast model, Claude Opus 5.5 under the annotator protocol on its subset, and the four main judges pooled). The upper block counts labels by position relative to the decisive interval and the point of no return; the lower block gives the direct probes at the labeled step (zero-drop: $V(h_{m-1})-V(h_m)\le0$; significant: one-sided Fisher $p<\alpha$), the H1 share (post-cause or zero-drop) with its Wilson interval, and, on runs with an explicit failure signal, labels within one step of $t^*$ and of $t_{\text{sym}}$. By shape, the pooled H1 share is 69\% on cliffs (68 labels) and 80\% on erosion (149); 24 of 60 trajectories are silent (no failure signal after the first edit). Three readings of the table. First, the H1 share is high for every source, from 65\% to 87\%, and the sources differ mainly in how they miss: the GPT-6 judges miss early (``pre'': the labeled step reads as a mistake but the curve has not moved) and the GPT-5.5/OpenAI o3 lineage misses late (``post-PNR'': the task was certified lost before the label). Second, the zero-drop row is the strongest evidence, because it does not depend on where $t^*$ is: at the labeled step itself, probed with $K\ge{}$16 on both sides, the success probability does not fall for about two thirds of the labels of every source, and a drop of $0.2$ or more is found at a few labels only. Third, the labels are not the first symptom either: on symptomatic runs almost none lie within one step of $t_{\text{sym}}$, and only a minority within one step of $t^*$; what judges mark is the step where a wrong idea is stated or acted upon, which the curve usually dates earlier. Conclusion: on this policy the first-mistake label is not a measurement of where the task was lost. The result holds for every judge, for both lineages, for the annotator-protocol model, for cliffs and erosions, and without the L2 stratum (Table~\ref{arm:tab:rq1-noL2}); what varies between sources is the direction of the miss, not its existence.

\begin{table}[htbp]
\centering
{\scriptsize\setlength{\tabcolsep}{1.6pt}
\begin{tabular}[t]{lcccccc}
\toprule
 & GPT-6.1 & GPT-5.5 & OpenAI & GPT-6 & GPT-6 & Pooled \\  & Sol &  & o3 & Astra & Luna &  \\
\midrule
trajectories & 60 & 60 & 60 & 60 & 60 & 60 \\
labels & 59 & 54 & 45 & 59 & 55 & 217 \\
\midrule \\
pre & 29 & 17 & 11 & 31 & 24 & 88 \\
on & 4 & 4 & 1 & 5 & 4 & 14 \\
post & 19 & 20 & 14 & 19 & 19 & 72 \\
post-PNR & 7 & 13 & 19 & 4 & 8 & 43 \\
no mistake & 1 & 6 & 15 & 1 & 5 & 23 \\
\midrule \\
zero-drop & 35/59 & 30/54 & 35/45 & 36/59 & 20/36 & 136/217 \\
drop $\ge0.2$ & 3/59 & 3/54 & 2/45 & 4/59 & 2/36 & 12/217 \\
significant & 3/59 & 3/54 & 2/45 & 4/59 & 2/36 & 12/217 \\
H1 & 71\% & 76\% & 87\% & 75\% & 65\% & 76\% \\
H1 95\% CI & 59--81 & 63--85 & 74--94 & 62--84 & 52--77 & 70--82 \\
near $t^*$ & 3/35 & 3/33 & 2/28 & 4/35 & 3/32 & 12/131 \\
near $t_{\text{sym}}$ & 0/35 & 1/33 & 0/28 & 0/35 & 1/32 & 1/131 \\
\bottomrule
\end{tabular}}
\caption{Labels against the curves on the frozen set (60 trajectories), every source and quantity.}
\label{arm:tab:rq1-full}
\end{table}\vspace{-5pt}
Table~\ref{arm:tab:rq1-noL2} repeats the analysis without the L2 stratum (base runs that hit the cap), on 17 trajectories: the pooled H1 share is 67\% (36/54) against 76\% on the full set.

\begin{table}[htbp]
\centering
{\scriptsize\setlength{\tabcolsep}{1.6pt}
\begin{tabular}[t]{lcccccc}
\toprule
 & GPT-6.1 & GPT-5.5 & OpenAI & GPT-6 & GPT-6 & Pooled \\  & Sol &  & o3 & Astra & Luna &  \\
\midrule
trajectories & 17 & 17 & 17 & 17 & 17 & 17 \\
labels & 16 & 14 & 8 & 16 & 13 & 54 \\
\midrule \\
pre & 10 & 7 & 5 & 11 & 8 & 33 \\
on & 0 & 1 & 0 & 1 & 0 & 2 \\
post & 5 & 5 & 1 & 4 & 4 & 15 \\
post-PNR & 1 & 1 & 2 & 0 & 1 & 4 \\
no mistake & 1 & 3 & 9 & 1 & 4 & 14 \\
\midrule \\
zero-drop & 8/16 & 7/14 & 5/8 & 10/16 & 4/9 & 30/54 \\
drop $\ge0.2$ & 0/16 & 1/14 & 0/8 & 1/16 & 0/9 & 2/54 \\
significant & 0/16 & 1/14 & 0/8 & 1/16 & 0/9 & 2/54 \\
H1 & 62\% & 64\% & 62\% & 75\% & 62\% & 67\% \\
H1 95\% CI & 39--82 & 39--84 & 31--86 & 51--90 & 36--82 & 53--78 \\
near $t^*$ & 0/9 & 1/8 & 0/5 & 1/9 & 0/7 & 2/31 \\
near $t_{\text{sym}}$ & 0/9 & 0/8 & 0/5 & 0/9 & 0/7 & 0/31 \\
\bottomrule
\end{tabular}}
\caption{The same without the L2 stratum (17 trajectories whose base run submitted).}
\label{arm:tab:rq1-noL2}
\end{table}\vspace{-5pt}

\FloatBarrier
\subsection{Restart policies, in full}
\label{arm:app:restart}

Table~\ref{arm:tab:restart-full} lists every restart policy evaluated offline on the 60 frozen curves (margin $\mu=0.1$): how many trajectories it restarts, the mean restart step, its expected success on the failed runs alone and on a random run of the task (successes included), and the gain over never restarting. Table~\ref{arm:tab:decoys} gives the judge policies' behaviour on the successful decoys, where a label is a restart of a run that would have succeeded. Tables~\ref{arm:tab:restart-traj} and \ref{arm:tab:restart-traj-judges} give, per trajectory, the restart step and the expected success $R(x,B-t)$ of the value-based policies and of each judge's label. The value-based policies restart early (mean step 6 at the restart point, 25 at $t^*$) and almost always: on a failed run the only way to recover value is a fresh start while enough budget remains. The judge policies restart on 59/60 (GPT-6.1 Sol) to 45/60 (OpenAI o3) trajectories, and late (mean step 33 for GPT-5.5, 42 for OpenAI o3), so that the fresh run has little budget left; the fixed policies show the same trade-off along a single axis, restarting everything at step 10 being too eager (it forfeits successful runs that pass step 10) and at step 50 too late. The decoy table quantifies the eagerness of the judge policies on successful runs: a label on a decoy is a restart that forfeits a success, and the GPT-6 judges, which label most decoys, lose most there. The per-trajectory tables show where the judge policies lose: the GPT-6 labels fall within three steps of $t^*$ on 26 of 120 judge--trajectory pairs, and there the restart value is close to the value-based policy's; elsewhere, and for the GPT-5.5/OpenAI o3 lineage on most trajectories, the label falls long after $t^*$, often after the point of no return, where $R(x,B-t)$ has decayed to a few percent. Conclusion: the gain of a restart policy is decided by two things, how early it restarts failing runs and how rarely it restarts succeeding ones; the measured curve gives both, a certified point of no return gives the second with a known error rate at some cost in the first, and a judge's label gives neither, since it comes late on failed runs and fires on successful ones.

\begin{table}[htbp]
\centering
{\scriptsize\setlength{\tabcolsep}{2pt}
\begin{tabular}[t]{lrrrrr}
\toprule
Policy & Restarts & $\bar t$ & Failed & Random & Gain \\
\midrule
never & 0/60 & -- & 0.00 & 0.45 & +0.00 \\
restart point $t_r$ & 15/60 & 6 & 0.11 & 0.50 & +0.05 \\
$t_{\text{pnr}}$ & 50/60 & 49 & 0.04 & 0.47 & +0.02 \\
$t^*$ & 51/60 & 25 & 0.17 & 0.54 & +0.09 \\
label, GPT-6.1 Sol & 59/60 & 23 & 0.24 & 0.33 & -0.13 \\
label, GPT-5.5 & 54/60 & 33 & 0.16 & 0.42 & -0.03 \\
label, OpenAI o3 & 45/60 & 42 & 0.11 & 0.41 & -0.04 \\
label, GPT-6 Astra & 59/60 & 20 & 0.26 & 0.34 & -0.11 \\
fixed $t=10$ & 60/60 & 10 & 0.36 & 0.36 & -0.09 \\
fixed $t=20$ & 60/60 & 20 & 0.25 & 0.25 & -0.20 \\
fixed $t=30$ & 60/60 & 30 & 0.15 & 0.15 & -0.30 \\
fixed $t=50$ & 58/60 & 50 & 0.01 & 0.02 & -0.43 \\
\bottomrule
\end{tabular}}
\caption{Restart policies on the frozen set (60 trajectories). $\bar t$: mean restart step; Failed / Random: expected success on failed runs / on a random run.}
\label{arm:tab:restart-full}
\end{table}\vspace{-5pt}
\begin{table}[htbp]
\centering
{\scriptsize\setlength{\tabcolsep}{3pt}
\begin{tabular}[t]{lrrr}
\toprule
Judge & Decoys & Restarted & Loss \\
\midrule
GPT-6.1 Sol & 10 & 10 & 0.51 \\
GPT-5.5 & 10 & 3 & 0.23 \\
OpenAI o3 & 10 & 3 & 0.20 \\
GPT-6 Astra & 10 & 10 & 0.51 \\
\bottomrule
\end{tabular}}
\caption{Judge policies on successful decoys: labelled decoys are restarted; expected loss $1-R(x,B-m)$ per decoy.}
\label{arm:tab:decoys}
\end{table}\vspace{-5pt}
\begin{table}[htbp]
\centering
{\scriptsize\setlength{\tabcolsep}{3pt}
\begin{tabular}[t]{lrrrr}
\toprule
Trajectory & $p(x)$ & $t_r$ & $t_{\text{pnr}}$ & $t^*$ \\
\midrule
django-11333 s6 & 0.38 & 7/0.25 & 36/0.00 & 7/0.25 \\
django-11451 s0 & 0.50 & -- & 71/0.00 & 1/0.50 \\
django-11451 s2 & 0.50 & -- & -- & -- \\
django-11999 s0 & 0.25 & -- & 60/0.00 & 71/0.00 \\
django-11999 s1 & 0.25 & -- & 20/0.00 & 40/0.00 \\
django-11999 s6 & 0.25 & -- & 48/0.00 & 21/0.00 \\
django-12713 s3 & 0.62 & 17/0.62 & 35/0.12 & 34/0.12 \\
django-13028 s2 & 0.25 & -- & 25/0.00 & 11/0.12 \\
django-13028 s5 & 0.25 & -- & -- & -- \\
django-13028 s6 & 0.25 & 0/0.25 & 10/0.12 & 12/0.12 \\
django-13089 s5 & 0.62 & -- & -- & -- \\
django-13551 s4 & 0.62 & 23/0.62 & 42/0.25 & 23/0.62 \\
django-13821 s6 & 0.75 & 0/0.75 & 56/0.00 & 37/0.12 \\
django-13821 s7 & 0.75 & -- & 89/0.00 & 71/0.00 \\
django-13837 s0 & 0.38 & 5/0.38 & 20/0.25 & 1/0.38 \\
django-13837 s2 & 0.38 & -- & 64/0.00 & 26/0.12 \\
django-13933 s0 & 0.50 & -- & 75/0.00 & 11/0.38 \\
django-13933 s6 & 0.50 & -- & 61/0.00 & 11/0.38 \\
django-16116 s0 & 0.50 & -- & -- & -- \\
django-16116 s4 & 0.50 & -- & 90/0.00 & 63/0.00 \\
django-16139 s0 & 0.50 & -- & 91/0.00 & 83/0.00 \\
django-16139 s1 & 0.50 & -- & 66/0.00 & 5/0.50 \\
django-16139 s3 & 0.50 & -- & 80/0.00 & 24/0.38 \\
django-16899 s0 & 0.38 & -- & 70/0.00 & 60/0.00 \\
django-16899 s3 & 0.38 & -- & 39/0.00 & 39/0.00 \\
django-16899 s7 & 0.38 & -- & 35/0.00 & 35/0.00 \\
xarray-3305 s1 & 0.50 & -- & 49/0.00 & 6/0.38 \\
xarray-3305 s5 & 0.50 & -- & 45/0.00 & 26/0.12 \\
xarray-3305 s7 & 0.50 & -- & 46/0.00 & 6/0.38 \\
xarray-4695 s3 & 0.25 & -- & 35/0.00 & 34/0.00 \\
xarray-4695 s4 & 0.25 & 0/0.25 & 35/0.00 & 32/0.00 \\
xarray-7393 s0 & 0.62 & 0/0.62 & 36/0.12 & 7/0.50 \\
xarray-7393 s4 & 0.62 & -- & 96/0.00 & 46/0.00 \\
pytest-5631 s0 & 0.62 & 2/0.50 & 25/0.38 & 1/0.50 \\
pytest-5631 s6 & 0.62 & 1/0.50 & 40/0.12 & 1/0.50 \\
pytest-5631 s7 & 0.62 & 2/0.50 & 51/0.00 & 1/0.50 \\
pytest-6202 s3 & 0.25 & -- & 25/0.12 & 11/0.12 \\
pytest-6202 s7 & 0.25 & -- & 70/0.00 & 57/0.00 \\
pytest-7432 s0 & 0.75 & 0/0.75 & 46/0.12 & 17/0.62 \\
pytest-7521 s6 & 0.50 & 29/0.25 & 30/0.25 & 28/0.25 \\
pytest-7982 s5 & 0.50 & -- & 71/0.00 & 57/0.00 \\
scikit-learn-13439 s2 & 0.75 & -- & -- & -- \\
scikit-learn-13439 s4 & 0.75 & -- & -- & -- \\
scikit-learn-14983 s2 & 0.62 & -- & 90/0.00 & 83/0.00 \\
scikit-learn-15100 s6 & 0.75 & -- & -- & -- \\
sphinx-11445 s3 & 0.25 & -- & 46/0.00 & 33/0.00 \\
sphinx-11445 s6 & 0.25 & -- & 25/0.00 & 3/0.12 \\
sphinx-7757 s5 & 0.75 & -- & -- & -- \\
sphinx-8459 s0 & 0.25 & 4/0.25 & 10/0.25 & 1/0.25 \\
sphinx-9320 s7 & 0.75 & -- & -- & -- \\
sphinx-9658 s1 & 0.25 & -- & 60/0.00 & 26/0.12 \\
sphinx-9658 s3 & 0.25 & -- & 50/0.00 & 6/0.25 \\
sphinx-9658 s6 & 0.25 & -- & -- & 36/0.00 \\
sympy-13615 s1 & 0.25 & -- & 61/0.00 & 6/0.12 \\
sympy-13615 s3 & 0.25 & -- & 49/0.00 & 19/0.12 \\
sympy-13615 s6 & 0.25 & -- & 41/0.00 & 9/0.12 \\
sympy-19783 s0 & 0.38 & -- & 30/0.12 & 6/0.38 \\
sympy-19783 s7 & 0.38 & 2/0.38 & 16/0.12 & 1/0.38 \\
sympy-22714 s0 & 0.38 & -- & 35/0.00 & 10/0.25 \\
sympy-22714 s5 & 0.38 & -- & 70/0.00 & 17/0.12 \\
\bottomrule
\end{tabular}}
\caption{Value-based policies per trajectory: restart step / expected success $R(x,B-t)$ (-- = no restart).}
\label{arm:tab:restart-traj}
\end{table}\vspace{-5pt}
\begin{table}[htbp]
\centering
{\scriptsize\setlength{\tabcolsep}{3pt}
\begin{tabular}[t]{lrrrr}
\toprule
 & GPT-6.1 & GPT-6 & GPT-5.5 & OpenAI \\ Trajectory & Sol & Astra &  & o3 \\
\midrule
django-11333 s6 & 7/0.25 & 7/0.25 & 7/0.25 & 22/0.12 \\
django-11451 s0 & 19/0.50 & 17/0.50 & 19/0.50 & 19/0.50 \\
django-11451 s2 & 21/0.50 & 21/0.50 & 21/0.50 & 21/0.50 \\
django-11999 s0 & 28/0.00 & 15/0.12 & 99/0.00 & 99/0.00 \\
django-11999 s1 & 11/0.12 & 11/0.12 & 43/0.00 & 43/0.00 \\
django-11999 s6 & 9/0.12 & 9/0.12 & 24/0.00 & 38/0.00 \\
django-12713 s3 & 7/0.62 & 7/0.62 & 32/0.12 & 33/0.12 \\
django-13028 s2 & 2/0.12 & 2/0.12 & 32/0.00 & -- \\
django-13028 s5 & 24/0.00 & 24/0.00 & 24/0.00 & -- \\
django-13028 s6 & 10/0.12 & 10/0.12 & 10/0.12 & 59/0.00 \\
django-13089 s5 & 23/0.50 & 10/0.62 & 23/0.50 & 23/0.50 \\
django-13551 s4 & 42/0.25 & 42/0.25 & 42/0.25 & -- \\
django-13821 s6 & 31/0.38 & 31/0.38 & 31/0.38 & 57/0.00 \\
django-13821 s7 & 75/0.00 & 75/0.00 & 75/0.00 & -- \\
django-13837 s0 & 8/0.38 & 8/0.38 & 8/0.38 & 14/0.38 \\
django-13837 s2 & 15/0.38 & 14/0.38 & 28/0.12 & 60/0.00 \\
django-13933 s0 & 20/0.25 & 20/0.25 & 32/0.00 & 32/0.00 \\
django-13933 s6 & 13/0.25 & 13/0.25 & 13/0.25 & 63/0.00 \\
django-16116 s0 & 13/0.00 & 13/0.00 & 11/0.00 & -- \\
django-16116 s4 & 34/0.00 & 34/0.00 & 43/0.00 & 43/0.00 \\
django-16139 s0 & 12/0.50 & 12/0.50 & -- & -- \\
django-16139 s1 & 38/0.00 & 38/0.00 & 84/0.00 & 86/0.00 \\
django-16139 s3 & 21/0.50 & 21/0.50 & 21/0.50 & 21/0.50 \\
django-16899 s0 & 22/0.12 & 22/0.12 & 43/0.00 & 43/0.00 \\
django-16899 s3 & 21/0.12 & 39/0.00 & 39/0.00 & -- \\
django-16899 s7 & 37/0.00 & 20/0.12 & 43/0.00 & 37/0.00 \\
xarray-3305 s1 & 13/0.25 & 13/0.25 & 13/0.25 & 7/0.25 \\
xarray-3305 s5 & 26/0.12 & 26/0.12 & 44/0.00 & 44/0.00 \\
xarray-3305 s7 & 10/0.25 & 10/0.25 & 10/0.25 & 10/0.25 \\
xarray-4695 s3 & 38/0.00 & 38/0.00 & 38/0.00 & 61/0.00 \\
xarray-4695 s4 & 6/0.25 & 6/0.25 & 6/0.25 & 8/0.25 \\
xarray-7393 s0 & 34/0.12 & 20/0.25 & 82/0.00 & 82/0.00 \\
xarray-7393 s4 & 50/0.00 & 13/0.38 & 50/0.00 & 50/0.00 \\
pytest-5631 s0 & 4/0.50 & 4/0.50 & 13/0.38 & 18/0.38 \\
pytest-5631 s6 & 6/0.50 & 3/0.50 & 18/0.38 & 46/0.00 \\
pytest-5631 s7 & 6/0.50 & 6/0.50 & 6/0.50 & 72/0.00 \\
pytest-6202 s3 & 83/0.00 & 16/0.12 & 83/0.00 & 83/0.00 \\
pytest-6202 s7 & 74/0.00 & 74/0.00 & 74/0.00 & 74/0.00 \\
pytest-7432 s0 & 12/0.75 & 12/0.75 & 12/0.75 & 17/0.62 \\
pytest-7521 s6 & 17/0.38 & 17/0.38 & 80/0.00 & -- \\
pytest-7982 s5 & 29/0.25 & 29/0.25 & 74/0.00 & 74/0.00 \\
scikit-learn-13439 s2 & -- & -- & -- & -- \\
scikit-learn-13439 s4 & 20/0.62 & 20/0.62 & 20/0.62 & 22/0.62 \\
scikit-learn-14983 s2 & 27/0.25 & 27/0.25 & 27/0.25 & 97/0.00 \\
scikit-learn-15100 s6 & 5/0.75 & 5/0.75 & -- & -- \\
sphinx-11445 s3 & 16/0.00 & 15/0.00 & 16/0.00 & 16/0.00 \\
sphinx-11445 s6 & 10/0.12 & 10/0.12 & -- & -- \\
sphinx-7757 s5 & 16/0.75 & 16/0.75 & -- & 27/0.50 \\
sphinx-8459 s0 & 5/0.25 & 5/0.25 & 81/0.00 & 5/0.25 \\
sphinx-9320 s7 & 8/0.75 & 8/0.75 & 8/0.75 & -- \\
sphinx-9658 s1 & 27/0.00 & 14/0.12 & 42/0.00 & -- \\
sphinx-9658 s3 & 85/0.00 & 85/0.00 & -- & -- \\
sphinx-9658 s6 & 38/0.00 & 38/0.00 & 55/0.00 & -- \\
sympy-13615 s1 & 7/0.12 & 7/0.12 & 11/0.12 & 30/0.12 \\
sympy-13615 s3 & 8/0.12 & 8/0.12 & 8/0.12 & 74/0.00 \\
sympy-13615 s6 & 9/0.12 & 9/0.12 & 9/0.12 & 61/0.00 \\
sympy-19783 s0 & 33/0.12 & 33/0.12 & 33/0.12 & 33/0.12 \\
sympy-19783 s7 & 26/0.12 & 7/0.38 & 7/0.38 & 40/0.00 \\
sympy-22714 s0 & 10/0.25 & 10/0.25 & 10/0.25 & 12/0.25 \\
sympy-22714 s5 & 16/0.12 & 16/0.12 & 20/0.12 & 16/0.12 \\
\bottomrule
\end{tabular}}
\caption{Judge-label policies per trajectory: restart step / expected success (-- = no mistake, no restart).}
\label{arm:tab:restart-traj-judges}
\end{table}
Table~\ref{arm:tab:restart-noL2} repeats the evaluation without the L2 stratum (17 trajectories): the gain at the restart point is +0.02 (full set +0.05) and the judge policies gain -0.16 to -0.04.

\begin{table}[htbp]
\centering
{\scriptsize\setlength{\tabcolsep}{2pt}
\begin{tabular}[t]{lrrrrr}
\toprule
Policy & Restarts & $\bar t$ & Failed & Random & Gain \\
\midrule
never & 0/17 & -- & 0.00 & 0.52 & +0.00 \\
restart point $t_r$ & 2/17 & 12 & 0.07 & 0.55 & +0.02 \\
$t_{\text{pnr}}$ & 7/17 & 48 & 0.02 & 0.53 & +0.01 \\
$t^*$ & 8/17 & 24 & 0.10 & 0.57 & +0.05 \\
label, GPT-6.1 Sol & 16/17 & 21 & 0.31 & 0.36 & -0.16 \\
label, GPT-5.5 & 14/17 & 27 & 0.21 & 0.48 & -0.04 \\
label, OpenAI o3 & 8/17 & 26 & 0.15 & 0.47 & -0.05 \\
label, GPT-6 Astra & 16/17 & 19 & 0.32 & 0.37 & -0.16 \\
fixed $t=10$ & 17/17 & 10 & 0.47 & 0.47 & -0.05 \\
fixed $t=20$ & 17/17 & 20 & 0.37 & 0.37 & -0.15 \\
fixed $t=30$ & 17/17 & 30 & 0.26 & 0.26 & -0.26 \\
fixed $t=50$ & 15/17 & 50 & 0.03 & 0.08 & -0.44 \\
\bottomrule
\end{tabular}}
\caption{Restart policies without the L2 stratum (17 trajectories).}
\label{arm:tab:restart-noL2}
\end{table}\vspace{-5pt}

\FloatBarrier
\subsection{Cheaper schedules, in full}
\label{arm:app:cheap}

Table~\ref{arm:tab:cheap-full} reports the parametric bootstrap of cheaper schedules (20 replications per trajectory on 60 curves): continuations used, agreement of the estimated decisive step with the fit's largest drop (exact, within 1, within 3), the share of replications that find a drop at all, and the share of the true drop captured within $\pm10$ steps. Even the full schedule is exact in 30\% of replications. Judge-guided schedules probe the judge's label first and fall back to a search when the drop there is too small; Table~\ref{arm:tab:cheap-paths} counts the paths taken, and Table~\ref{arm:tab:cheap-shape} splits the comparison by failure shape. The rows order themselves by cost: the full schedule and the judge-guided schedules cost the most, screening with bisection but no confirmation about half, screening at $K=4$ and sparse anchors less again. Accuracy does not follow cost: the sparse schedule matches the full one within noise, because the largest drop is rarely between two anchors ten steps apart, while dropping confirmation loses exact agreement because the $K=8$ estimates at the bracketing points are too noisy to order neighbouring drops. The value-captured column is the quantity a controller cares about and it is far more forgiving than exact agreement: a schedule that misses $t^*$ by a few steps still captures most of the drop. The judge-guided schedules take the ``label verified'' path in a small minority of replications; in the rest they fall back to a local search around the label and then to the full schedule, which is why their cost equals the full schedule's plus the probe. By shape, cliffs are located far more reliably than erosions under every schedule, which is the expected behaviour of an estimator that looks for one large drop. Conclusion: the decisive step is an expensive and, on erosions, ill-defined target, while the drop itself can be captured at a fraction of the cost; a practical localizer should be evaluated on value captured and budgeted for screening and certification rather than for exact localization, and should not route its search through a judge's label.

\begin{table}[htbp]
\centering
{\scriptsize\setlength{\tabcolsep}{3pt}
\begin{tabular}[t]{lrrrrrr}
\toprule
Schedule & $n$ & Exact & $\pm1$ & $\pm3$ & Found & Capt. \\
\midrule
full schedule & 289 & 30\% & 36\% & 46\% & 73\% & 0.47 \\
screening + bisection & 162 & 25\% & 31\% & 39\% & 90\% & 0.40 \\
screening at $K=4$ & 213 & 24\% & 30\% & 36\% & 94\% & 0.37 \\
sparse anchors & 241 & 32\% & 38\% & 46\% & 66\% & 0.49 \\
judge-guided, GPT-6.1 Sol & 265 & 31\% & 38\% & 47\% & 72\% & 0.48 \\
judge-guided, GPT-5.5 & 258 & 33\% & 37\% & 47\% & 74\% & 0.50 \\
judge-guided, OpenAI o3 & 270 & 32\% & 39\% & 47\% & 74\% & 0.50 \\
judge-guided, GPT-6 Astra & 261 & 33\% & 39\% & 48\% & 72\% & 0.50 \\
\bottomrule
\end{tabular}}
\caption{Cheaper schedules: continuations, agreement with the full curve's decisive step, drops found, value captured.}
\label{arm:tab:cheap-full}
\end{table}\vspace{-5pt}
\begin{table}[htbp]
\centering
{\scriptsize\setlength{\tabcolsep}{3pt}
\begin{tabular}[t]{p{0.3\columnwidth}p{0.55\columnwidth}}
\toprule
Schedule & Paths \\
\midrule
judge-guided, GPT-6.1 Sol & fallback full 502, label verified 25, local search 653, no label: full 20 \\
judge-guided, GPT-5.5 & fallback full 430, label verified 49, local search 601, no label: full 120 \\
judge-guided, OpenAI o3 & fallback full 396, label verified 12, local search 492, no label: full 300 \\
judge-guided, GPT-6 Astra & fallback full 489, label verified 53, local search 638, no label: full 20 \\
\bottomrule
\end{tabular}}
\caption{Paths taken by the judge-guided schedules (replications).}
\label{arm:tab:cheap-paths}
\end{table}\vspace{-5pt}
\begin{table}[htbp]
\centering
{\scriptsize\setlength{\tabcolsep}{3pt}
\begin{tabular}[t]{llrrr}
\toprule
Shape & Schedule & Exact & $\pm3$ & Capt. \\
\midrule
cli & full schedule & 66\% & 76\% & 0.70 \\
cli & screening + bisection & 54\% & 63\% & 0.58 \\
cli & screening at $K=4$ & 50\% & 59\% & 0.53 \\
cli & sparse anchors & 69\% & 80\% & 0.74 \\
cli & judge-guided, GPT-6.1 Sol & 66\% & 76\% & 0.70 \\
cli & judge-guided, GPT-5.5 & 71\% & 81\% & 0.75 \\
cli & judge-guided, OpenAI o3 & 67\% & 78\% & 0.72 \\
cli & judge-guided, GPT-6 Astra & 67\% & 78\% & 0.72 \\
ero & full schedule & 12\% & 31\% & 0.35 \\
ero & screening + bisection & 10\% & 27\% & 0.31 \\
ero & screening at $K=4$ & 10\% & 25\% & 0.29 \\
ero & sparse anchors & 13\% & 29\% & 0.37 \\
ero & judge-guided, GPT-6.1 Sol & 13\% & 32\% & 0.37 \\
ero & judge-guided, GPT-5.5 & 13\% & 29\% & 0.37 \\
ero & judge-guided, OpenAI o3 & 15\% & 31\% & 0.39 \\
ero & judge-guided, GPT-6 Astra & 16\% & 32\% & 0.39 \\
\bottomrule
\end{tabular}}
\caption{Cheaper schedules by failure shape.}
\label{arm:tab:cheap-shape}
\end{table}\vspace{-5pt}

\FloatBarrier
\subsection{Ranking localizers, in full}
\label{arm:app:rq3}

Table~\ref{arm:tab:rq3-full} scores every localizer by value captured within $k$ steps (VC$_k$), by agreement with the labels of the judges of the other lineages within $k$ steps (A$_k$; a lineage consensus exists on only 40 trajectories), and by agreement with $t^*$ ($t^*_k$). Kendall's $\tau$ between the rankings by VC$_1$ and by A$_1$ over the 9 localizers scored on all trajectories is 0.44 (permutation $p=0.12$); by VC$_1$ the order is GPT-6 Astra, GPT-5.5, GPT-6.1 Sol, GPT-6 Luna (contrast), step 1, OpenAI o3, first edit, first verification, before first symptom; by A$_1$ it is GPT-5.5, GPT-6 Luna (contrast), GPT-6.1 Sol, GPT-6 Astra, first edit, OpenAI o3, before first symptom, first verification, step 1. Three features of the table. The absolute values of VC$_k$ are low for every localizer, judges and heuristics alike: on erosion curves the drop is spread over many steps and no single step can capture much of it, and on cliffs the judges' labels usually fall a few steps after the drop, so widening the window to $k=3$ roughly doubles the value captured without changing the order. The heuristics are informative baselines: ``first edit'' captures as much value as a mid-ranked judge because the decisive step is usually at or before the first edit (\S\ref{sec:shapes}), while ``step 1'' and ``before first symptom'' capture almost nothing, the first because the drop is rarely at the very start and the second because the symptom comes long after the cause. Agreement with labels (A$_k$) rewards the judges that label like the other judges: the contrast model and GPT-5.5 rank first and second by A$_1$ and in the middle by VC$_1$, while the GPT-6 judges, which capture the most value, rank below them by agreement because their early labels do not coincide with the later labels of the other lineage. Agreement with $t^*$ ($t^*_k$) is low throughout and is reported for reference only; it inherits the instability of a single decisive step on erosions (Appendix~\ref{arm:app:cheap}). Conclusion: a benchmark that ranks localizers by agreement with labels and one that ranks them by value captured would pick different winners, and the second criterion is the one that corresponds to what a controller gains; the measured curves make the second criterion computable for any proposed localizer, including ones not studied here.

\begin{table}[htbp]
\centering
{\scriptsize\setlength{\tabcolsep}{2.5pt}
\begin{tabular}[t]{lrrrrrrrrr}
\toprule
Localizer & $n$ & VC$_0$ & VC$_1$ & VC$_3$ & A$_0$ & A$_1$ & A$_3$ & $t^*_0$ & $t^*_3$ \\
\midrule
GPT-6 Astra & 60 & 0.05 & 0.10 & 0.18 & 0.29 & 0.31 & 0.36 & 0.08 & 0.22 \\
GPT-5.5 & 60 & 0.05 & 0.08 & 0.15 & 0.44 & 0.46 & 0.49 & 0.07 & 0.20 \\
GPT-6.1 Sol & 60 & 0.04 & 0.08 & 0.16 & 0.37 & 0.37 & 0.40 & 0.07 & 0.22 \\
\twoline{GPT-6 Luna}{(contrast)} & 60 & 0.03 & 0.07 & 0.11 & 0.39 & 0.41 & 0.48 & 0.07 & 0.13 \\
step 1 & 60 & 0.03 & 0.06 & 0.07 & 0.00 & 0.01 & 0.02 & 0.12 & 0.13 \\
OpenAI o3 & 60 & 0.01 & 0.04 & 0.10 & 0.25 & 0.26 & 0.32 & 0.02 & 0.13 \\
first edit & 60 & 0.02 & 0.03 & 0.07 & 0.26 & 0.27 & 0.28 & 0.03 & 0.07 \\
random step & 60 & 0.01 & 0.03 & 0.07 & -- & -- & -- & -- & -- \\
first verification & 60 & 0.00 & 0.02 & 0.04 & 0.01 & 0.02 & 0.05 & 0.02 & 0.05 \\
before first symptom & 60 & 0.00 & 0.01 & 0.01 & 0.05 & 0.08 & 0.14 & 0.00 & 0.02 \\
\bottomrule
\end{tabular}}
\caption{Localizers by value captured (VC$_k$), label agreement (A$_k$) and agreement with $t^*$; sorted by VC$_1$.}
\label{arm:tab:rq3-full}
\end{table}\vspace{-5pt}

\FloatBarrier
\clearpage
\subsection{All V-curves}
\label{arm:app:gallery}

\begin{figure}[H]
\centering
\includegraphics[width=0.96\textwidth]{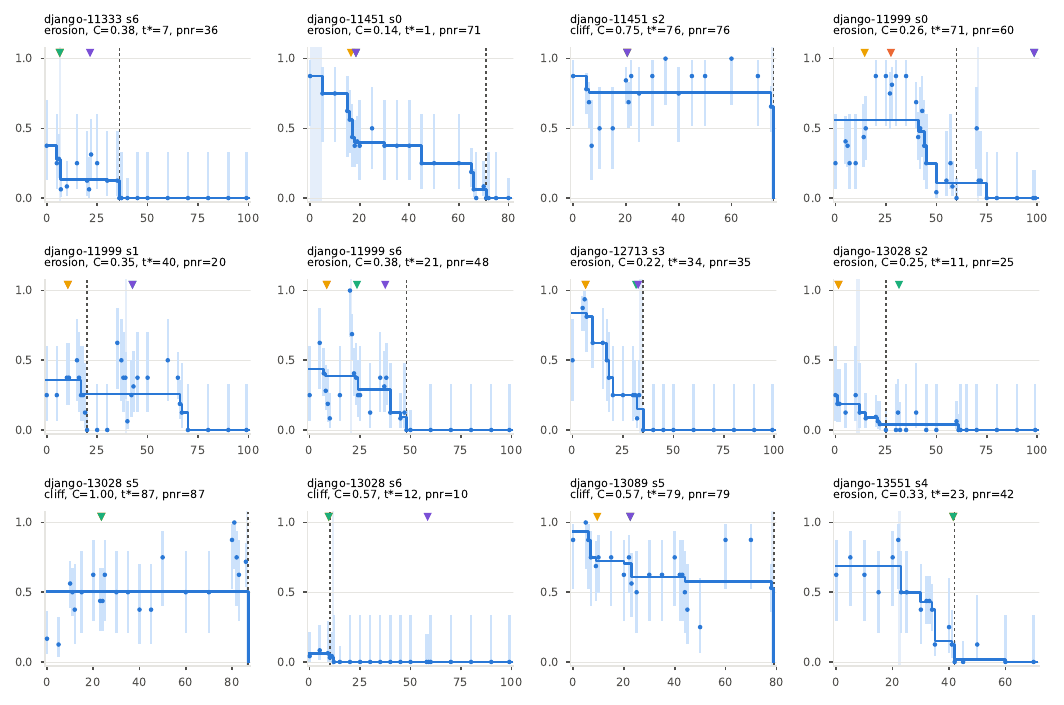}
\caption{V-curves of the second policy's frozen set (1 of 5); one diagnostic note per curve follows.}
\label{arm:fig:gallery_arm1}
\end{figure}

\begin{figure}[H]
\centering
\includegraphics[width=0.96\textwidth]{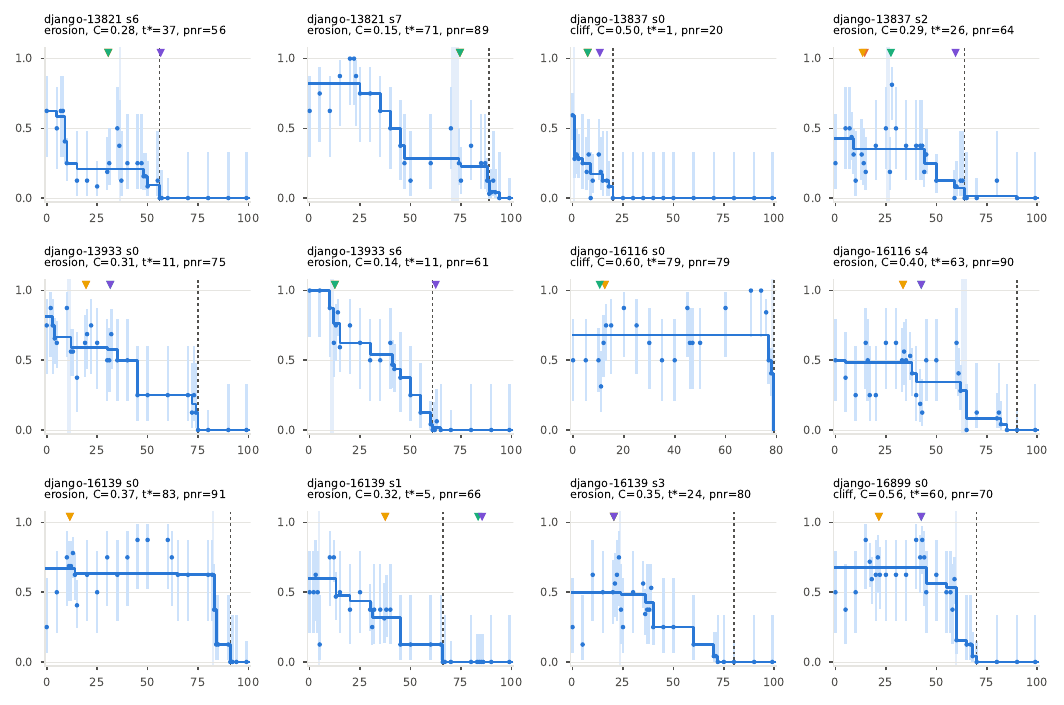}
\caption{V-curves of the second policy's frozen set (2 of 5); one diagnostic note per curve follows.}
\label{arm:fig:gallery_arm2}
\end{figure}

\begin{figure}[H]
\centering
\includegraphics[width=0.96\textwidth]{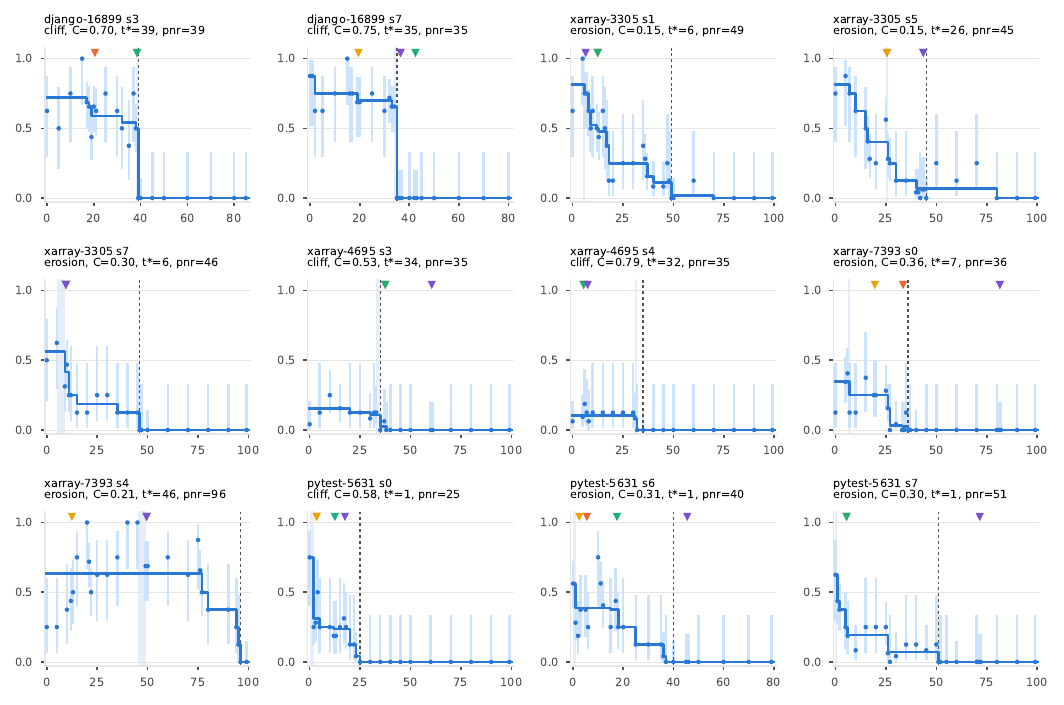}
\caption{V-curves of the second policy's frozen set (3 of 5); one diagnostic note per curve follows.}
\label{arm:fig:gallery_arm3}
\end{figure}

\begin{figure}[H]
\centering
\includegraphics[width=0.96\textwidth]{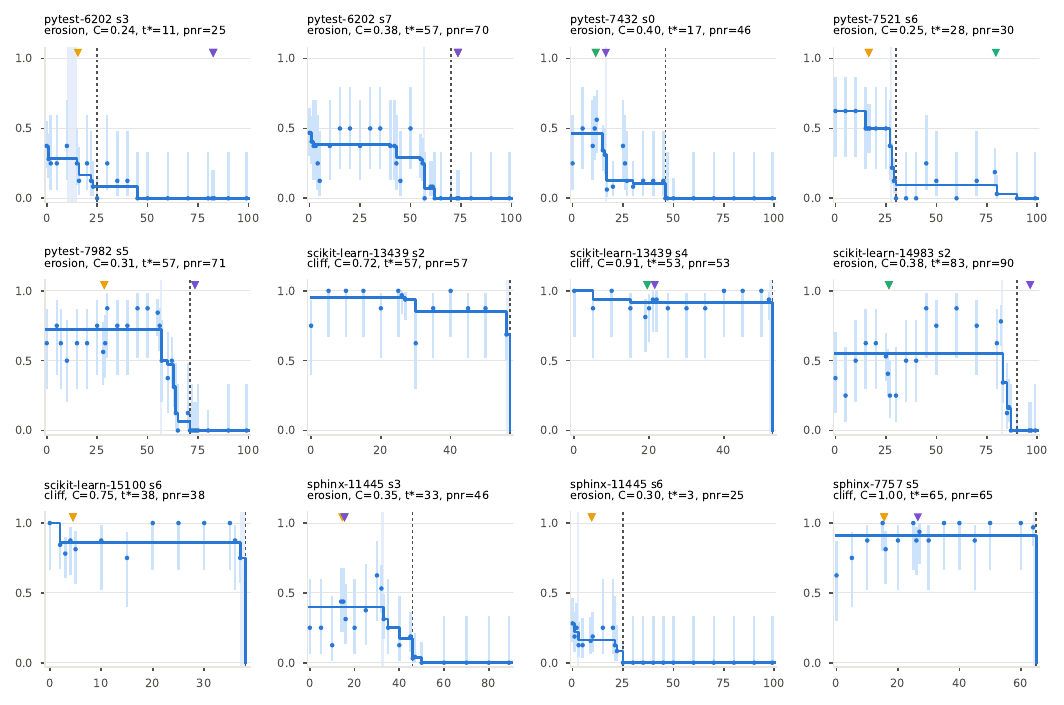}
\caption{V-curves of the second policy's frozen set (4 of 5); one diagnostic note per curve follows.}
\label{arm:fig:gallery_arm4}
\end{figure}

\begin{figure}[H]
\centering
\includegraphics[width=0.96\textwidth]{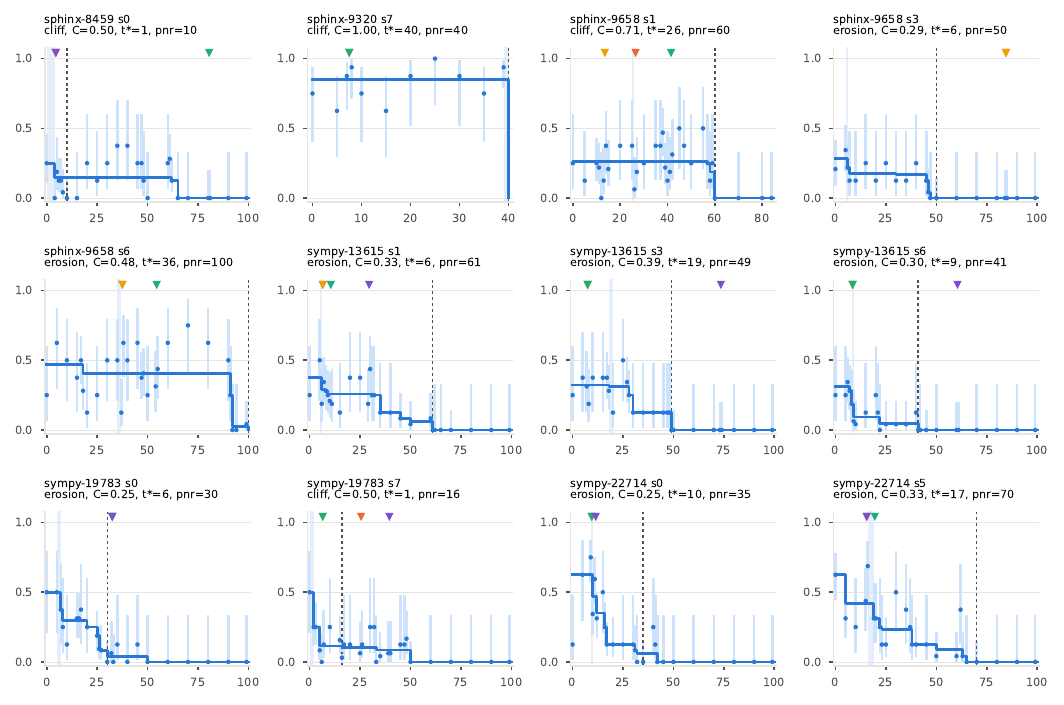}
\caption{V-curves of the second policy's frozen set (5 of 5); one diagnostic note per curve follows.}
\label{arm:fig:gallery_arm5}
\end{figure}

Figures~\ref{arm:fig:gallery_arm1}--\ref{arm:fig:gallery_arm5} show every curve of the second policy's frozen set in processing order, with the quantities of Table~\ref{arm:tab:alltraj} in each title: the isotonic fit (line), the raw estimates with their Wilson 95\% bars (dots), the decisive interval (shaded), the certified point of no return (dotted) and the main judges' labels (triangles above the axis: GPT-6.1 Sol orange, GPT-6 Astra yellow, GPT-5.5 green, OpenAI o3 purple). A few patterns are visible across the pages. Cliffs show a single step from a high plateau to zero (their decisive step sits at a median of 53\% of the run) with the labels clustered a few steps to its right; erosions show a staircase whose steps are each within the noise of a $K=8$ point, so their decisive interval is often wide (15 of the 19 wide intervals are erosions) and their point of no return lies a median of 28 steps after $t^*$ against 0 on cliffs. Curves that start low (a task with $p(x)$ near $0.25$) leave little to localize and their drops are correspondingly small. On some curves two judges' triangles of different lineages sit at the same step; these are the trajectories on which the lineages agree, and they are not confined to cliffs: the shared step is usually the one at which the wrong idea is acted upon, which both lineages recognize, whether or not the value had already moved. Conclusion: the gallery is the evidence behind Table~\ref{tab:rq1} in its rawest form; a reader can verify on any curve that the triangles rarely sit on the shaded interval: a main judge's label lies inside it on 6 of the 60 curves (1 cliffs, 5 erosions).

Across the 60 curves: 20 cliffs and 40 erosions; in 28 the value is flat (within $10\%$ of $\hat V(h_0)$) until the decisive interval, so everything before $t^*$ was harmless, while in 16 at least $30\%$ of the value had already leaked away before the largest drop, the erosion pattern; 19 decisive intervals are wider than one step (bisection stopped where neighbouring drops were indistinguishable at $K=8$); 9 curves lose the task only at the last step, where the agent submitted a patch that was still viable or ran out of budget; 43 show a rise of at least $0.25$ between consecutive raw points, which at $K\le32$ is usually within noise; in 5 of them the two 95\% intervals do not overlap, a genuine recovery that the monotone fit flattens and that is named in the notes below; the point of no return coincides with $t^*$ in 11. On the labels: 6 curves have at least one main-judge label inside the decisive interval, 18 have every label after it, and 9 have every label before it.

How to read a curve for diagnosis: the height at $t=0$ is the task's pass rate under this policy, so a curve that starts low has little to lose and its drops are small by construction; a flat stretch followed by one step to zero (a cliff) means that one decision ended the run, and the steps on the shaded interval are where to look in the transcript; a staircase (an erosion) means the agent spent its chances over many steps, and the decisive interval is only the largest of several losses; a drop that coincides with the first edit means the fix itself was wrong, a drop before it means the diagnosis was wrong before any code changed; the dotted line is the earliest step at which the agent's prospects were certifiably below $\epsilon$, so any label to its right marks something that could no longer matter; a curve whose dotted line is far to the right of its shaded interval is one on which the agent kept a small chance for a long time, typically because a later retry could still have worked. The triangles above the axis are the judges' labels; the question the paper asks is simply whether they sit on the shaded interval, and on most curves they do not.

\clearpage
\onecolumn
\ifdefined\nolinenumbers\nolinenumbers\fi
\paragraph{Figure~\ref{arm:fig:gallery_arm1}.}
\begin{multicols}{2}\scriptsize\begin{itemize}\setlength{\itemsep}{1pt}
\item \textbf{django-11333 s6} (erosion, $C=0.38$; $\hat V(h_0)=0.38$, $T=100$, L2): the decisive step 7 (resolved to one step) comes before the first edit (edit at 22); point of no return at 36 (29 steps later); the raw estimates rise by 0.25 from step 21 (1/16) to 22 (5/16), within the noise of the two samples; first failure signal at 27. Labels: GPT-6.1 Sol 7, GPT-6 Astra 7, GPT-5.5 7, OpenAI o3 22 (3 inside the decisive interval; 1 after it).
\item \textbf{django-11451 s0} (erosion, $C=0.14$; $\hat V(h_0)=0.88$, $T=81$, L0): the decisive step 1 (in an interval of 5 steps) comes before the first edit (edit at 19), with the value flat until then; point of no return at 71 (70 steps later); first failure signal at 22. Labels: GPT-6.1 Sol 19, GPT-6 Astra 17, GPT-5.5 19, OpenAI o3 19 (4 after it).
\item \textbf{django-11451 s2} (cliff, $C=0.75$; $\hat V(h_0)=0.88$, $T=76$, L0): the decisive step 76 (resolved to one step) comes after the first edit (edit at 21); the task is lost only at the last step; point of no return at 76 (same step); the raw estimates rise by 0.34 from step 15 (4/8) to 20 (27/32), within the noise of the two samples; first failure signal at 27. Labels: GPT-6.1 Sol 21, GPT-6 Astra 21, GPT-5.5 21, OpenAI o3 21 (4 before it).
\item \textbf{django-11999 s0} (erosion, $C=0.26$; $\hat V(h_0)=0.56$, $T=100$, L2): the decisive step 71 (resolved to one step) comes after the first edit (edit at 28), with 81\% of the value already gone before it; point of no return at 60 (-11 steps later); the raw estimates rise by 0.50 from step 60 (0/24) to 70 (4/8), a genuine recovery (the 95\% intervals do not overlap) that the monotone fit flattens; first failure signal at 35. Labels: GPT-6.1 Sol 28, GPT-6 Astra 15, GPT-5.5 99, OpenAI o3 99 (2 before it; 2 after the point of no return).
\item \textbf{django-11999 s1} (erosion, $C=0.35$; $\hat V(h_0)=0.36$, $T=100$, L2): the decisive step 40 (resolved to one step) comes before the first edit (edit at 43); point of no return at 20 (-20 steps later); the raw estimates rise by 0.62 from step 30 (0/8) to 35 (5/8), within the noise of the two samples; first failure signal at 53. Labels: GPT-6.1 Sol 11, GPT-6 Astra 11, GPT-5.5 43, OpenAI o3 43 (2 before it; 2 after the point of no return).
\item \textbf{django-11999 s6} (erosion, $C=0.38$; $\hat V(h_0)=0.44$, $T=100$, L2): the decisive step 21 (resolved to one step) comes on a run that never edits the repository; point of no return at 48 (27 steps later); the raw estimates rise by 0.75 from step 15 (2/8) to 20 (8/8), a genuine recovery (the 95\% intervals do not overlap) that the monotone fit flattens; no explicit failure signal. Labels: GPT-6.1 Sol 9, GPT-6 Astra 9, GPT-5.5 24, OpenAI o3 38 (2 before it; 2 after it).
\item \textbf{django-12713 s3} (erosion, $C=0.22$; $\hat V(h_0)=0.84$, $T=100$, L2): the decisive step 34 (in an interval of 2 steps) comes on a run that never edits the repository, with 82\% of the value already gone before it; point of no return at 35 (1 steps later); the raw estimates rise by 0.38 from step 0 (4/8) to 5 (28/32), within the noise of the two samples; no explicit failure signal. Labels: GPT-6.1 Sol 7, GPT-6 Astra 7, GPT-5.5 32, OpenAI o3 33 (4 before it).
\item \textbf{django-13028 s2} (erosion, $C=0.25$; $\hat V(h_0)=0.25$, $T=100$, L2): the decisive step 11 (in an interval of 2 steps) comes on a run that never edits the repository; point of no return at 25 (14 steps later); no explicit failure signal. Labels: GPT-6.1 Sol 2, GPT-6 Astra 2, GPT-5.5 32, OpenAI o3 none (2 before it; 1 after the point of no return; 1 ``no mistake'').
\item \textbf{django-13028 s5} (cliff, $C=1.00$; $\hat V(h_0)=0.50$, $T=87$, L0): the decisive step 87 (resolved to one step) comes after the first edit (edit at 39), with the value flat until then; the task is lost only at the last step; point of no return at 87 (same step); the raw estimates rise by 0.44 from step 5 (3/24) to 10 (18/32), a genuine recovery (the 95\% intervals do not overlap) that the monotone fit flattens; first failure signal at 58. Labels: GPT-6.1 Sol 24, GPT-6 Astra 24, GPT-5.5 24, OpenAI o3 none (3 before it; 1 ``no mistake'').
\item \textbf{django-13028 s6} (cliff, $C=0.57$; $\hat V(h_0)=0.06$, $T=100$, L2): the decisive step 12 (resolved to one step) comes on a run that never edits the repository, with 43\% of the value already gone before it; point of no return at 10 (-2 steps later); no explicit failure signal. Labels: GPT-6.1 Sol 10, GPT-6 Astra 10, GPT-5.5 10, OpenAI o3 59 (3 before it; 1 after the point of no return).
\item \textbf{django-13089 s5} (cliff, $C=0.57$; $\hat V(h_0)=0.94$, $T=79$, L0): the decisive step 79 (resolved to one step) comes after the first edit (edit at 23), with 43\% of the value already gone before it; the task is lost only at the last step; point of no return at 79 (same step); the raw estimates rise by 0.62 from step 50 (2/8) to 60 (7/8), within the noise of the two samples; first failure signal at 58. Labels: GPT-6.1 Sol 23, GPT-6 Astra 10, GPT-5.5 23, OpenAI o3 23 (4 before it).
\item \textbf{django-13551 s4} (erosion, $C=0.33$; $\hat V(h_0)=0.69$, $T=71$, L0): the decisive step 23 (resolved to one step) comes before the first edit (edit at 42), with the value flat until then; point of no return at 42 (19 steps later); the raw estimates rise by 0.25 from step 15 (4/8) to 20 (6/8), within the noise of the two samples; no explicit failure signal. Labels: GPT-6.1 Sol 42, GPT-6 Astra 42, GPT-5.5 42, OpenAI o3 none (3 after it; 1 ``no mistake'').
\end{itemize}\end{multicols}
\clearpage
\paragraph{Figure~\ref{arm:fig:gallery_arm2}.}
\begin{multicols}{2}\scriptsize\begin{itemize}\setlength{\itemsep}{1pt}
\item \textbf{django-13821 s6} (erosion, $C=0.28$; $\hat V(h_0)=0.62$, $T=100$, L2): the decisive step 37 (resolved to one step) comes on a run that never edits the repository, with 67\% of the value already gone before it; point of no return at 56 (19 steps later); the raw estimates rise by 0.25 from step 31 (4/16) to 35 (4/8), within the noise of the two samples; no explicit failure signal. Labels: GPT-6.1 Sol 31, GPT-6 Astra 31, GPT-5.5 31, OpenAI o3 57 (3 before it; 1 after the point of no return).
\item \textbf{django-13821 s7} (erosion, $C=0.15$; $\hat V(h_0)=0.82$, $T=100$, L2): the decisive step 71 (in an interval of 4 steps) comes after the first edit (edit at 55), with 66\% of the value already gone before it; point of no return at 89 (18 steps later); the raw estimates rise by 0.25 from step 10 (5/8) to 15 (7/8), within the noise of the two samples; first failure signal at 83. Labels: GPT-6.1 Sol 75, GPT-6 Astra 75, GPT-5.5 75, OpenAI o3 none (3 after it; 1 ``no mistake'').
\item \textbf{django-13837 s0} (cliff, $C=0.50$; $\hat V(h_0)=0.59$, $T=100$, L2): the decisive step 1 (resolved to one step) comes on a run that never edits the repository, with the value flat until then; point of no return at 20 (19 steps later); no explicit failure signal. Labels: GPT-6.1 Sol 8, GPT-6 Astra 8, GPT-5.5 8, OpenAI o3 14 (4 after it).
\item \textbf{django-13837 s2} (erosion, $C=0.29$; $\hat V(h_0)=0.43$, $T=100$, L2): the decisive step 26 (in an interval of 2 steps) comes before the first edit (edit at 52); point of no return at 64 (38 steps later); the raw estimates rise by 0.62 from step 27 (3/16) to 28 (13/16), a genuine recovery (the 95\% intervals do not overlap) that the monotone fit flattens; first failure signal at 64. Labels: GPT-6.1 Sol 15, GPT-6 Astra 14, GPT-5.5 28, OpenAI o3 60 (2 before it; 2 after it).
\item \textbf{django-13933 s0} (erosion, $C=0.31$; $\hat V(h_0)=0.81$, $T=100$, L2): the decisive step 11 (in an interval of 2 steps) comes before the first edit (edit at 32); point of no return at 75 (64 steps later); the raw estimates rise by 0.25 from step 5 (20/32) to 10 (7/8), within the noise of the two samples; first failure signal at 39. Labels: GPT-6.1 Sol 20, GPT-6 Astra 20, GPT-5.5 32, OpenAI o3 32 (4 after it).
\item \textbf{django-13933 s6} (erosion, $C=0.14$; $\hat V(h_0)=1.00$, $T=100$, L2): the decisive step 11 (in an interval of 2 steps) comes before the first edit (edit at 63); point of no return at 61 (50 steps later); first failure signal at 77. Labels: GPT-6.1 Sol 13, GPT-6 Astra 13, GPT-5.5 13, OpenAI o3 63 (3 after it; 1 after the point of no return).
\item \textbf{django-16116 s0} (cliff, $C=0.60$; $\hat V(h_0)=0.68$, $T=79$, L0): the decisive step 79 (resolved to one step) comes after the first edit (edit at 37), with 40\% of the value already gone before it; the task is lost only at the last step; point of no return at 79 (same step); the raw estimates rise by 0.38 from step 40 (4/8) to 45 (7/8), within the noise of the two samples; first failure signal at 41. Labels: GPT-6.1 Sol 13, GPT-6 Astra 13, GPT-5.5 11, OpenAI o3 none (3 before it; 1 ``no mistake'').
\item \textbf{django-16116 s4} (erosion, $C=0.40$; $\hat V(h_0)=0.50$, $T=100$, L2): the decisive step 63 (in an interval of 3 steps) comes after the first edit (edit at 43), with 44\% of the value already gone before it; point of no return at 90 (27 steps later); the raw estimates rise by 0.38 from step 10 (2/8) to 15 (5/8), within the noise of the two samples; first failure signal at 61. Labels: GPT-6.1 Sol 34, GPT-6 Astra 34, GPT-5.5 43, OpenAI o3 43 (4 before it).
\item \textbf{django-16139 s0} (erosion, $C=0.37$; $\hat V(h_0)=0.67$, $T=100$, L2): the decisive step 83 (resolved to one step) comes after the first edit (edit at 51), with the value flat until then; point of no return at 91 (8 steps later); the raw estimates rise by 0.25 from step 0 (2/8) to 5 (4/8), within the noise of the two samples; first failure signal at 59. Labels: GPT-6.1 Sol 12, GPT-6 Astra 12, GPT-5.5 none, OpenAI o3 none (2 before it; 2 ``no mistake'').
\item \textbf{django-16139 s1} (erosion, $C=0.32$; $\hat V(h_0)=0.60$, $T=100$, L2): the decisive step 5 (resolved to one step) comes before the first edit (edit at 87), with the value flat until then; point of no return at 66 (61 steps later); the raw estimates rise by 0.62 from step 5 (1/8) to 10 (6/8), within the noise of the two samples; first failure signal at 94. Labels: GPT-6.1 Sol 38, GPT-6 Astra 38, GPT-5.5 84, OpenAI o3 86 (2 after it; 2 after the point of no return).
\item \textbf{django-16139 s3} (erosion, $C=0.35$; $\hat V(h_0)=0.50$, $T=100$, L2): the decisive step 24 (resolved to one step) comes on a run that never edits the repository, with the value flat until then; point of no return at 80 (56 steps later); the raw estimates rise by 0.50 from step 5 (1/8) to 10 (5/8), within the noise of the two samples; no explicit failure signal. Labels: GPT-6.1 Sol 21, GPT-6 Astra 21, GPT-5.5 21, OpenAI o3 21 (4 before it).
\item \textbf{django-16899 s0} (cliff, $C=0.56$; $\hat V(h_0)=0.68$, $T=100$, L2): the decisive step 60 (resolved to one step) comes after the first edit (edit at 35); point of no return at 70 (10 steps later); the raw estimates rise by 0.38 from step 10 (4/8) to 15 (7/8), within the noise of the two samples; first failure signal at 53. Labels: GPT-6.1 Sol 22, GPT-6 Astra 22, GPT-5.5 43, OpenAI o3 43 (4 before it).
\end{itemize}\end{multicols}
\clearpage
\paragraph{Figure~\ref{arm:fig:gallery_arm3}.}
\begin{multicols}{2}\scriptsize\begin{itemize}\setlength{\itemsep}{1pt}
\item \textbf{django-16899 s3} (cliff, $C=0.70$; $\hat V(h_0)=0.72$, $T=86$, L0): the decisive step 39 (resolved to one step) comes at the first edit (edit at 39), with 30\% of the value already gone before it; point of no return at 39 (same step); the raw estimates rise by 0.38 from step 35 (3/8) to 37 (6/8), within the noise of the two samples; first failure signal at 45. Labels: GPT-6.1 Sol 21, GPT-6 Astra 39, GPT-5.5 39, OpenAI o3 none (2 inside the decisive interval; 1 before it; 1 ``no mistake'').
\item \textbf{django-16899 s7} (cliff, $C=0.75$; $\hat V(h_0)=0.88$, $T=81$, L0): the decisive step 35 (resolved to one step) comes before the first edit (edit at 37); point of no return at 35 (same step); the raw estimates rise by 0.25 from step 10 (6/8) to 15 (8/8), within the noise of the two samples; no explicit failure signal. Labels: GPT-6.1 Sol 37, GPT-6 Astra 20, GPT-5.5 43, OpenAI o3 37 (1 before it; 3 after the point of no return).
\item \textbf{xarray-3305 s1} (erosion, $C=0.15$; $\hat V(h_0)=0.81$, $T=100$, L2): the decisive step 6 (resolved to one step) comes before the first edit (edit at 48), with the value flat until then; point of no return at 49 (43 steps later); the raw estimates rise by 0.38 from step 0 (5/8) to 5 (8/8), within the noise of the two samples; first failure signal at 63. Labels: GPT-6.1 Sol 13, GPT-6 Astra 13, GPT-5.5 13, OpenAI o3 7 (4 after it).
\item \textbf{xarray-3305 s5} (erosion, $C=0.15$; $\hat V(h_0)=0.81$, $T=100$, L2): the decisive step 26 (resolved to one step) comes before the first edit (edit at 44), with 50\% of the value already gone before it; point of no return at 45 (19 steps later); the raw estimates rise by 0.31 from step 20 (2/8) to 25 (18/32), within the noise of the two samples; first failure signal at 51. Labels: GPT-6.1 Sol 26, GPT-6 Astra 26, GPT-5.5 44, OpenAI o3 44 (2 inside the decisive interval; 2 after it).
\item \textbf{xarray-3305 s7} (erosion, $C=0.30$; $\hat V(h_0)=0.56$, $T=100$, L2): the decisive step 6 (in an interval of 4 steps) comes before the first edit (edit at 34), with the value flat until then; point of no return at 46 (40 steps later); first failure signal at 37. Labels: GPT-6.1 Sol 10, GPT-6 Astra 10, GPT-5.5 10, OpenAI o3 10 (4 after it).
\item \textbf{xarray-4695 s3} (cliff, $C=0.53$; $\hat V(h_0)=0.16$, $T=100$, L2): the decisive step 34 (in an interval of 2 steps) comes before the first edit (edit at 56), with 31\% of the value already gone before it; point of no return at 35 (1 steps later); no explicit failure signal. Labels: GPT-6.1 Sol 38, GPT-6 Astra 38, GPT-5.5 38, OpenAI o3 61 (4 after the point of no return).
\item \textbf{xarray-4695 s4} (cliff, $C=0.79$; $\hat V(h_0)=0.11$, $T=100$, L2): the decisive step 32 (resolved to one step) comes on a run that never edits the repository; point of no return at 35 (3 steps later); no explicit failure signal. Labels: GPT-6.1 Sol 6, GPT-6 Astra 6, GPT-5.5 6, OpenAI o3 8 (4 before it).
\item \textbf{xarray-7393 s0} (erosion, $C=0.36$; $\hat V(h_0)=0.35$, $T=100$, L2): the decisive step 7 (resolved to one step) comes before the first edit (edit at 82), with the value flat until then; point of no return at 36 (29 steps later); the raw estimates rise by 0.25 from step 10 (1/8) to 15 (3/8), within the noise of the two samples; first failure signal at 86. Labels: GPT-6.1 Sol 34, GPT-6 Astra 20, GPT-5.5 82, OpenAI o3 82 (2 after it; 2 after the point of no return).
\item \textbf{xarray-7393 s4} (erosion, $C=0.21$; $\hat V(h_0)=0.64$, $T=100$, L2): the decisive step 46 (in an interval of 4 steps) comes before the first edit (edit at 50), with the value flat until then; point of no return at 96 (50 steps later); the raw estimates rise by 0.25 from step 13 (8/16) to 15 (6/8), within the noise of the two samples; first failure signal at 53. Labels: GPT-6.1 Sol 50, GPT-6 Astra 13, GPT-5.5 50, OpenAI o3 50 (1 before it; 3 after it).
\item \textbf{pytest-5631 s0} (cliff, $C=0.58$; $\hat V(h_0)=0.75$, $T=100$, L2): the decisive step 1 (in an interval of 2 steps) comes on a run that never edits the repository, with the value flat until then; point of no return at 25 (24 steps later); no explicit failure signal. Labels: GPT-6.1 Sol 4, GPT-6 Astra 4, GPT-5.5 13, OpenAI o3 18 (4 after it).
\item \textbf{pytest-5631 s6} (erosion, $C=0.31$; $\hat V(h_0)=0.56$, $T=80$, L0): the decisive step 1 (resolved to one step) comes before the first edit (edit at 46), with the value flat until then; point of no return at 40 (39 steps later); the raw estimates rise by 0.50 from step 6 (4/16) to 10 (6/8), within the noise of the two samples; first failure signal at 50. Labels: GPT-6.1 Sol 6, GPT-6 Astra 3, GPT-5.5 18, OpenAI o3 46 (3 after it; 1 after the point of no return).
\item \textbf{pytest-5631 s7} (erosion, $C=0.30$; $\hat V(h_0)=0.62$, $T=100$, L2): the decisive step 1 (resolved to one step) comes before the first edit (edit at 73), with the value flat until then; point of no return at 51 (50 steps later); first failure signal at 84. Labels: GPT-6.1 Sol 6, GPT-6 Astra 6, GPT-5.5 6, OpenAI o3 72 (3 after it; 1 after the point of no return).
\end{itemize}\end{multicols}
\clearpage
\paragraph{Figure~\ref{arm:fig:gallery_arm4}.}
\begin{multicols}{2}\scriptsize\begin{itemize}\setlength{\itemsep}{1pt}
\item \textbf{pytest-6202 s3} (erosion, $C=0.24$; $\hat V(h_0)=0.38$, $T=100$, L2): the decisive step 11 (in an interval of 5 steps) comes before the first edit (edit at 83); point of no return at 25 (14 steps later); the raw estimates rise by 0.25 from step 25 (0/24) to 30 (2/8), within the noise of the two samples; first failure signal at 86. Labels: GPT-6.1 Sol 83, GPT-6 Astra 16, GPT-5.5 83, OpenAI o3 83 (1 after it; 3 after the point of no return).
\item \textbf{pytest-6202 s7} (erosion, $C=0.38$; $\hat V(h_0)=0.47$, $T=100$, L2): the decisive step 57 (resolved to one step) comes before the first edit (edit at 74), with 47\% of the value already gone before it; point of no return at 70 (13 steps later); the raw estimates rise by 0.38 from step 45 (1/8) to 50 (4/8), within the noise of the two samples; first failure signal at 76. Labels: GPT-6.1 Sol 74, GPT-6 Astra 74, GPT-5.5 74, OpenAI o3 74 (4 after the point of no return).
\item \textbf{pytest-7432 s0} (erosion, $C=0.40$; $\hat V(h_0)=0.46$, $T=100$, L2): the decisive step 17 (resolved to one step) comes at the first edit (edit at 17), with 33\% of the value already gone before it; point of no return at 46 (29 steps later); the raw estimates rise by 0.29 from step 20 (2/24) to 25 (3/8), within the noise of the two samples; first failure signal at 32. Labels: GPT-6.1 Sol 12, GPT-6 Astra 12, GPT-5.5 12, OpenAI o3 17 (1 inside the decisive interval; 3 before it).
\item \textbf{pytest-7521 s6} (erosion, $C=0.25$; $\hat V(h_0)=0.62$, $T=100$, L2): the decisive step 28 (resolved to one step) comes before the first edit (edit at 80), with 40\% of the value already gone before it; point of no return at 30 (2 steps later); the raw estimates rise by 0.25 from step 40 (0/8) to 45 (2/8), within the noise of the two samples; first failure signal at 81. Labels: GPT-6.1 Sol 17, GPT-6 Astra 17, GPT-5.5 80, OpenAI o3 none (2 before it; 1 after the point of no return; 1 ``no mistake'').
\item \textbf{pytest-7982 s5} (erosion, $C=0.31$; $\hat V(h_0)=0.72$, $T=100$, L2): the decisive step 57 (resolved to one step) comes after the first edit (edit at 40), with the value flat until then; point of no return at 71 (14 steps later); the raw estimates rise by 0.25 from step 29 (10/16) to 30 (7/8), within the noise of the two samples; first failure signal at 42. Labels: GPT-6.1 Sol 29, GPT-6 Astra 29, GPT-5.5 74, OpenAI o3 74 (2 before it; 2 after the point of no return).
\item \textbf{scikit-learn-13439 s2} (cliff, $C=0.72$; $\hat V(h_0)=0.95$, $T=57$, L0): the decisive step 57 (resolved to one step) comes after the first edit (edit at 23); the task is lost only at the last step; point of no return at 57 (same step); the raw estimates rise by 0.25 from step 0 (6/8) to 5 (8/8), within the noise of the two samples; first failure signal at 26. Labels: GPT-6.1 Sol none, GPT-6 Astra none, GPT-5.5 none, OpenAI o3 none (4 ``no mistake'').
\item \textbf{scikit-learn-13439 s4} (cliff, $C=0.91$; $\hat V(h_0)=1.00$, $T=53$, L0): the decisive step 53 (resolved to one step) comes after the first edit (edit at 20), with the value flat until then; the task is lost only at the last step; point of no return at 53 (same step); no explicit failure signal. Labels: GPT-6.1 Sol 20, GPT-6 Astra 20, GPT-5.5 20, OpenAI o3 22 (4 before it).
\item \textbf{scikit-learn-14983 s2} (erosion, $C=0.38$; $\hat V(h_0)=0.55$, $T=100$, L2): the decisive step 83 (resolved to one step) comes after the first edit (edit at 27), with the value flat until then; point of no return at 90 (7 steps later); the raw estimates rise by 0.38 from step 40 (4/8) to 45 (7/8), within the noise of the two samples; no explicit failure signal. Labels: GPT-6.1 Sol 27, GPT-6 Astra 27, GPT-5.5 27, OpenAI o3 97 (3 before it; 1 after the point of no return).
\item \textbf{scikit-learn-15100 s6} (cliff, $C=0.75$; $\hat V(h_0)=1.00$, $T=38$, L0): the decisive step 38 (resolved to one step) comes after the first edit (edit at 19); the task is lost only at the last step; point of no return at 38 (same step); the raw estimates rise by 0.25 from step 15 (6/8) to 20 (8/8), within the noise of the two samples; first failure signal at 23. Labels: GPT-6.1 Sol 5, GPT-6 Astra 5, GPT-5.5 none, OpenAI o3 none (2 before it; 2 ``no mistake'').
\item \textbf{sphinx-11445 s3} (erosion, $C=0.35$; $\hat V(h_0)=0.40$, $T=90$, L0): the decisive step 33 (resolved to one step) comes before the first edit (edit at 63), with the value flat until then; point of no return at 46 (13 steps later); the raw estimates rise by 0.31 from step 10 (1/8) to 14 (7/16), within the noise of the two samples; first failure signal at 68. Labels: GPT-6.1 Sol 16, GPT-6 Astra 15, GPT-5.5 16, OpenAI o3 16 (4 before it).
\item \textbf{sphinx-11445 s6} (erosion, $C=0.30$; $\hat V(h_0)=0.28$, $T=100$, L2): the decisive step 3 (resolved to one step) comes before the first edit (edit at 83); point of no return at 25 (22 steps later); no explicit failure signal. Labels: GPT-6.1 Sol 10, GPT-6 Astra 10, GPT-5.5 none, OpenAI o3 none (2 after it; 2 ``no mistake'').
\item \textbf{sphinx-7757 s5} (cliff, $C=1.00$; $\hat V(h_0)=0.91$, $T=65$, L0): the decisive step 65 (resolved to one step) comes after the first edit (edit at 27), with the value flat until then; the task is lost only at the last step; point of no return at 65 (same step); no explicit failure signal. Labels: GPT-6.1 Sol 16, GPT-6 Astra 16, GPT-5.5 none, OpenAI o3 27 (3 before it; 1 ``no mistake'').
\end{itemize}\end{multicols}
\clearpage
\paragraph{Figure~\ref{arm:fig:gallery_arm5}.}
\begin{multicols}{2}\scriptsize\begin{itemize}\setlength{\itemsep}{1pt}
\item \textbf{sphinx-8459 s0} (cliff, $C=0.50$; $\hat V(h_0)=0.25$, $T=100$, L2): the decisive step 1 (in an interval of 4 steps) comes before the first edit (edit at 61), with the value flat until then; point of no return at 10 (9 steps later); the raw estimates rise by 0.25 from step 15 (0/8) to 20 (2/8), within the noise of the two samples; no explicit failure signal. Labels: GPT-6.1 Sol 5, GPT-6 Astra 5, GPT-5.5 81, OpenAI o3 5 (3 after it; 1 after the point of no return).
\item \textbf{sphinx-9320 s7} (cliff, $C=1.00$; $\hat V(h_0)=0.85$, $T=40$, L0): the decisive step 40 (resolved to one step) comes after the first edit (edit at 20), with the value flat until then; the task is lost only at the last step; point of no return at 40 (same step); the raw estimates rise by 0.25 from step 5 (5/8) to 7 (14/16), within the noise of the two samples; no explicit failure signal. Labels: GPT-6.1 Sol 8, GPT-6 Astra 8, GPT-5.5 8, OpenAI o3 none (3 before it; 1 ``no mistake'').
\item \textbf{sphinx-9658 s1} (cliff, $C=0.71$; $\hat V(h_0)=0.26$, $T=85$, L0): the decisive step 26 (resolved to one step) comes before the first edit (edit at 42), with the value flat until then; point of no return at 60 (34 steps later); the raw estimates rise by 0.25 from step 13 (2/16) to 14 (6/16), within the noise of the two samples; no explicit failure signal. Labels: GPT-6.1 Sol 27, GPT-6 Astra 14, GPT-5.5 42, OpenAI o3 none (1 before it; 2 after it; 1 ``no mistake'').
\item \textbf{sphinx-9658 s3} (erosion, $C=0.29$; $\hat V(h_0)=0.29$, $T=100$, L2): the decisive step 6 (resolved to one step) comes before the first edit (edit at 85), with the value flat until then; point of no return at 50 (44 steps later); no explicit failure signal. Labels: GPT-6.1 Sol 85, GPT-6 Astra 85, GPT-5.5 none, OpenAI o3 none (2 after the point of no return; 2 ``no mistake'').
\item \textbf{sphinx-9658 s6} (erosion, $C=0.48$; $\hat V(h_0)=0.47$, $T=100$, L2): the decisive step 36 (in an interval of 2 steps) comes before the first edit (edit at 55); point of no return at 100 (64 steps later); the raw estimates rise by 0.50 from step 37 (2/16) to 38 (10/16), a genuine recovery (the 95\% intervals do not overlap) that the monotone fit flattens; no explicit failure signal. Labels: GPT-6.1 Sol 38, GPT-6 Astra 38, GPT-5.5 55, OpenAI o3 none (3 after it; 1 ``no mistake'').
\item \textbf{sympy-13615 s1} (erosion, $C=0.33$; $\hat V(h_0)=0.38$, $T=100$, L2): the decisive step 6 (resolved to one step) comes before the first edit (edit at 30), with the value flat until then; point of no return at 61 (55 steps later); the raw estimates rise by 0.25 from step 0 (2/8) to 5 (4/8), within the noise of the two samples; no explicit failure signal. Labels: GPT-6.1 Sol 7, GPT-6 Astra 7, GPT-5.5 11, OpenAI o3 30 (4 after it).
\item \textbf{sympy-13615 s3} (erosion, $C=0.39$; $\hat V(h_0)=0.32$, $T=100$, L2): the decisive step 19 (in an interval of 2 steps) comes before the first edit (edit at 74), with the value flat until then; point of no return at 49 (30 steps later); the raw estimates rise by 0.38 from step 20 (1/8) to 25 (4/8), within the noise of the two samples; first failure signal at 92. Labels: GPT-6.1 Sol 8, GPT-6 Astra 8, GPT-5.5 8, OpenAI o3 74 (3 before it; 1 after the point of no return).
\item \textbf{sympy-13615 s6} (erosion, $C=0.30$; $\hat V(h_0)=0.31$, $T=100$, L2): the decisive step 9 (resolved to one step) comes before the first edit (edit at 61), with 40\% of the value already gone before it; point of no return at 41 (32 steps later); first failure signal at 65. Labels: GPT-6.1 Sol 9, GPT-6 Astra 9, GPT-5.5 9, OpenAI o3 61 (3 inside the decisive interval; 1 after the point of no return).
\item \textbf{sympy-19783 s0} (erosion, $C=0.25$; $\hat V(h_0)=0.50$, $T=100$, L2): the decisive step 6 (in an interval of 2 steps) comes before the first edit (edit at 33), with the value flat until then; point of no return at 30 (24 steps later); first failure signal at 35. Labels: GPT-6.1 Sol 33, GPT-6 Astra 33, GPT-5.5 33, OpenAI o3 33 (4 after the point of no return).
\item \textbf{sympy-19783 s7} (cliff, $C=0.50$; $\hat V(h_0)=0.50$, $T=100$, L2): the decisive step 1 (in an interval of 2 steps) comes before the first edit (edit at 43), with the value flat until then; point of no return at 16 (15 steps later); first failure signal at 75. Labels: GPT-6.1 Sol 26, GPT-6 Astra 7, GPT-5.5 7, OpenAI o3 40 (2 after it; 2 after the point of no return).
\item \textbf{sympy-22714 s0} (erosion, $C=0.25$; $\hat V(h_0)=0.62$, $T=100$, L2): the decisive step 10 (resolved to one step) comes on a run that never edits the repository, with the value flat until then; point of no return at 35 (25 steps later); the raw estimates rise by 0.50 from step 0 (1/8) to 5 (5/8), within the noise of the two samples; no explicit failure signal. Labels: GPT-6.1 Sol 10, GPT-6 Astra 10, GPT-5.5 10, OpenAI o3 12 (3 inside the decisive interval; 1 after it).
\item \textbf{sympy-22714 s5} (erosion, $C=0.33$; $\hat V(h_0)=0.62$, $T=100$, L2): the decisive step 17 (in an interval of 3 steps) comes on a run that never edits the repository, with 33\% of the value already gone before it; point of no return at 70 (53 steps later); the raw estimates rise by 0.38 from step 25 (3/24) to 30 (4/8), within the noise of the two samples; no explicit failure signal. Labels: GPT-6.1 Sol 16, GPT-6 Astra 16, GPT-5.5 20, OpenAI o3 16 (3 before it; 1 after it).
\end{itemize}\end{multicols}
\ifdefined\linenumbers\linenumbers\fi
\twocolumn